\documentclass[10pt,twocolumn,letterpaper]{article}

\usepackage{cvpr}              
\usepackage{graphicx}
\usepackage{siunitx}
\usepackage{enumitem}
\usepackage{multirow}
\usepackage{fp}
\usepackage{amsmath}
\usepackage{amssymb}
\usepackage{xcolor}
\usepackage{etoolbox}
\usepackage{algorithm}
\usepackage{algorithmicx}
\usepackage{algpseudocode}
\usepackage{wrapfig}
\usepackage{textcomp, gensymb}
\usepackage{calc}
\usepackage{booktabs}       
\usepackage{amsfonts}       
\usepackage{wrapfig}
\usepackage{graphicx}
\usepackage{caption}
\usepackage{subcaption}
\usepackage{booktabs}
\usepackage{multirow}
\usepackage{colortbl}
\usepackage{float}
\usepackage{tcolorbox}
\usepackage[super]{nth}
\usepackage{tikz}
\usepackage{pgfplots}
\usepackage{stmaryrd}
\usepackage{svg}
\usepackage{soul}

\usepackage[normalem]{ulem}
\usepackage{afterpage}
\usepackage{float}
\usepackage{fontawesome5} 
\definecolor{indianred}{rgb}{0.8, 0.36, 0.36}
\definecolor{bleudefrance}{rgb}{0.19, 0.55, 0.91}
\definecolor{forestgreen}{rgb}{0.0, 0.5, 0.0}
\definecolor{ashgrey}{rgb}{0.7, 0.75, 0.71}
\definecolor{darkorange}{rgb}{1.0, 0.55, 0.0}
\definecolor{darkorchid}{rgb}{0.6, 0.2, 0.8}

\definecolor{wdcolor}{RGB}{128, 0, 255}

\definecolor{wd_question_color}{RGB}{255, 0, 0}

\newcommand{\icoyes}{\textcolor{forestgreen}{\faCheckCircle}\xspace}
\newcommand{\icono}{\textcolor{ashgrey}{\faTimesCircle}\xspace}
\usepackage{tikz}

\usepackage{pifont}
\newcommand{\Gray}[0]{\rowcolor{gray!20}}
\newcommand{\Lgray}[0]{\rowcolor{gray!10}}

\newcommand{\icohalf}{\textcolor{darkorange}{\ding{51}\kern-0.65em\ding{55}}}

\definecolor{cvprblue}{rgb}{0.21,0.49,0.74}
\usepackage[pagebackref,breaklinks,colorlinks,allcolors=cvprblue]{hyperref}

\def\paperID{12221} 
\def\confName{CVPR}
\def\confYear{2025}

\usepackage[bb=boondox,bbscaled=.95,cal=boondoxo]{mathalfa} 
\usepackage{amssymb}

\title{ XLRS-Bench: Could Your Multimodal LLMs Understand Extremely Large Ultra-High-Resolution Remote Sensing Imagery?}

\author{%
  Fengxiang Wang\textsuperscript{1},
  Hongzhen Wang\textsuperscript{2}\ , 
  Mingshuo Chen\textsuperscript{3},
  Di Wang\textsuperscript{4,5}, 
  Yulin Wang\textsuperscript{2} ,\\
  Zonghao Guo\textsuperscript{2}, 
  Qiang Ma\textsuperscript{2},
  Long Lan\textsuperscript{1}, 
  Wenjing Yang\textsuperscript{1}\thanks{Corresponding authors}, 
  Jing Zhang\textsuperscript{4,6},
  Zhiyuan Liu\textsuperscript{2},
  Maosong Sun\textsuperscript{2}
  \\
  \textsuperscript{1} College of Computer Science and Technology, National University of Defense Technology, China \\
  \textsuperscript{2} Tsinghua University, China
  \textsuperscript{3} Beijing University of Posts and Telecommunications, China \\
  \textsuperscript{4} School of Computer Science, Wuhan University, China
   \textsuperscript{5} Zhongguancun Academy, China \\
   \textsuperscript{6} School of Artificial Intelligence, Wuhan University, China
   \\ \\
   {\centering}
    \url{https://xlrs-bench.github.io/}
}

\let\oldtwocolumn\twocolumn
\renewcommand\twocolumn[1][]{%
    \oldtwocolumn[{#1}{
\begin{center}
\centering
\includegraphics[width=1\linewidth]{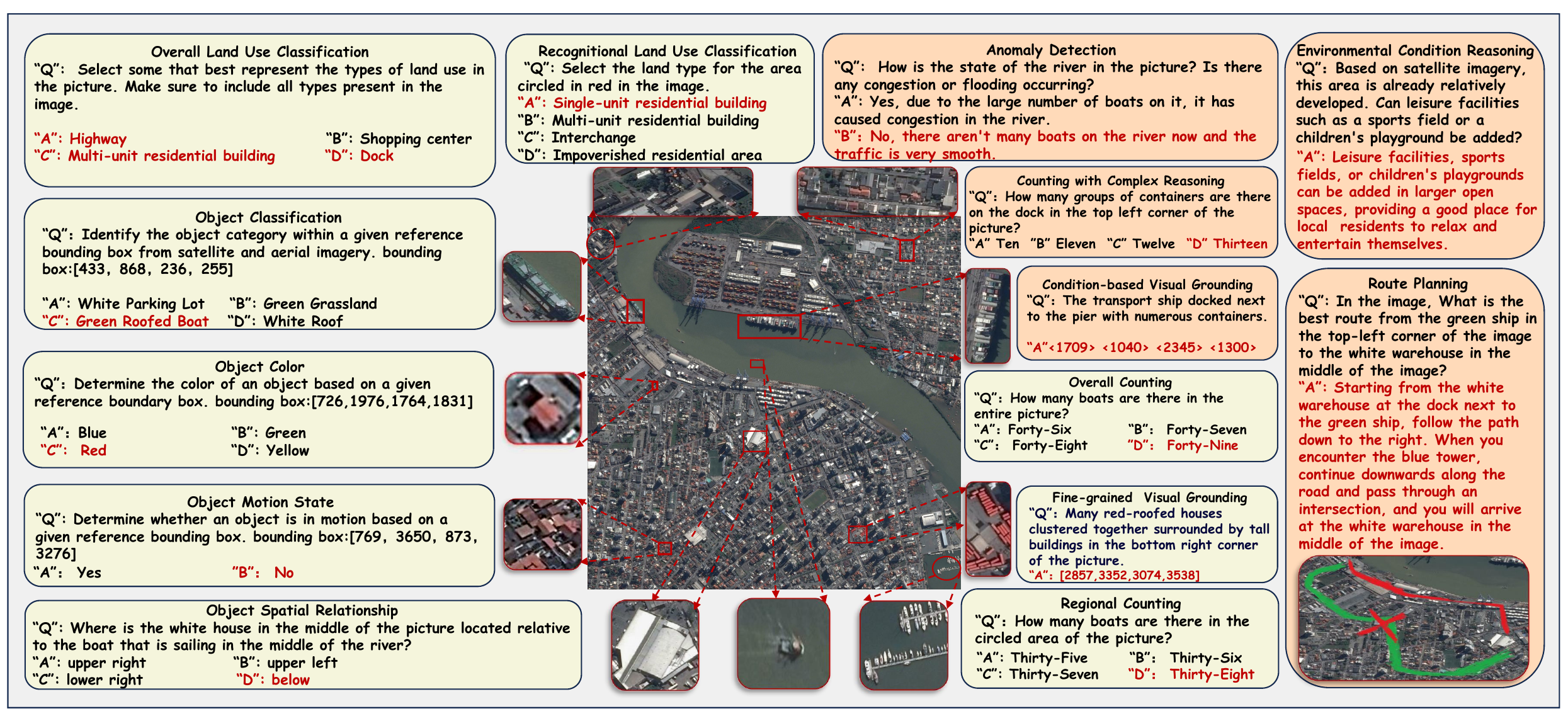}
\captionof{figure}{A typical example from our XLRS-Bench. XLRS-Bench focuses on extremely large ultra-high-resolution RS imagery, integrating over 10 multimodal vision-language perception and reasoning tasks within the same image.}
\label{fig:main}
\end{center}
    }]
}

\begin{document}
\sethlcolor{yellow}
\maketitle
\begin{abstract}
The astonishing breakthrough of multimodal large language models (MLLMs) has necessitated new benchmarks to quantitatively assess their capabilities, reveal their limitations, and indicate future research directions. However, this is challenging in the context of remote sensing (RS), since the imagery features ultra-high resolution that incorporates extremely complex semantic relationships. Existing benchmarks usually adopt notably smaller image sizes than real-world RS scenarios, suffer from limited annotation quality, and consider insufficient dimensions of evaluation. To address these issues, we present XLRS-Bench: a comprehensive benchmark for evaluating the perception and reasoning capabilities of MLLMs in ultra-high-resolution RS scenarios. XLRS-Bench boasts the largest average image size (8500$\times$8500) observed thus far, with all evaluation samples meticulously annotated manually, assisted by a novel semi-automatic captioner on ultra-high-resolution RS images. On top of the XLRS-Bench, 16 sub-tasks are defined to evaluate MLLMs' 10 kinds of perceptual capabilities and 6 kinds of reasoning capabilities, with a primary emphasis on advanced cognitive processes that facilitate real-world decision-making and the capture of spatiotemporal changes. The results of both general and RS-focused MLLMs on XLRS-Bench indicate that further efforts are needed for real-world RS applications. We have open-sourced  \href{https://xlrs-bench.github.io/}{\color{magenta}XLRS-Bench} to support further research in developing more powerful MLLMs for remote sensing.
\end{abstract}    
\section{Introduction}
\label{sec:intro}

Recent advancements in multimodal large language models (MLLMs)~\cite{gemini,gpt4,llama,minigpt,qwen} have significantly enhanced visual understanding and reasoning. As real-world applications require more detailed visual processing, many MLLMs~\cite{InternLM-XComposer,llava-uhd,llava-next,slime,mini-gemini} have been developed to improve understanding of high-resolution images. To fully assess and leverage their potential, benchmarking is essential, leading to the creation of various related datasets~\cite{mmbench,mmerealworld,mme,mmtbench,seedbench,cvbench,blink,visitbench}.

Remote sensing (RS) images have become essential for monitoring and understanding human environments, driving advancements in applications like precision agriculture \cite{agriculture_1}, urban planning \cite{urban_plan}, and disaster assessment \cite{disa_ass}. As such, assessing the performance of MLLMs in this field is crucial. However, the high resolution and complex semantic relationships in RS imagery make evaluating MLLMs in real-world remote sensing contexts particularly challenging. While recent studies \cite{vrsbench, lhrsbot, earthgpt} have proposed benchmarks and metrics to assess MLLM performance in RS, these efforts remain limited in three key areas:

\textbf{Image Size.}
Real ultra-high resolution RS images often capture scenes at the least city-level or above with large size (\textit{e.g.}, 10,000$\times$10,000). However, benchmarks like VRSBench~\cite{vrsbench} merely utilize 512$\times$512 image slices to evaluate the performance of MLLMs on cross-modal perception and understanding (like visual question answering (VQA), image captioning and visual grounding), which fails to provide a comprehensive assessment of the models' capabilities to capture long-range spatial-semantic relationships. 

\textbf{Human Annotation.} 
The prohibitive labor costs associated with comprehensive manual annotation significantly limit the scalability of existing RS multimodal benchmarks, such as LHRS-Bench~\cite{lhrsbot} (only 108 RS images paired with 690 QA pairs). In response, GPT-based annotators have become widely employed to automate data construction for both training and evaluating MLLMs, thus generating substantial data volumes. Following this trend, larger-scale RS benchmarks have been created~\cite{vrsbench}. However, the absence of stringent manual calibration in automated annotation processes may lead to hallucinations and errors, compromising the quality of the data. Additionally, empirical evidence~\cite{mmerealworld} indicates that benchmarks annotated by MLLMs, such as GPT-4V, often incorporate linguistic biases, which may inadvertently boost the performance of corresponding models on these benchmarks, despite attempting to manually adjust content biases.
    
\textbf{Evaluation Dimension.}
Existing RS benchmarks~\cite{vrsbench, lhrsbot, earthgpt} primarily concentrate on basic perceptual capabilities such as attribute recognition, semantic discrimination, and spatial localization.
However, existing benchmarks fail to fully assess MLLMs' interactive perception and complex reasoning, which are crucial for tasks like visual grounding, path planning, and intent inference.

\begin{figure}[t!]
\centering
\includegraphics[width=1\linewidth]{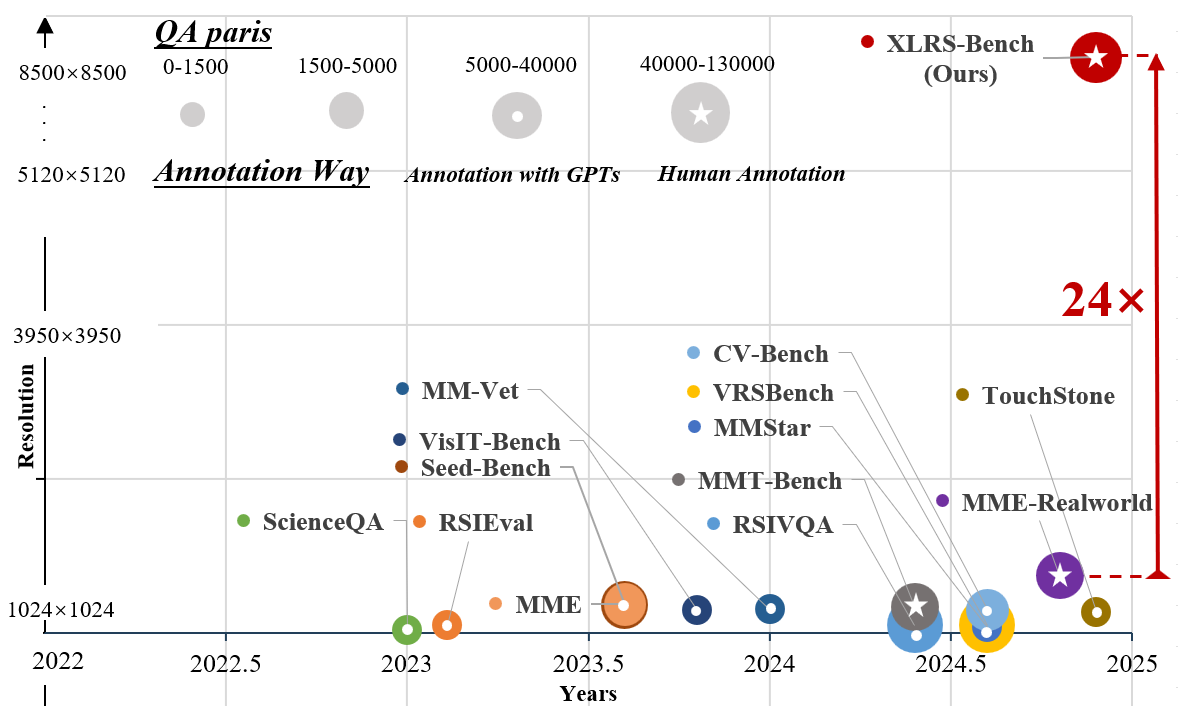}
\caption{Advantages of XLRS-Bench: XLRS-Bench boasts an average image size that is 24 times larger than existing datasets.}
\label{fig:abstract}
\vspace{-3mm}
\end{figure}

To address these challenges, we introduce XLRS-Bench, a benchmark designed to assess the perception and reasoning capabilities of MLLMs on ultra-high-resolution RS scenarios. We first collected 1,400 real-world ultra-high-resolution RS images with large sizes (8,500$\times$8,500 pixels on average). Among them, 530 images come from detection datasets (\textit{e.g.}, DOTA-v2~\cite{dota}) and 870 images are from segmentation datasets (\textit{e.g.}, MiniFrance~\cite{minifrance}). Then, 45 experts are involved in hand-crafted annotation and cross-verification for 16 sub-tasks. The annotations are presented as three vision-language formats including VQA, captioning and visual grounding. Specifically, we implemented a semi-automated pipeline for image captioning, by leveraging GPT-4o for pre-annotation and human for verification. We additionally invite 10 specialists in the field of MLLM for further quality control. Finally, XLRS-Bench incorporates 45,942 annotations covering 10 ability indicators for perception and 6 for reasoning, comprising 32,389 VQA pairs, 12,619 visual grounding instances, and 934 detailed captions, as shown in Fig. \ref{fig:main}. Based on XLRS-Bench, we evaluated a series of general MLLMs and an RS-specific MLLM and conducted an in-depth analysis. As shown in Fig.~\ref{fig:abstract}, the key advantages of XLRS-Bench over existing benchmarks include:

\begin{itemize}
    \item \textbf{Ultra-high Resolution}. XLRS-Bench features the largest image sizes available, 10$\thicksim$20$\times$ than that of existing datasets, with 840 images out of all images at a resolution of 10,000$\times$10,000 pixels.
    \item  \textbf{High-quality Annotation}. All the annotations are human involved and manually verified through iterations, resulting in a high-quality benchmark for evaluating MLLMs on real ultra-high-resolution RS scenarios.
    \item \textbf{Comprehensive Evaluation Dimensions}: XLRS-Bench covers 10 perception indicators and 6 reasoning dimensions to assess MLLMs' capabilities, encompassing 16 sub-tasks with a total of 45,942 questions. Especially, XLRS-Bench includes complex reasoning tasks to explore MLLMs' potential in conducting planning and change detection in long spatial-temporal RS scenarios.  
\end{itemize}

In summary, our main contributions are as follows:
\begin{itemize}
    \item We establish XLRS-Bench, a new benchmark for assessing the perception and reasoning capabilities of MLLMs across 16 sub-tasks in ultra-high-resolution RS scenarios.
    \item We develop a semi-automated pipeline to expand detailed captions, showcasing the potential of the scalability across various RS tasks.
    \item We evaluate prominent MLLMs across computer vision and RS fields on XLRS-Bench, providing insights into future directions for understanding real RS scenarios.
\end{itemize}

\section{Related Work}
\label{sec:related}

\textbf{General Multimodal Benchmark.} 
Large-scale vision-language models (VLMs) have demonstrated great potential in multimodal tasks like complex scene understanding and visual sentiment analysis. Recently, multimodal evaluation datasets have been developed to assess the capabilities of these models quantitatively. However, previous benchmarks mainly focused on specific domains and assessed models on one or a few tasks (\textit{e.g.}, captioning~\cite{coco_caption, nocaps, flickr30k} or VQA~\cite{gqa,ok-vqa,vqav2,vizwiz,textvqa,youcook2}). The evaluation for large-scale VLMs requires more comprehensive benchmarks. Therefore, MME ~\citep{mme} offers a benchmark spanning 14 perceptual and cognitive tasks, and MMBench ~\citep{mmbench} includes over 3,000 questions across 20 skill dimensions, like object localization and social reasoning. Seed-Bench~\citep{seedbench} expands the sample size with 19,000 questions, and MMT-Bench~\citep{mmtbench} incorporates real-world data from areas like autonomous driving and embedded AI. MME-Realworld~\citep{mmerealworld}, extending to five real-world scenarios, provides the highest resolution natural scene benchmark so far, with images at 2,000$\times$1,500. However, these general-purpose benchmarks have two main limitations: limited data and text annotations for RS scenarios, and smaller image sizes compared to real RS data. Even MME-Realworld, with the most RS data, includes only three types of QA pairs that are related to RS scenarios. Furthermore, its average resolution of 2,000$\times$1,500 is still far below the required of real RS tasks (\textit{e.g.}, HRSCD~\cite{hrscd} at 10,000$\times$10,000 for RS segmentation).

\textbf{Remote Sensing Multimodal Benchmark.} 
With the recent advancements of large multimodal models in general domains, the RS field has also experienced rapid growth on MLLMs~\cite{geochat, lhrsbot}. This has led to the creations of relevant evaluation benchmarks. RSIEval~\cite{rsgpt} offers 100 human-annotated captions and 936 visual question-answer pairs, mainly for image captioning and VQA tasks. LHRS-Bench~\cite{lhrsbot} provides 108 images and 690 questions, with VQA questions spanning five dimensions. VLEO-BENCH~\cite{VLEO-BENCH} covers scenarios such as urban monitoring, disaster relief, land use, and conservation, evaluating VLMs in scene understanding, localization, counting, and change detection tasks. What's more, RSSA~\cite{h2rsvlm} introduces a benchmark focused on hallucination, while FIT-RSRC~\cite{skysensegpt} targets understanding object relationships in RS scenes. VRSBench~\cite{vrsbench} includes 29,614 images with paired descriptions, 52,472 object references, and 123,221 question-answer pairs. Nevertheless, these benchmarks have two main limitations. First, they lack a diverse range of sub-tasks, resulting in less objective and comprehensive evaluations. Second, their image sizes are typically limited to 512$\times$512, which is far smaller than the requirements of real RS tasks (\textit{e.g.}, DOTA~\cite{dota} images reach 7,000$\times$4,000 for detection tasks, HRSCD~\cite{hrscd} images reach 10,000$\times$10,000 for segmentation), hindering the evaluation of MLLMs for long-range spatial-semantic cognition in real-world RS scenarios.

\textbf{Multimodal Large Language Model.}
Leveraging advanced LLMs like GPTs~\citep{gpt4} and LLaMA~\citep{llama}, MLLMs have shown impressive capabilities~\cite{learning,DecodingTrust}. Proprietary models such as Gemini~\citep{gemini} and GPT-4o~\citep{gpt4} exhibit strong understanding and reasoning skills, while open-source models, including Qwen-VL~\citep{qwen}, InternLM-XComposer~\citep{InternLM-XComposer}, MiniCPM~\citep{minicpm}, LLaVA~\citep{visualllama} and MiniGPT-4~\citep{minigpt}, also demonstrate notable performance. Since not specifically optimized for high-resolution tasks, these models only support image resolutions from 2K to 4K.
Recently, several MLLMs have addressed low-resolution constraints to enable high-resolution processing. For example, LLaVA-Next~\citep{llava-next} divides high-resolution images into patches, encodes each patch independently, and then links the patch tokens with the global image tokens. Furthermore, models such as Monkey~\citep{monkey} and LLaVA-UHD~\citep{llava-uhd} compress these patches to avoid redundant tokens. Alternatively, Mini-Gemini~\citep{mini-gemini} uses a dual encoder—one for high-resolution images and another for low-resolution embeddings, while Cambrian~\cite{cambrian} uses learnable latent queries to interact with multiple visual features, and SliME~\citep{slime} compresses local patches twice, preserving global features.
In the RS field~\cite{selectivemae, efficient,ship_my,roma}, MLLMs have advanced as well. Geochat~\citep{geochat}, based on LLaVA-1.5~\citep{llama}, enables multi-task dialogues with RS images. LHRS-Bot~\citep{lhrsbot} enhances image understanding through a multi-level vision-language alignment strategy and curriculum learning, and EarthGPT~\citep{earthgpt} unifies multisensor interpretation tasks for general RS understanding.

Despite these advancements, both RS and general-domain MLLMs face challenges. General MLLMs, while focused on high-resolution processing, lack rigorous testing on larger-size benchmarks in real-world RS scenarios. Meanwhile, RS models have only been tested on benchmarks with small images (\textit{e.g.}, 512$\times$512). In contrast, our XLRS-Bench provides a rigorous test bed to assess models' perception and reasoning capabilities on large-size images from real RS environments.
\section{XLRS-Bench}
\label{sec:method}

XLRS-Bench stands out from existing multimodal understanding benchmarks with four key features:
i) the largest average image size of 8,500$\times$8,500, featuring extensive 10,000$\times$10,000 imagery; ii) 16 sub-tasks designed to evaluate MLLMs’ capabilities in ultra-high-resolution RS scenes; iii) strict quality control to ensure sample accuracy and validity;  iv) bilingual support for evaluating VLM performance in both English and Chinese. Further details on the construction of XLRS-Bench are provided below.

\subsection{Evaluation Dimensions}
\begin{figure}[tbp]

\centering
\includegraphics[width=\linewidth]{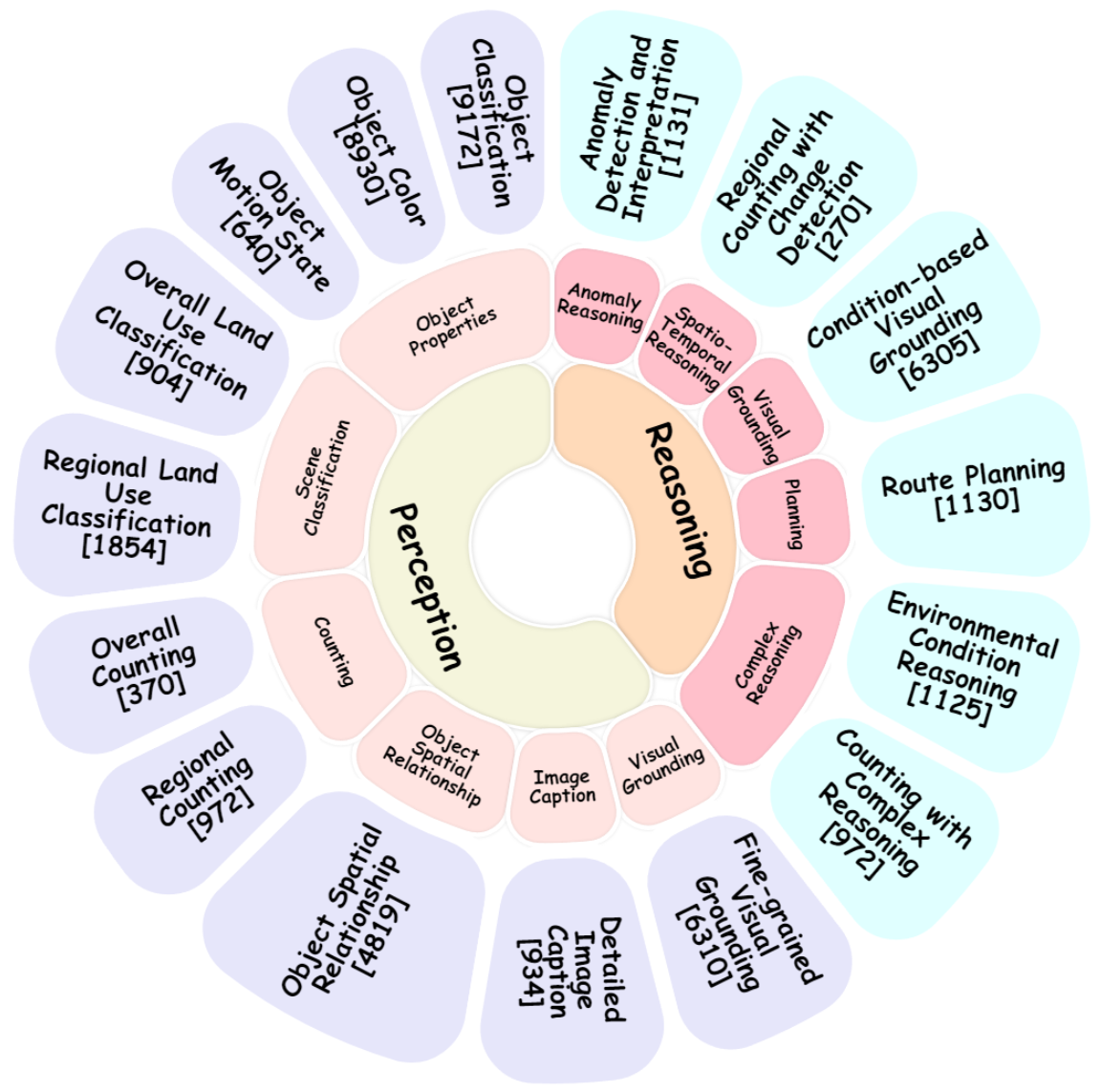}
\caption{
XLRS-Bench evaluates the perception and reasoning capabilities of MLLMs across three levels and 16 sub-tasks.
}
\label{fig:dimensions}
\vspace{-5mm}
\end{figure}

Humans have exceptional perception and reasoning abilities that are foundational to complex cognition. Perception gathers sensory input, while reasoning draws conclusions. Together, these abilities enable tasks like object recognition, problem-solving, and decision-making. To achieve true artificial general intelligence in the real ultra large-scale RS scenarios, MLLMs must also exhibit robust perception and reasoning capabilities. We therefore classify perception and reasoning as primary (\textbf{L-1}) abilities, further refined into 11 \textbf{L-2} and 16 \textbf{L-3} capability dimensions. Figure \ref{fig:dimensions} illustrates this classification, whose detailed explanations are presented in the following text. 

\subsubsection{Perception}
We employ two typical tasks: image captioning and VQA, to measure the L-1 perception ability. To assess the capabilities in capturing fine-grained information for large-size ultra-high-resolution RS scenes, at the L-2 level, we additionally definite 6 evaluation aspects, which are further divided into 10 more refined indicators in L-3 grade. Next, we separately introduce the L-2 capabilities,  with the L-3 tasks are also presented.

\begin{itemize}
    \item \textbf{Image Captioning.} In this task, we go beyond simple single-sentence captioning tasks to\ \textit{Detailed Image Captioning}, as large-size ultra-high-resolution RS images contain abundant details. 
    \item \textbf{Scene Classification.} Scene Classification is evaluated through two L-3 sub-tasks: \textit{Overall Land Use Classification} and \textit{Regional Land Use Classification}. The former focuses on global image information with multiple-choice questions, while the latter examines detailed scene understanding in localized areas using single-choice questions.
    \item \textbf{Counting.} This capability contains two L-3 dimensions: \textit{Overall Counting and Regional Counting}. Overall Counting requires comprehensive object recognition for the whole image, which is challenging even for humans. Regional Counting narrows the focus to smaller areas. 
    \item \textbf{Object Spatial Relationship.} This dimension requires the model to identify two specified objects and determine their spatial relationship, demanding advanced spatial perception and understanding ability.
    \item \textbf{Object Properties.} Object understanding is assessed across three L-3 dimensions: \textit{Object Classification, Object Color, and Object Motion State.} We exclude properties like shape and size~\cite{vrsbench}, as they fall under Object Classification. Here, we conduct a separate evaluation of Object Color, since it differs significantly from attributes like shape and size. Object Motion State is especially relevant, as it tests reasoning through context clues, such as inferring a ship's movement from its wake. 
    \item \textbf{Visual Grounding.} This dimension tests the model’s ability to precisely locate objects.
    We assess \textit{Fine-grained Visual Grounding}, which is essential in RS, where the model is required to detect small object types in large-scale high-resolution scenarios. 
    Notably, some objects in these large-size images may be as small as 5-10 pixels, which presents a challenge for current MLLMs that primarily support 2K resolution, compressing such objects to just 1-2 pixels.
\end{itemize}

\begin{table*}[t!]
\footnotesize
    \centering
        \caption{Comparison between existing vision-language benchmarks and our benchmark. \icoyes, \icono and \icohalf separately represent the annotations are machine generated, manually written and semi-automated, \textit{i.e.}, machine generation followed by human verification.}
    \resizebox{0.98\textwidth}{!}{
    \begin{tabular}{l|c|c|cc|cc|cc}
        \toprule
        \multirow{2}{*}{Dataset} & \multirow{2}{*}{Average Resolution} & \multirow{2}{*}{Remote Sensing} & \multicolumn{2}{c}{Detailed Caption} & \multicolumn{2}{c}{Visual Grounding} & \multicolumn{2}{c}{VQA} \\
        \cline{4-5} \cline{6-7} \cline{8-9}
         &  &  & Volume & Average Length & Volume & Annotation Method & Volume & Annotation Method   \\
        \hline
MMStar~\cite{mmstar} & 512$\times$375& N & - & - & - & \icono & 1,500 & \icono  \\
ScienceQA~\cite{scienceqa} & 378$\times$249&  N & -&- & - & \icono &21,000 & \icono \\
MM-Vet~\cite{mmvet} & 1,200$\times$675& N  & -& - &- & \icono &218 & \icono  \\
Seed-Bench~\cite{seedbench}  & 1,024$\times$931& N  &- &- &- & \icono &19,242 & \icono  \\
MMT-Bench~\cite{mmtbench} & 2,365$\times$377&  N & - & - & - & \icono & 32,325 & \icono \\
TouchStone~\cite{touchstone} & 897$\times$803& N & - & - & - & \icono & 908 & \icono \\
VisIT-Bench~\cite{visitbench}  & 765$\times$1,024&  N & - & - &- & \icono & 1,159 & \icono \\
BLINK~\cite{blink} & 620$\times$1,024& N & -& - & - & \icono & 3,807 & \icono  \\
CV-Bench~\cite{cvbench} & 1,024$\times$768&  N & - & - & - & \icono & 2,638&\icono \\ 
MME~\cite{mme} & 1,161$\times$840&  N & - & - & - & \icono & 2,374 & \icoyes  \\
MMBench~\cite{mmbench}  & 512$\times$270 &  N & - & - & - & \icono & 3,217 & \icoyes \\ 
MME-Realworld~\cite{mmerealworld}  & 2,000$\times$1,500&  N & -& - &- & \icono & 29,429 & \icoyes \\ \midrule
        UCM-Captions~\cite{ucmcaption} & 250$\times$250 & Y & 10,500  & 12 & - & \icono & - & \icono  \\
        RSICD \cite{rsicd} & 500$\times$500 & Y & 54,605  & 12 & - & \icono & - & \icono  \\
        RSICap~\cite{rsgpt} & 512$\times$512 & Y & 2,585 & 60 & - & \icono & - & \icono\\
        RSVG~\cite{rsvg} & 800$\times$800 & Y & -& - & 7,933 & \icohalf & - & \icono \\
        DIOR-RSVG \cite{diorrsvg} & 800$\times$800 & Y & - & - & 38,320 & \icohalf & - & \icono  \\
        RRSIS-D \cite{rrsisd} & 800$\times$800 &Y & - & - & 17,402 & \icohalf & - & \icono \\
        RSVQA-HR \cite{rsvqahr} & 1,024$\times$1,024 & Y & - & -&  -& \icono & 1,066,316 & \icono  \\
        RSIVQA \cite{rsivqa} & 512$\times$512 & Y & - & - & - & \icono & 111,134 & \icohalf \\
        RSIEval \cite{rsgpt} & 512$\times$512 & Y & -& - & - & \icono & 933 & \icohalf\\
        VRSBench \cite{vrsbench}                   &  512$\times$512  & Y & 29,614 & 52 & 52,472 & \icohalf & 123,231 & \icohalf  \\ \midrule
       XLRS-Bench& 8,500 $\times$8,500 &Y&934&379&12,619& \icoyes&32,389& \icoyes\\
        \bottomrule
    \end{tabular}
    }
    \label{tab:main}
    \vspace{-5mm}
\end{table*}

\subsubsection{Reasoning}
 Unlike previous assessments of simple reasoning \cite{vrsbench}, we would like to evaluate the model's ability to perform Complex Reasoning based on rich visual information in ultra-high-resolution RS scenarios. 
 We designed challenging 5 L-2 dimensions and 6 L-3 dimensions to test reasoning in realistic ultra-high-resolution RS scenarios:
\begin{itemize}
\item \textbf{Complex Reasoning. } It is divided into two L-3 dimensions, \textit{Environmental Condition Reasoning} and \textit{Counting with Complex Reasoning}. The former dimension leverages the comprehension information in ultra-high-resolution images, requiring the model to make contextual inferences.
\textit{Counting with Complex Reasoning} requires not only perception-based counting but also the ability to perform complex reasoning.
 \item \textbf{Anomaly Reasoning.} The model is expected to detect anomalies and predict potential risks using contextual clues, named \textit{Anomaly Detection and Interpretation}. For instance, in a town near a mountain, the model should identify exposed forest land and, considering climate conditions, predict possible landslide risks.
 \item \textbf{Planning.} Beyond reasoning, the model should plan a route based on provided conditions, named \textit{Route Planning} (L-3 ability). Given a starting point described in natural language, the model must locate it accurately on the image and select an optimal route based on criteria like shortest distance or minimal intersections.
 \item \textbf{Visual Grounding.} We also include conditional grounding to evaluate reasoning-based localization abilities, named \textit{Condition-based Visual Grounding} (L-3 ability). Here, the model locates objects by inferring based on complex conditions in the image. 
 \item  \textbf{Spatiotemporal Reasoning.} The \textit{Regional Counting with Change Detection} is unique in the RS field, as it evaluates changes in the number of objects between two temporal images of the same location.
\end{itemize}

\subsection{Data Collection and Quality Control}
\textbf{Data Source}
To create a benchmark with diverse evaluation dimensions, we collected large-size high-resolution RS images rich in visual details, in order to design varied and challenging assessment tasks. Specifically, we compiled 1,400 images from realistic RS scenarios for different downstream tasks, selecting them rigorously based on diversity and quality. For detection tasks, we sourced 270 images at 4,096$\times$4,096 and 210 images at 7,360$\times$4,912 from DOTA-v2~\cite{dota}, and added 50 images at a size of 3,744$\times$5,616 from the ITCVD~\cite{itcvd} dataset. For segmentation tasks, we used 457 images at 10,000$\times$10,000 resolution from MiniFrance~\cite{minifrance}, 13 images at 11,500$\times$7,500 from Toronto~\cite{toronto}, and 30 images at 6,000$\times$6,000 from Potsdam~\cite{potsdam}. Additionally, for change detection tasks, we included 185 pairs (370 images) at 10,000$\times$10,000 resolution from the HRSCD~\cite{hrscd} dataset.
Each image underwent multi-round cross-validation by annotators to ensure a rich, diverse selection of large-size and ultra-high-resolution samples. To be noticed, our dataset includes 840 images at 10,000$\times$10,000 pixels—a significant milestone in benchmark works. To the best of our knowledge, this is the first large-scale evaluation using 10,000$\times$10,000 images. Although challenging for human evaluation, we believe it is an essential step toward realistic RS applications.

\textbf{Visual Question Answering}
In XLRS-Bench, we compile visual language question-answer pairs in a multiple-choice format for each L-3 capability, excluding grounding and captioning dimensions. Each pair $( P_i = (Q_i, C_i, I_i, T_i) )$ includes: $ Q_i $ (the question), $ C_i $ (a set of $n$ options, $2 \leq n \leq 4)$, $I_i$ (the relevant image), and $ T_i $ (the correct answer).
It is important to note that research indicates GPT-based tools can introduce bias in benchmarks, favoring similar models and compromising their rigor and real-world assessment capacity~\cite{mmerealworld}. To address this, trained annotators expand the question set through human-only annotation, avoiding the assistance of GPTs.
To minimize design bias, we implemented cross-validation and formed an external review team. Specifically, three annotation groups, each with 15 members, cross-validated each other's work, while a review team of 10 experts with MLLM experience resolved discrepancies and validated question design, ensuring that all questions genuinely assess MLLM capabilities.

\textbf{Visual Grounding}
In XLRS-Bench, we evaluated visual grounding across two L-3 dimensions. We analyze each image by selecting 5-10 objects per L-3 dimension, resulting in 10-20 objects per image. Concretely, we carefully crafted reference sentences for each object to enable independent and precise identification. Distinctive features in these sentences such as color, shape, position, size, and relative spatial attributes distinguished each reference object from others. For the \textit{Condition-based Visual Grounding} task, annotators used detailed image information to create specific descriptions of conditions, including possible constraints on state or geographic details. Notably, all visual grounding annotations were performed by the annotation and quality control team of the VQA group, ensuring high-quality manual annotation throughout.

\begin{figure}[tbp]

\centering
\includegraphics[width=\linewidth]{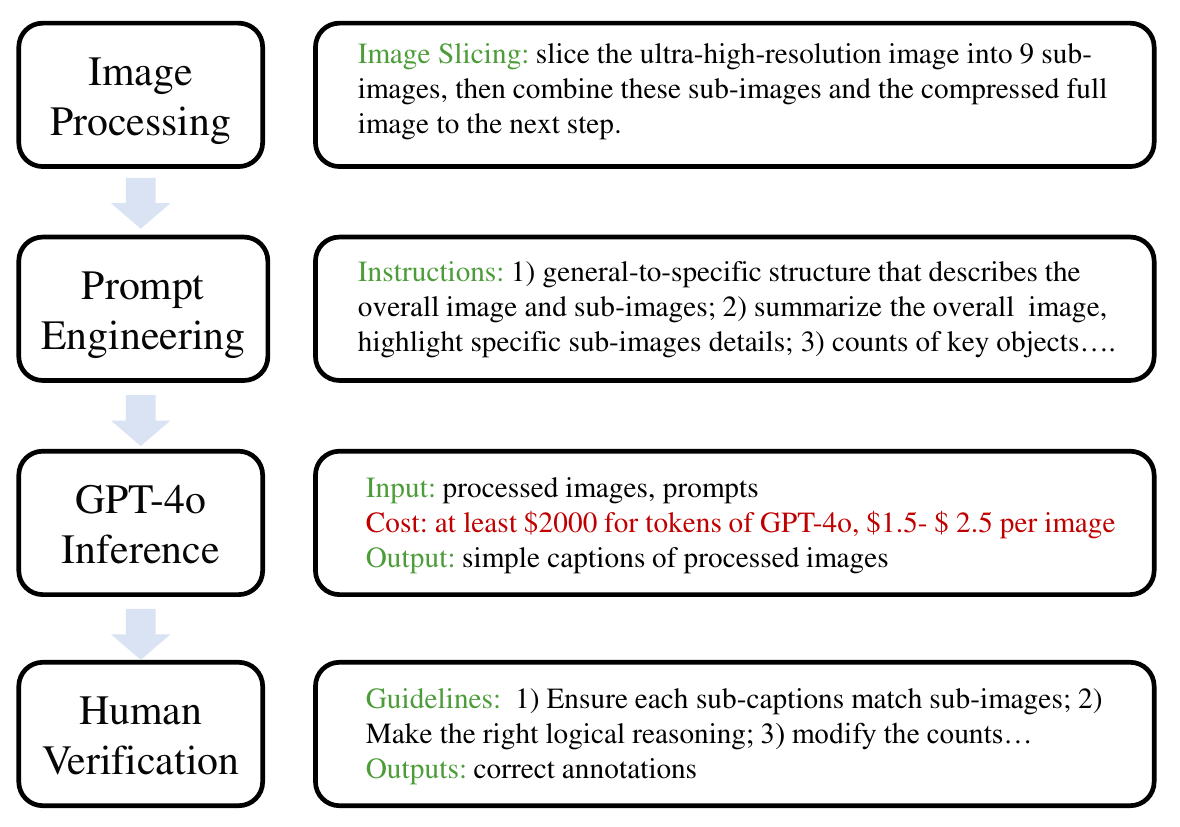}
\caption{Semi-automated pipeline for detailed image captioning in XLRS-Bench.}
\label{fig:caption}
\vspace{-4mm}
\end{figure}

\textbf{Image Caption}
Unlike other approaches that assess MLLM’s understanding of RS images with single, simple descriptions, we believe the rich details in ultra-high-resolution images require thorough and precise descriptions. For captioning, we use a semi-automated pipeline to interpret and generate text annotations of the image, as shown in Figure \ref{fig:caption}. 
We begin by dividing each image into nine sub-images, as using GPT-based tools to annotate the entire image often results in low-quality captions that lose important details, whereas the sub-image approach helps mitigate this issue.

In the next steps, we input 10 images (nine sub-images plus a compressed full image) into GPT-4o using complex prompts to generate detailed and comprehensive captions. 
For instance, for a 10,000 $\times$ 10,000 image where a river spans the scene with a wetland on one side and urban areas on the other, using sub-images alone cannot reflect the river's flow and overall layout, while only adopting the full image lacks local details. Our combined input method addresses both problems.
For the ultra-high-resolution RS scenarios, we adopt a general-to-specific structure that separately describes the overall image and each sub-image, where the key objects in the description of sub-images (like vehicles, boats, and buildings) are counted. Then we require the GPT-4o to count and display prominent objects across the entire image, identify anomalies, and predict potential risks or developments.
Detailed captions and multi-image inputs resulted in extensive token usage with GPT-4o, costing over \$2,000 for captioning 1,000 images. 
However, despite precise prompts, GPT-4 still struggles with complex tasks such as counting and reasoning, necessitating substantial human intervention to ensure the quality of the captions. To address this, annotators refine the captions using existing VQA annotations for detail and quality control. These refinements include adjusting summaries, correcting object counts, refining scene types, ordering regional descriptions by prominence, and manually annotating anomalies.

\begin{table*}[htbp]
\footnotesize
\caption{
Experimental results on the perception and reasoning dimensions of VQA tasks, with models ranked by average performance. Proprietary models are highlighted in gray. 'Avg' represents the average accuracy across sub-tasks.
}
\label{tab:vqa}
\centering
\resizebox{0.98\textwidth}{!}{%
\begin{tabular}{llc|cccc|cccc|c}
\toprule \Gray
\multicolumn{1}{c}{\textbf{Method}} & \multicolumn{1}{c}{\textbf{LLM}}& \multicolumn{1}{c}{\textbf{Language}} & \multicolumn{4}{c}{\textbf{Perception}}  & \multicolumn{4}{c}{\textbf{Reasoning}} &  \\  \midrule
\multicolumn{3}{c}{\multirow{2}{*}{\textbf{Sub-tasks (L-2 Capability)}}} & \multicolumn{1}{c}{\multirow{2}{*}{\textbf{Counting}}} & \textbf{Scene} & \textbf{Object} & \textbf{Object}  & \textbf{Complex} & \multicolumn{1}{c}{\multirow{2}{*}{\textbf{Planning}}} & \textbf{Spatiotemporal} & \textbf{Anomaly} &\multicolumn{1}{c}{\multirow{2}{*}{\textbf{Avg.}}}\\  
 & && & \textbf{Classification} & \textbf{Spatial Relationship} & \textbf{Properties}  & \textbf{Reasoning} &  & \textbf{Reasoning} & \textbf{Reasoning} \\ \midrule

CogVLM2 & Llama3-8B & en & 36.66 & 46.26 & \textbf{35.92} & \textbf{36.89} & 60.23 & 34.16 & - & 69.85 & \textbf{39.80} \\
Qwen2-VL & Qwen2-7B & en & \textbf{39.72} & 48.30 & 32.04 & 35.07 & 64.23 & 32.12 & \textbf{45.93} & 68.35 & 38.99 \\
LLaVA-OneVision & Qwen2-7B & en & 35.84 & 47.76 & 32.60 & 35.95 & 60.37 & 24.07 & 37.78 & 71.88 & 38.90 \\
\Lgray GPT-4o-mini & - & en & 28.32 & 45.46 & 29.96 & 36.29 & 54.84 & 33.81 & 15.19 & 72.06 & 37.99 \\
InternVL2 & InternLM2.5-7B & en & 33.91 & 44.93 & 26.60 & 34.84 & \textbf{66.43} & 33.01 & 44.44 & \textbf{74.54} & 37.89 \\
InternLM-XComposer-2.5 & InternLM2-7B & en & 35.84 & 47.43 & 32.75 & 30.60 & 64.76 & 35.31 & 32.22 & 69.50 & 36.33 \\
LLaVA-Next & Llama3-8B & en & 38.00 & 40.79 & 32.50 & 31.22 & 61.85 & 26.02 & 32.22 & 69.10 & 35.65 \\
\Lgray GPT-4o & - & en & 29.51 & \textbf{48.55} & 32.35 & 24.78 & 52.60 & \textbf{41.24} & 21.85 & 72.06 & 32.15 \\
LLaVA-1.5 & Vicuna-7B & en & 22.65 & 19.69 & 23.05 & 20.21 & 33.86 & 38.67 & 29.26 & 34.70 & 22.81 \\
GeoChat & Vicuna-7B & en & 22.65 & 18.96 & 23.30 & 20.21 & 30.33 & 32.23 & - & 33.25 & 22.03 \\

\midrule
Qwen2-VL & Qwen2-7B & zh & \textbf{39.4}9 & \textbf{49.28} & 33.26 & \textbf{37.89} & 67.62 & 24.34 & \textbf{44.44} & \textbf{76.57} & \textbf{41.10} \\
InternVL2 & InternLM2.5-7B & zh & 34.42 & 39.12 & 34.24 & 37.54 & \textbf{68.57} & 40.09 & 44.07 & 76.57 & 40.58 \\
LLaVA-OneVision & Qwen2-7B & zh & 37.48 & 47.06 & 31.73 & 34.72 & 62.56 & 29.12 & 30.37 & 74.80 & 38.42 \\
\Lgray GPT-4o-mini & - & zh & 29.51 & 44.13 & 29.59 & 35.41 & 55.36 & \textbf{41.86} & 21.85 & 74.71 & 37.82 \\
InternLM-XComposer-2.5 & InternLM2-7B & zh & 37.70 & 43.15 & 32.62 & 33.40 & 62.99 & 29.47 & 24.44 & 69.14 & 37.26 \\
CogVLM2 & Llama3-8B & zh & 36.21 & 45.79 & \textbf{34.57} & 31.03 & 59.56 & 26.19 & - & 70.47 & 35.84 \\
LLaVA-Next & Llama3-8B & zh & 33.08 & 39.52 & 31.98 & 29.34 & 54.07 & 21.59 & 25.56 & 69.85 & 33.48 \\
\Lgray GPT-4o & - & zh & 22.28 & 45.47 & 31.25 & 24.93 & 45.45 & 27.08 & 15.19 & 69.41 & 30.40 \\
LLaVA-1.5 & Vicuna-7B & zh & 22.65 & 19.07 & 23.01 & 20.21 & 33.95 & 38.67 & 29.26 & 37.58 & 22.86 \\
GeoChat & Vicuna-7B & zh & 22.65 & 19.17 & 23.05 & 20.21 & 24.75 & 22.79 & - & 23.74 & 20.99 \\

 \bottomrule
\end{tabular}%
}
\vspace{-0.4cm}
\end{table*}
\subsection{Analysis}
\textbf{Resolution.}
There are some benchmarks that adopt the image with sizes up to 2,000 $\times$ 1,500 pixels, but they are primarily designed for natural images rather than RS scenes. In contrast, current RS benchmarks typically remain below 512 $\times$ 512 pixels, as shown in Table \ref{tab:main}. In contrast, our XLRS-Bench, with an average size of 8,500 $\times$ 8,500 pixels, provides extensive views that facilitate a deeper understanding of real RS scenarios. Remarkably, it includes 840 images at 10,000$\times$ 10,000 pixels, producing over 10,000 QA pairs. In fact, even annotators must carefully analyze these images to answer questions, with cross-validation employed to reduce errors. We hope this large-size RS benchmark is able to drive the development of MLLMs.

\textbf{Length of Caption.}
XLRS-Bench offers comprehensive descriptions that cover global image context, regional details, specific object attributes, object counts, anomalies, and reasoning information. Each caption starts with a general image overview, followed by precise details across various regions. Regional details include object attributes such as count, color, shape, size, and spatial positioning—both absolute values within the image and relative to other objects like buildings, roads, and trees. In summary, these descriptions emphasize clear, distinct features, avoiding ambiguity, and typically average 19 sentences (379 words) in English and 20 sentences (663 words) in Chinese.

\textbf{Bilingual.}
Traditional VQA methods~\cite{mmbench} use translation engines to convert QA pairs from English to Chinese, but this often leads to visual-text misalignment~\cite{mtvqa} and fails to handle subtleties, contextual nuances, language bias, and diverse question types. These issues are particularly evident in detailed captioning, where long content makes machine translation prone to errors. To build a high-quality Chinese benchmark, we engaged six bilingual experts fluent in Chinese and English to translate and cross-check all long-text content.
Finally, 1,400 images, 32,389 QA pairs, 12,619 groundings, and 934 detailed captions are served for the Chinese version of XLRS-Bench, maintaining similar task types, image quality, and difficulty.

\begin{table}[tbp]
\footnotesize
    \caption{Detailed image captioning performance on XLRS-Bench.}
    \centering
    \resizebox{0.45\textwidth}{!}{
    \begin{tabular}{lc|cccccc}
        \toprule \Gray
        \textbf{Method} &\textbf{Language} & \textbf{BLEU-1} & \textbf{BLEU-2} &\textbf{ BLEU-3 }& \textbf{BLEU-4} & \textbf{METEOR} & \textbf{ROUGE\_L} \\
        \hline
GeoChat & en & 16.74 & 8.38 & 4.49 & 2.45 & 10.37 & 16.72 \\
\Lgray GPT-4o & en & 34.69 & 17.67 & 8.56 & 4.04 & 23.54 & 20.93 \\
\Lgray GPT-4o-mini & en & 38.29 & 19.75 & 9.76 & 4.29 & \textbf{23.94} & \textbf{21.30} \\
Qwen2-VL & en & 26.74 & 12.79 & 5.99 & 2.53 & 19.32 & 19.76 \\
LLaVA-OneVision & en & \textbf{41.12} & \textbf{20.42} & \textbf{9.94} & \textbf{4.56} & 19.99 & 21.03 \\
LLaVA-Next & en & 27.62 & 13.45 & 6.82 & 3.52 & 17.78 & 20.65 \\
LLaVA-1.5 & en & 35.82 & 17.62 & 8.92 & 4.33 & 16.49 & 20.80 \\
CogVLM2 & en & 30.27 & 14.46 & 6.80 & 3.09 & 19.37 & 19.17 \\
InternLM-XComposer-2.5 & en & 35.17 & 15.91 & 7.00 & 3.02 & 19.99 & 17.95 \\
InternVL2 & en & 25.71 & 12.44 & 5.84 & 2.58 & 19.55 & 19.43 \\
\midrule
GeoChat & zh & 6.77 & 1.49 & 0.68 & 0.26 & 7.84 & 15.79 \\
\Lgray GPT-4o & zh & 31.08 & 3.86 & 1.43 & 0.43 & \textbf{26.41} & 36.41 \\
\Lgray GPT-4o-mini & zh & 34.13 & 5.37 & 2.20 & 0.58 & 25.73 & \textbf{37.11} \\
Qwen2-VL & zh & 21.80 & 3.50 & 1.41 & 0.33 & 22.92 & 31.04 \\
LLaVA-OneVision & zh & 33.05 & 5.67 & \textbf{2.47} & \textbf{0.98} & 20.24 & 31.95 \\
LLaVA-Next & zh & 13.01 & 2.10 & 0.82 & 0.20 & 15.12 & 27.72 \\
LLaVA-1.5 & zh & 28.56 & 4.26 & 1.73 & 0.00 & 16.36 & 29.18 \\
CogVLM2 & zh & 19.78 & 2.23 & 0.79 & 0.18 & 22.53 & 28.33 \\
InternLM-XComposer-2.5 & zh & \textbf{37.30} & \textbf{6.12} & 2.39 & 0.58 & 20.86 & 32.97 \\
InternVL2 & zh & 16.49 & 3.16 & 1.39 & 0.48 & 22.10 & 25.76 \\

        \bottomrule
    \end{tabular}
    }
    \label{tab_cap}
    \vspace{-5mm}
\end{table}

\section{Experiment}
\label{sec:experiment}

\subsection{Experimental Setup}
The MLLMs evaluated on XLRS-Bench are grouped into three categories: 
(a) open-source VLMs, including Qwen2-VL~\cite{qwen2}, LLava-Onevision~\cite{llava-onevision}, LLava-Next~\cite{llava-next}, LLaVA-1.5~\cite{llama}, CogVLM2~\cite{cogvlm2}, InternLM-XComposer-2.5~\cite{ixc2.5} and InternVL-2~\citep{internvl2};
(b) closed-source VLMs, such as GPT-4o~\cite{gpt4o}
and GPT-4o-mini~\cite{gpt4omini}; and (c) the specialized RS model Geochat~\cite{geochat}. For fair comparison, we used a zero-shot setting with uniform prompts for all VLMs. The appendix details the architecture and parameter sizes of each open-source VLM and includes additional results across various settings. Except for GeoChat which was evaluated using its native framework, all other models were evaluated using LMMs-Eval~\cite{zhang2024lmmsevalrealitycheckevaluation,lmms_eval2024}. More results are shown in the Appendix.
\begin{table*}[t!]
\footnotesize
 \vspace{-5mm}
    \caption{Visual grounding performance on XLRS-Bench and VRSBench~\cite{vrsbench}. *: VRSBench uses GPT-4V to assess this task.}
    \centering
    \resizebox{0.98\textwidth}{!}{
     \begin{tabular}{l|ccccccccccccc}
        \toprule \Gray
        \textbf{Benchmark} & \textbf{Method}  & \textbf{GPT-4o}* & \textbf{GPT-4o-mini} & \textbf{Qwen2-VL} & \textbf{LLaVA-OneVision} & \textbf{LLaVA-Next} & \textbf{LLaVA-1.5} & \textbf{CogVLM2} & \textbf{InternLM-XComposer-2.5} & \textbf{InternVL2} & \textbf{GeoChat} \\
        \hline
       \multirow{2}{*}{XLRS-Bench-EN}&Acc@0.5  &  \textbf{0.46} & 0.09 & 0.15 & 0.16 & 0.18 & 0.09 & 0.01 & 0.02 & 0.33 & 0.14 \\
                                    &Acc@0.7  &  0.05 & 0.03 & 0.03 & 0.00 & 0.04 & 0.00 & 0.00 & 0.01 & \textbf{0.12} & 0.01 \\ \midrule
      \multirow{2}{*}{XLRS-Bench-ZH}&Acc@0.5 &  \textbf{0.45} & 0.21 & 0.14 & 0.13 & 0.07 & 0.12 & 0.03 & 0.06 & 0.19 & 0.14 \\
                                    &Acc@0.7  & 0.03 & 0.03 & 0.01 & 0.01 & 0.02 & 0.02 & 0.00 & 0.00 & \textbf{0.06} & 0.01 \\
 \midrule
         \multirow{2}{*}{VRSBench}&   Acc@0.5 & 5.1 & -  & - & - & - & - & - & - & - & - \\ 
        &Acc@0.7  & 1.1 & -  & - & -& - & - & - & - & - & - \\
        
        \bottomrule
    \end{tabular}
    }

    \label{tab_ref}
    \vspace{-7mm}
    
\end{table*}
\subsection{Evaluation Strategy}
In the VQA task, we manually created four options for each question: one correct answer and three distractors, derived from text within the image or similar alternatives. This increases the difficulty by requiring models to deeply understand image details. Following MMBench~\cite{mmbench} and MME-Realworld~\cite{mmerealworld} methods, we evaluated the accuracy of L-2 capability dimension for the VQA task and reported the average accuracy across L-2 dimensions, with L-3 results available in the appendix.
For the L3 sub-task of Overall Land Use Classification, where a question may have multiple correct options, a prediction is considered correct only if it matches the ground truth exactly.
For the Grounding task, we used precision, assessing accuracy based on the intersection between predicted and ground truth bounding boxes, with predictions deemed correct if IoU exceeds a threshold. We tested two IoU thresholds: 0.5 and 0.7. For the captioning task, we used standard metrics, including BLEU~\cite{bleu}, ROUGE\_L~\cite{rouge}, and METEOR~\cite{meteor}. We consider the n-gram precision for BLEU with n-values of 1, 2, 3, and 4. All scores in Tables \ref{tab:vqa}, \ref{tab_cap}, and \ref{tab_ref} are reported as percentages (\%).

\subsection{Main Results}
\textbf{Results on VQA tasks.}
Table \ref{tab:vqa} compares various models on the different four L-2 dimensions of perception and reasoning abilities, respectively. Qwen2-VL excels in both English and Chinese proficiency, outperforming both proprietary and most open-source models. Nevertheless, their performance varies across tasks. From the results, we can draw the following key insights:
(1) \textit{GPT-4o's Weakness in Spatiotemporal Reasoning}:
GPT-4o underperforms compared to open-source models on complex spatiotemporal reasoning tasks, with accuracy consistently below 25\%.
These tasks, focused on local counting for change detection, assess MLLMs' ability to capture temporal details across images. GPT-4o likely lacks pretraining on RS scene change detection and sometimes refuses to answer questions citing privacy or harmful content. By contrast, Qwen2-VL performs reliably on these sub-tasks.
(2) \textit{Benefits of Higher-Resolution Input Models}: Models like Qwen2-VL, which allow higher-resolution inputs, significantly outperform those using visual encoders like CLIP (\textit{e.g.}, LLaVA1.5). The limit of input image size often necessitates compression, but Qwen2-VL's ability to process richer information at larger image sizes leads to superior performance.
(3) \textit{Better Abstract Reasoning in RS}: In high-resolution RS scenarios, models excel at abstract reasoning tasks like anomaly detection, which require minimal local detail and can rely on compressed regional data. Conversely, for perception-heavy tasks requiring fine detail recognition, MLLMs are constrained by their architecture, achieving accuracy rates of only 30\%–50\%.

\textbf{Results on image captioning tasks.}
Our captions have average hl{379 words in English and 663} in Chinese, offering richly detailed descriptions with broad applicability. However, evaluations show MLLMs struggle with these lengthy, real-world captions; open-source models like Qwen2-VL and the RS-specific GeoChat perform poorly. Notably, GPT-4o and GPT-4o-mini, even without extensive RS pretraining, excel in long-text generation, significantly outperforming other models. Previous evaluations in RS captioning focused on shorter texts, missing GPT-4o’s strengths with longer content. 
For Chinese caption tasks, GPT-4o still outperforms other models, showcasing its robust understanding and generation abilities as a valuable asset for future research.

\textbf{Results on grounding tasks.}
While we anticipated challenges in visual localization during annotation, the poor performance of each MLLM was surprising. To validate the benchmark’s effectiveness, we conducted additional human verification: we randomly selected 100 questions from each VQA sub-task (L-3 dimensions) and had two groups answer them concurrently. We also included results from the low-resolution RS visual grounding dataset, VRSBench. Notably, even in 512$\times$512 low-resolution RS scenarios, GPT-4V achieved only 1.1\% and 5.1\% accuracy; when scaled to resolutions near 8,500$\times$8,500, the accuracy rates of 3.2\% and 0.48\% are understandable. Some objects in these large images are as small as 5–10 pixels, posing a challenge for current MLLMs. 

\textbf{Limitations and discussions.}
1) \textit{Limitations in Spatiotemporal Understanding}: General models (like LLaVA-series and GPT-4o) lack large-scale training tailored to RS spatiotemporal understandings, leading to poor performance on such tasks. For instance, GPT-4o achieves less than 25\% accuracy (Table \ref{tab:vqa}). Given the widespread application of these tasks in RS, domain-specific training is essential. Specialized models like GeoChat have yet to support multi-image input tasks, highlingting the need for further research in this area.
2) \textit{Challenges in Ultra-High-Resolution Scenarios}: Current high-resolution MLLMs are limited to 4K images, requiring significant compression for larger-size images, resulting in substantial information loss. This issue is critical for satellite imagery, where the small objects are often measured only 5–10 pixels and shrink to a single pixel after compression, rendering them uninformative. Developing specialized MLLMs for RS super-resolution is anticipated to mitigate this limitation.

\section{Conclusion}
\label{sec:conclusion}
In this paper, we introduce XLRS-Bench, a comprehensive benchmark for evaluating the perception and reasoning capabilities of multimodal large language models (MLLMs) in ultra-high-resolution remote sensing (RS) scenarios. XLRS-Bench features the largest average image size to date, high-quality human-verified annotations, and 16 sub-tasks across three ability levels, offering a multidimensional evaluation. It supports both English and Chinese, making it the largest manually annotated ultra-high-resolution RS vision-language dataset, surpassing existing benchmarks in data capacity and task diversity. By emphasizing real-world decision-making and spatiotemporal change detection, XLRS-Bench promotes advanced cognitive processes. Experimental results demonstrate that current general-purpose and RS-specific MLLMs still struggle to understand ultra-high-resolution RS imagery, underscoring the need for further improvements.

\section{Acknowledgements}
This work was partially supported by the National Natural Science Foundation of China (No. 62372459, No.62376282 and No. 624B2109)

{
    \small
    \bibliographystyle{ieeenat_fullname}
    \bibliography{main}

\begin{thebibliography}{81}
\providecommand{\natexlab}[1]{#1}
\providecommand{\url}[1]{\texttt{#1}}
\expandafter\ifx\csname urlstyle\endcsname\relax
  \providecommand{\doi}[1]{doi: #1}\else
  \providecommand{\doi}{doi: \begingroup \urlstyle{rm}\Url}\fi

\bibitem[Agrawal et~al.(2019)Agrawal, Desai, Wang, Chen, Jain, Johnson, Batra, Parikh, Lee, and Anderson]{nocaps}
Harsh Agrawal, Karan Desai, Yufei Wang, Xinlei Chen, Rishabh Jain, Mark Johnson, Dhruv Batra, Devi Parikh, Stefan Lee, and Peter Anderson.
\newblock Nocaps: Novel object captioning at scale.
\newblock In \emph{Proceedings of the IEEE/CVF international conference on computer vision}, pages 8948--8957, 2019.

\bibitem[Bai et~al.(2023{\natexlab{a}})Bai, Bai, Yang, Wang, Tan, Wang, Lin, Zhou, and Zhou]{qwen}
Jinze Bai, Shuai Bai, Shusheng Yang, Shijie Wang, Sinan Tan, Peng Wang, Junyang Lin, Chang Zhou, and Jingren Zhou.
\newblock Qwen-vl: A frontier large vision-language model with versatile abilities.
\newblock \emph{arXiv preprint arXiv:2308.12966}, 2023{\natexlab{a}}.

\bibitem[Bai et~al.(2023{\natexlab{b}})Bai, Yang, Bai, Wang, Zhang, Lin, Wang, Zhou, and Zhou]{touchstone}
Shuai Bai, Shusheng Yang, Jinze Bai, Peng Wang, Xingxuan Zhang, Junyang Lin, Xinggang Wang, Chang Zhou, and Jingren Zhou.
\newblock Touchstone: Evaluating vision-language models by language models.
\newblock \emph{arXiv preprint arXiv:2308.16890}, 2023{\natexlab{b}}.

\bibitem[Banerjee and Lavie(2005)]{meteor}
Satanjeev Banerjee and Alon Lavie.
\newblock Meteor: An automatic metric for mt evaluation with improved correlation with human judgments.
\newblock In \emph{Proceedings of the acl workshop on intrinsic and extrinsic evaluation measures for machine translation and/or summarization}, pages 65--72, 2005.

\bibitem[Bitton et~al.(2023)Bitton, Bansal, Hessel, Shao, Zhu, Awadalla, Gardner, Taori, and Schimdt]{visitbench}
Yonatan Bitton, Hritik Bansal, Jack Hessel, Rulin Shao, Wanrong Zhu, Anas Awadalla, Josh Gardner, Rohan Taori, and Ludwig Schimdt.
\newblock Visit-bench: a benchmark for vision-language instruction following inspired by real-world use.
\newblock In \emph{Proceedings of the 37th International Conference on Neural Information Processing Systems}, pages 26898--26922, 2023.

\bibitem[Castillo-Navarro et~al.(2022)Castillo-Navarro, Le~Saux, Boulch, Audebert, and Lef{\`e}vre]{minifrance}
Javiera Castillo-Navarro, Bertrand Le~Saux, Alexandre Boulch, Nicolas Audebert, and S{\'e}bastien Lef{\`e}vre.
\newblock Semi-supervised semantic segmentation in earth observation: The minifrance suite, dataset analysis and multi-task network study.
\newblock \emph{Machine Learning}, 111\penalty0 (9):\penalty0 3125--3160, 2022.

\bibitem[Chen et~al.(2024)Chen, Li, Dong, Zhang, Zang, Chen, Duan, Wang, Qiao, Lin, et~al.]{mmstar}
Lin Chen, Jinsong Li, Xiaoyi Dong, Pan Zhang, Yuhang Zang, Zehui Chen, Haodong Duan, Jiaqi Wang, Yu Qiao, Dahua Lin, et~al.
\newblock Are we on the right way for evaluating large vision-language models?
\newblock \emph{arXiv preprint arXiv:2403.20330}, 2024.

\bibitem[Chen et~al.(2015)Chen, Fang, Lin, Vedantam, Gupta, Doll{\'a}r, and Zitnick]{coco_caption}
Xinlei Chen, Hao Fang, Tsung-Yi Lin, Ramakrishna Vedantam, Saurabh Gupta, Piotr Doll{\'a}r, and C~Lawrence Zitnick.
\newblock Microsoft coco captions: Data collection and evaluation server.
\newblock \emph{arXiv preprint arXiv:1504.00325}, 2015.

\bibitem[Chin-Yew(2004)]{rouge}
Lin Chin-Yew.
\newblock Rouge: A package for automatic evaluation of summaries.
\newblock In \emph{Proceedings of the Workshop on Text Summarization Branches Out, 2004}, 2004.

\bibitem[Daudt et~al.(2019)Daudt, Le~Saux, Boulch, and Gousseau]{hrscd}
Rodrigo~Caye Daudt, Bertrand Le~Saux, Alexandre Boulch, and Yann Gousseau.
\newblock Multitask learning for large-scale semantic change detection.
\newblock \emph{Computer Vision and Image Understanding}, 187:\penalty0 102783, 2019.

\bibitem[Dell'Acqua and Gamba(2012)]{disa_ass}
Fabio Dell'Acqua and Paolo Gamba.
\newblock Remote sensing and earthquake damage assessment: Experiences, limits, and perspectives.
\newblock \emph{Proceedings of the IEEE}, 100\penalty0 (10):\penalty0 2876--2890, 2012.

\bibitem[Fu et~al.(2023)Fu, Chen, Shen, Qin, Zhang, Lin, Yang, Zheng, Li, Sun, et~al.]{mme}
Chaoyou Fu, Peixian Chen, Yunhang Shen, Yulei Qin, Mengdan Zhang, Xu Lin, Jinrui Yang, Xiawu Zheng, Ke Li, Xing Sun, et~al.
\newblock Mme: A comprehensive evaluation benchmark for multimodal large language models.
\newblock \emph{arXiv preprint arXiv:2306.13394}, 2023.

\bibitem[Fu et~al.(2024)Fu, Hu, Li, Feng, Wang, Lin, Roth, Smith, Ma, and Krishna]{blink}
Xingyu Fu, Yushi Hu, Bangzheng Li, Yu Feng, Haoyu Wang, Xudong Lin, Dan Roth, Noah~A Smith, Wei-Chiu Ma, and Ranjay Krishna.
\newblock Blink: Multimodal large language models can see but not perceive.
\newblock In \emph{European Conference on Computer Vision}, pages 148--166. Springer, 2024.

\bibitem[Gebru et~al.(2021)Gebru, Morgenstern, Vecchione, Vaughan, Wallach, Iii, and Crawford]{datasheets}
Timnit Gebru, Jamie Morgenstern, Briana Vecchione, Jennifer~Wortman Vaughan, Hanna Wallach, Hal~Daum{\'e} Iii, and Kate Crawford.
\newblock Datasheets for datasets.
\newblock \emph{Communications of the ACM}, 64\penalty0 (12):\penalty0 86--92, 2021.

\bibitem[Goyal et~al.(2017)Goyal, Khot, Summers-Stay, Batra, and Parikh]{vqav2}
Yash Goyal, Tejas Khot, Douglas Summers-Stay, Dhruv Batra, and Devi Parikh.
\newblock Making the v in vqa matter: Elevating the role of image understanding in visual question answering.
\newblock In \emph{Proceedings of the IEEE conference on computer vision and pattern recognition}, pages 6904--6913, 2017.

\bibitem[Guo et~al.(2024)Guo, Xu, Yao, Cui, Ni, Ge, Chua, Liu, and Huang]{llava-uhd}
Zonghao Guo, Ruyi Xu, Yuan Yao, Junbo Cui, Zanlin Ni, Chunjiang Ge, Tat-Seng Chua, Zhiyuan Liu, and Gao Huang.
\newblock Llava-uhd: an lmm perceiving any aspect ratio and high-resolution images.
\newblock In \emph{European Conference on Computer Vision}, pages 390--406. Springer, 2024.

\bibitem[Gurari et~al.(2018)Gurari, Li, Stangl, Guo, Lin, Grauman, Luo, and Bigham]{vizwiz}
Danna Gurari, Qing Li, Abigale~J Stangl, Anhong Guo, Chi Lin, Kristen Grauman, Jiebo Luo, and Jeffrey~P Bigham.
\newblock Vizwiz grand challenge: Answering visual questions from blind people.
\newblock In \emph{Proceedings of the IEEE conference on computer vision and pattern recognition}, pages 3608--3617, 2018.

\bibitem[Hong et~al.(2024)Hong, Wang, Ding, Yu, Lv, Wang, Cheng, Huang, Ji, Xue, et~al.]{cogvlm2}
Wenyi Hong, Weihan Wang, Ming Ding, Wenmeng Yu, Qingsong Lv, Yan Wang, Yean Cheng, Shiyu Huang, Junhui Ji, Zhao Xue, et~al.
\newblock Cogvlm2: Visual language models for image and video understanding.
\newblock \emph{arXiv preprint arXiv:2408.16500}, 2024.

\bibitem[Hu et~al.(2024)Hu, Tu, Han, He, Cui, Long, Zheng, Fang, Huang, Zhao, et~al.]{minicpm}
Shengding Hu, Yuge Tu, Xu Han, Chaoqun He, Ganqu Cui, Xiang Long, Zhi Zheng, Yewei Fang, Yuxiang Huang, Weilin Zhao, et~al.
\newblock Minicpm: Unveiling the potential of small language models with scalable training strategies.
\newblock \emph{arXiv preprint arXiv:2404.06395}, 2024.

\bibitem[Hu et~al.(2023)Hu, Yuan, Wen, Lu, and Li]{rsgpt}
Yuan Hu, Jianlong Yuan, Congcong Wen, Xiaonan Lu, and Xiang Li.
\newblock Rsgpt: A remote sensing vision language model and benchmark.
\newblock \emph{arXiv preprint arXiv:2307.15266}, 2023.

\bibitem[Huang et~al.(2022)Huang, Wang, Zhang, and Xu]{ship_my}
Liang Huang, Fengxiang Wang, Yalun Zhang, and Qingxia Xu.
\newblock Fine-grained ship classification by combining cnn and swin transformer.
\newblock \emph{Remote Sensing}, 14\penalty0 (13):\penalty0 3087, 2022.

\bibitem[Hudson and Manning(2019)]{gqa}
Drew~A Hudson and Christopher~D Manning.
\newblock Gqa: A new dataset for real-world visual reasoning and compositional question answering.
\newblock In \emph{Proceedings of the IEEE/CVF conference on computer vision and pattern recognition}, pages 6700--6709, 2019.

\bibitem[Kuckreja et~al.(2024)Kuckreja, Danish, Naseer, Das, Khan, and Khan]{geochat}
Kartik Kuckreja, Muhammad~Sohail Danish, Muzammal Naseer, Abhijit Das, Salman Khan, and Fahad~Shahbaz Khan.
\newblock Geochat: Grounded large vision-language model for remote sensing.
\newblock In \emph{Proceedings of the IEEE/CVF Conference on Computer Vision and Pattern Recognition}, pages 27831--27840, 2024.

\bibitem[Lan et~al.(2024)Lan, Wang, Zheng, Wang, and Liu]{efficient}
Long Lan, Fengxiang Wang, Xiangtao Zheng, Zengmao Wang, and Xinwang Liu.
\newblock Efficient prompt tuning of large vision-language model for fine-grained ship classification.
\newblock \emph{IEEE Transactions on Geoscience and Remote Sensing}, 2024.

\bibitem[Li et~al.(2023)Li, Wang, Wang, Ge, Ge, and Shan]{seedbench}
Bohao Li, Rui Wang, Guangzhi Wang, Yuying Ge, Yixiao Ge, and Ying Shan.
\newblock Seed-bench: Benchmarking multimodal llms with generative comprehension.
\newblock \emph{arXiv preprint arXiv:2307.16125}, 2023.

\bibitem[Li et~al.(2024{\natexlab{a}})Li, Zhang, Zhang, Pu, Du, Dong, Liu, Zhang, Zhang, Li, and Liu]{lmms_eval2024}
Bo Li, Peiyuan Zhang, Kaichen Zhang, Fanyi Pu, Xinrun Du, Yuhao Dong, Haotian Liu, Yuanhan Zhang, Ge Zhang, Chunyuan Li, and Ziwei Liu.
\newblock Lmms-eval: Accelerating the development of large multimodal models.
\newblock \url{https://github.com/EvolvingLMMs-Lab/lmms-eval}, 2024{\natexlab{a}}.

\bibitem[Li et~al.(2024{\natexlab{b}})Li, Zhang, Guo, Zhang, Li, Zhang, Zhang, Li, Liu, and Li]{llava-onevision}
Bo Li, Yuanhan Zhang, Dong Guo, Renrui Zhang, Feng Li, Hao Zhang, Kaichen Zhang, Yanwei Li, Ziwei Liu, and Chunyuan Li.
\newblock Llava-onevision: Easy visual task transfer.
\newblock \emph{arXiv preprint arXiv:2408.03326}, 2024{\natexlab{b}}.

\bibitem[Li et~al.(2024{\natexlab{c}})Li, Ding, and Elhoseiny]{vrsbench}
Xiang Li, Jian Ding, and Mohamed Elhoseiny.
\newblock Vrsbench: A versatile vision-language benchmark dataset for remote sensing image understanding.
\newblock \emph{arXiv preprint arXiv:2406.12384}, 2024{\natexlab{c}}.

\bibitem[Li et~al.(2024{\natexlab{d}})Li, Zhang, Wang, Zhong, Chen, Chu, Liu, and Jia]{mini-gemini}
Yanwei Li, Yuechen Zhang, Chengyao Wang, Zhisheng Zhong, Yixin Chen, Ruihang Chu, Shaoteng Liu, and Jiaya Jia.
\newblock Mini-gemini: Mining the potential of multi-modality vision language models.
\newblock \emph{arXiv preprint arXiv:2403.18814}, 2024{\natexlab{d}}.

\bibitem[Li et~al.(2024{\natexlab{e}})Li, Yang, Liu, Ma, Zhang, Yang, Sun, Liu, and Bai]{monkey}
Zhang Li, Biao Yang, Qiang Liu, Zhiyin Ma, Shuo Zhang, Jingxu Yang, Yabo Sun, Yuliang Liu, and Xiang Bai.
\newblock Monkey: Image resolution and text label are important things for large multi-modal models.
\newblock In \emph{proceedings of the IEEE/CVF conference on computer vision and pattern recognition}, pages 26763--26773, 2024{\natexlab{e}}.

\bibitem[Liu et~al.(2023)Liu, Li, Wu, and Lee]{visualllama}
Haotian Liu, Chunyuan Li, Qingyang Wu, and Yong~Jae Lee.
\newblock Visual instruction tuning.
\newblock \emph{Advances in neural information processing systems}, 36:\penalty0 34892--34916, 2023.

\bibitem[Liu et~al.(2024{\natexlab{a}})Liu, Li, Li, Li, Zhang, Shen, and Lee]{llava-next}
Haotian Liu, Chunyuan Li, Yuheng Li, Bo Li, Yuanhan Zhang, Sheng Shen, and Yong~Jae Lee.
\newblock Llava-next: Improved reasoning, ocr, and world knowledge.
\newblock \url{https://llava-vl.github.io/blog/2024-01-30-llava-nextl}, 2024{\natexlab{a}}.

\bibitem[Liu et~al.(2024{\natexlab{b}})Liu, Ma, Zhang, Wang, Ji, Sun, and Ji]{rrsisd}
Sihan Liu, Yiwei Ma, Xiaoqing Zhang, Haowei Wang, Jiayi Ji, Xiaoshuai Sun, and Rongrong Ji.
\newblock Rotated multi-scale interaction network for referring remote sensing image segmentation.
\newblock In \emph{Proceedings of the IEEE/CVF Conference on Computer Vision and Pattern Recognition}, 2024{\natexlab{b}}.

\bibitem[Liu et~al.(2024{\natexlab{c}})Liu, Duan, Zhang, Li, Zhang, Zhao, Yuan, Wang, He, Liu, et~al.]{mmbench}
Yuan Liu, Haodong Duan, Yuanhan Zhang, Bo Li, Songyang Zhang, Wangbo Zhao, Yike Yuan, Jiaqi Wang, Conghui He, Ziwei Liu, et~al.
\newblock Mmbench: Is your multi-modal model an all-around player?
\newblock In \emph{European conference on computer vision}, pages 216--233. Springer, 2024{\natexlab{c}}.

\bibitem[Lobry et~al.(2020)Lobry, Marcos, Murray, and Tuia]{rsvqahr}
Sylvain Lobry, Diego Marcos, Jesse Murray, and Devis Tuia.
\newblock Rsvqa: Visual question answering for remote sensing data.
\newblock \emph{IEEE Transactions on Geoscience and Remote Sensing}, 58\penalty0 (12):\penalty0 8555--8566, 2020.

\bibitem[Lu et~al.(2017)Lu, Wang, Zheng, and Li]{rsicd}
Xiaoqiang Lu, Binqiang Wang, Xiangtao Zheng, and Xuelong Li.
\newblock Exploring models and data for remote sensing image caption generation.
\newblock \emph{IEEE Transactions on Geoscience and Remote Sensing}, 56\penalty0 (4):\penalty0 2183--2195, 2017.

\bibitem[Luo et~al.(2024)Luo, Pang, Zhang, Wang, Wang, Dang, Lao, Wang, Chen, Tan, et~al.]{skysensegpt}
Junwei Luo, Zhen Pang, Yongjun Zhang, Tingzhu Wang, Linlin Wang, Bo Dang, Jiangwei Lao, Jian Wang, Jingdong Chen, Yihua Tan, et~al.
\newblock Skysensegpt: A fine-grained instruction tuning dataset and model for remote sensing vision-language understanding.
\newblock \emph{arXiv preprint arXiv:2406.10100}, 2024.

\bibitem[Marino et~al.(2019)Marino, Rastegari, Farhadi, and Mottaghi]{ok-vqa}
Kenneth Marino, Mohammad Rastegari, Ali Farhadi, and Roozbeh Mottaghi.
\newblock Ok-vqa: A visual question answering benchmark requiring external knowledge.
\newblock In \emph{Proceedings of the IEEE/cvf conference on computer vision and pattern recognition}, pages 3195--3204, 2019.

\bibitem[Marsocci et~al.(2024)Marsocci, Jia, Bellier, Kerekes, Zeng, Hafner, Gerard, Brune, Yadav, Shibli, et~al.]{VLEO-BENCH}
Valerio Marsocci, Yuru Jia, Georges~Le Bellier, David Kerekes, Liang Zeng, Sebastian Hafner, Sebastian Gerard, Eric Brune, Ritu Yadav, Ali Shibli, et~al.
\newblock Pangaea: A global and inclusive benchmark for geospatial foundation models.
\newblock \emph{arXiv preprint arXiv:2412.04204}, 2024.

\bibitem[Masry et~al.(2022)Masry, Long, Tan, Joty, and Hoque]{chartqa}
Ahmed Masry, Do~Xuan Long, Jia~Qing Tan, Shafiq Joty, and Enamul Hoque.
\newblock Chartqa: A benchmark for question answering about charts with visual and logical reasoning.
\newblock \emph{arXiv preprint arXiv:2203.10244}, 2022.

\bibitem[Muhtar et~al.(2024)Muhtar, Li, Gu, Zhang, and Xiao]{lhrsbot}
Dilxat Muhtar, Zhenshi Li, Feng Gu, Xueliang Zhang, and Pengfeng Xiao.
\newblock Lhrs-bot: Empowering remote sensing with vgi-enhanced large multimodal language model.
\newblock In \emph{European Conference on Computer Vision}, pages 440--457. Springer, 2024.

\bibitem[OpenAI(2024{\natexlab{a}})]{gpt4o}
OpenAI.
\newblock Hello gpt-4o.
\newblock \url{https://openai.com/index/hello-gpt-4o}, 2024{\natexlab{a}}.

\bibitem[OpenAI(2024{\natexlab{b}})]{gpt4omini}
OpenAI.
\newblock Gpt-4o mini: advancing cost-efficient intelligence.
\newblock \url{https://openai.com/index/gpt-4o-mini-advancing-cost-efficient-intelligence}, 2024{\natexlab{b}}.

\bibitem[{OpenAI} et~al.(2024){OpenAI}, Achiam, Adler, Agarwal, Ahmad, Akkaya, Aleman, Almeida, Altenschmidt, Altman, et~al.]{gpt4}
{OpenAI}, Josh Achiam, Steven Adler, Sandhini Agarwal, Lama Ahmad, Ilge Akkaya, Florencia~Leoni Aleman, Diogo Almeida, Janko Altenschmidt, Sam Altman, et~al.
\newblock Gpt-4 technical report.
\newblock \emph{arXiv preprint arXiv:2303.08774}, 2024.

\bibitem[Pang et~al.(2024)Pang, Wu, Li, Liu, Sun, Li, Weng, Wang, Feng, Xia, et~al.]{h2rsvlm}
Chao Pang, Jiang Wu, Jiayu Li, Yi Liu, Jiaxing Sun, Weijia Li, Xingxing Weng, Shuai Wang, Litong Feng, Gui-Song Xia, et~al.
\newblock H2rsvlm: Towards helpful and honest remote sensing large vision language model.
\newblock \emph{arXiv preprint arXiv:2403.20213}, 2024.

\bibitem[Papineni et~al.(2002)Papineni, Roukos, Ward, and Zhu]{bleu}
Kishore Papineni, Salim Roukos, Todd Ward, and Wei-Jing Zhu.
\newblock Bleu: a method for automatic evaluation of machine translation.
\newblock In \emph{Proceedings of the 40th annual meeting of the Association for Computational Linguistics}, pages 311--318, 2002.

\bibitem[Qu et~al.(2016)Qu, Li, Tao, and Lu]{ucmcaption}
Bo Qu, Xuelong Li, Dacheng Tao, and Xiaoqiang Lu.
\newblock Deep semantic understanding of high resolution remote sensing image.
\newblock In \emph{2016 International conference on computer, information and telecommunication systems (Cits)}, pages 1--5. IEEE, 2016.

\bibitem[Rottensteiner et~al.(2014{\natexlab{a}})Rottensteiner, Sohn, Gerke, and Wegner]{potsdam}
Franz Rottensteiner, Gunho Sohn, Markus Gerke, and Jan~D Wegner.
\newblock Isprs semantic labeling contest.
\newblock \emph{ISPRS: Leopoldsh{\"o}he, Germany}, 1\penalty0 (4):\penalty0 4, 2014{\natexlab{a}}.

\bibitem[Rottensteiner et~al.(2014{\natexlab{b}})Rottensteiner, Sohn, Gerke, Wegner, Breitkopf, and Jung]{toronto}
Franz Rottensteiner, Gunho Sohn, Markus Gerke, Jan~Dirk Wegner, Uwe Breitkopf, and Jaewook Jung.
\newblock Results of the isprs benchmark on urban object detection and 3d building reconstruction.
\newblock \emph{ISPRS journal of photogrammetry and remote sensing}, 93:\penalty0 256--271, 2014{\natexlab{b}}.

\bibitem[Saikh et~al.(2022)Saikh, Ghosal, Mittal, Ekbal, and Bhattacharyya]{scienceqa}
Tanik Saikh, Tirthankar Ghosal, Amish Mittal, Asif Ekbal, and Pushpak Bhattacharyya.
\newblock Scienceqa: A novel resource for question answering on scholarly articles.
\newblock \emph{International Journal on Digital Libraries}, 23\penalty0 (3):\penalty0 289--301, 2022.

\bibitem[Singh et~al.(2019{\natexlab{a}})Singh, Natarajan, Shah, Jiang, Chen, Batra, Parikh, and Rohrbach]{singh2019towards}
Amanpreet Singh, Vivek Natarajan, Meet Shah, Yu Jiang, Xinlei Chen, Dhruv Batra, Devi Parikh, and Marcus Rohrbach.
\newblock Towards vqa models that can read.
\newblock In \emph{CVPR}, 2019{\natexlab{a}}.

\bibitem[Singh et~al.(2019{\natexlab{b}})Singh, Natarajan, Shah, Jiang, Chen, Batra, Parikh, and Rohrbach]{textvqa}
Amanpreet Singh, Vivek Natarajan, Meet Shah, Yu Jiang, Xinlei Chen, Dhruv Batra, Devi Parikh, and Marcus Rohrbach.
\newblock Towards vqa models that can read.
\newblock In \emph{Proceedings of the IEEE/CVF conference on computer vision and pattern recognition}, pages 8317--8326, 2019{\natexlab{b}}.

\bibitem[Sun et~al.(2022)Sun, Feng, Li, Ye, Kang, and Huang]{rsvg}
Yuxi Sun, Shanshan Feng, Xutao Li, Yunming Ye, Jian Kang, and Xu Huang.
\newblock Visual grounding in remote sensing images.
\newblock In \emph{Proceedings of the 30th ACM International Conference on Multimedia}, pages 404--412, 2022.

\bibitem[Tang et~al.(2024)Tang, Liu, Ye, Lu, Wei, Lin, Li, Mahmood, Feng, Zhao, et~al.]{mtvqa}
Jingqun Tang, Qi Liu, Yongjie Ye, Jinghui Lu, Shu Wei, Chunhui Lin, Wanqing Li, Mohamad Fitri Faiz~Bin Mahmood, Hao Feng, Zhen Zhao, et~al.
\newblock Mtvqa: Benchmarking multilingual text-centric visual question answering.
\newblock \emph{arXiv preprint arXiv:2405.11985}, 2024.

\bibitem[Team et~al.(2023)Team, Anil, Borgeaud, Wu, Alayrac, Yu, Soricut, Schalkwyk, Dai, Hauth, et~al.]{gemini}
Gemini Team, Rohan Anil, Sebastian Borgeaud, Yonghui Wu, Jean-Baptiste Alayrac, Jiahui Yu, Radu Soricut, Johan Schalkwyk, Andrew~M Dai, Anja Hauth, et~al.
\newblock Gemini: a family of highly capable multimodal models.
\newblock \emph{arXiv preprint arXiv:2312.11805}, 2023.

\bibitem[Tong et~al.(2024{\natexlab{a}})Tong, Brown, Wu, Woo, IYER, Akula, Yang, Yang, Middepogu, Wang, et~al.]{cvbench}
Peter Tong, Ellis Brown, Penghao Wu, Sanghyun Woo, Adithya Jairam~Vedagiri IYER, Sai~Charitha Akula, Shusheng Yang, Jihan Yang, Manoj Middepogu, Ziteng Wang, et~al.
\newblock Cambrian-1: A fully open, vision-centric exploration of multimodal llms.
\newblock \emph{Advances in Neural Information Processing Systems}, 37:\penalty0 87310--87356, 2024{\natexlab{a}}.

\bibitem[Tong et~al.(2024{\natexlab{b}})Tong, Brown, Wu, Woo, Middepogu, Akula, Yang, Yang, Iyer, Pan, et~al.]{cambrian}
Shengbang Tong, Ellis Brown, Penghao Wu, Sanghyun Woo, Manoj Middepogu, Sai~Charitha Akula, Jihan Yang, Shusheng Yang, Adithya Iyer, Xichen Pan, et~al.
\newblock Cambrian-1: A fully open, vision-centric exploration of multimodal llms.
\newblock \emph{arXiv preprint arXiv:2406.16860}, 2024{\natexlab{b}}.

\bibitem[Touvron et~al.(2023)Touvron, Lavril, Izacard, Martinet, Lachaux, Lacroix, Rozi{\`e}re, Goyal, Hambro, Azhar, et~al.]{llama}
Hugo Touvron, Thibaut Lavril, Gautier Izacard, Xavier Martinet, Marie-Anne Lachaux, Timoth{\'e}e Lacroix, Baptiste Rozi{\`e}re, Naman Goyal, Eric Hambro, Faisal Azhar, et~al.
\newblock Llama: Open and efficient foundation language models.
\newblock \emph{arXiv preprint arXiv:2302.13971}, 2023.

\bibitem[Wang et~al.(2023)Wang, Chen, Pei, Xie, Kang, Zhang, Xu, Xiong, Dutta, Schaeffer, et~al.]{DecodingTrust}
Boxin Wang, Weixin Chen, Hengzhi Pei, Chulin Xie, Mintong Kang, Chenhui Zhang, Chejian Xu, Zidi Xiong, Ritik Dutta, Rylan Schaeffer, et~al.
\newblock Decodingtrust: A comprehensive assessment of trustworthiness in gpt models.
\newblock In \emph{NeurIPS}, 2023.

\bibitem[Wang et~al.(2024{\natexlab{a}})Wang, Huang, Yang, Fan, and Lan]{learning}
Fengxiang Wang, Wanrong Huang, Shaowu Yang, Qi Fan, and Long Lan.
\newblock Learning to learn better visual prompts.
\newblock \emph{Proceedings of the AAAI Conference on Artificial Intelligence}, 38\penalty0 (6):\penalty0 5354--5363, 2024{\natexlab{a}}.

\bibitem[Wang et~al.(2024{\natexlab{b}})Wang, Wang, Wang, Guo, Zhong, Lan, Zhang, Liu, and Sun]{selectivemae}
Fengxiang Wang, Hongzhen Wang, Di Wang, Zonghao Guo, Zhenyu Zhong, Long Lan, Jing Zhang, Zhiyuan Liu, and Maosong Sun.
\newblock Scaling efficient masked autoencoder learning on large remote sensing dataset.
\newblock \emph{arXiv preprint arXiv:2406.11933}, 2024{\natexlab{b}}.

\bibitem[Wang et~al.(2025)Wang, Wang, Wang, Wang, Chen, Zhao, Sun, Wang, Lan, Yang, et~al.]{roma}
Fengxiang Wang, Hongzhen Wang, Yulin Wang, Di Wang, Mingshuo Chen, Haiyan Zhao, Yangang Sun, Shuo Wang, Long Lan, Wenjing Yang, et~al.
\newblock Roma: Scaling up mamba-based foundation models for remote sensing.
\newblock \emph{arXiv preprint arXiv:2503.10392}, 2025.

\bibitem[Wang et~al.(2024{\natexlab{c}})Wang, Bai, Tan, Wang, Fan, Bai, Chen, Liu, Wang, Ge, et~al.]{qwen2}
Peng Wang, Shuai Bai, Sinan Tan, Shijie Wang, Zhihao Fan, Jinze Bai, Keqin Chen, Xuejing Liu, Jialin Wang, Wenbin Ge, et~al.
\newblock Qwen2-vl: Enhancing vision-language model's perception of the world at any resolution.
\newblock \emph{arXiv preprint arXiv:2409.12191}, 2024{\natexlab{c}}.

\bibitem[Wang et~al.(2024{\natexlab{d}})Wang, Chen, Wang, Cao, Liu, Gao, Zhu, Zhu, Lu, Qiao, et~al.]{internvl2}
Weiyun Wang, Zhe Chen, Wenhai Wang, Yue Cao, Yangzhou Liu, Zhangwei Gao, Jinguo Zhu, Xizhou Zhu, Lewei Lu, Yu Qiao, et~al.
\newblock Enhancing the reasoning ability of multimodal large language models via mixed preference optimization.
\newblock \emph{arXiv preprint arXiv:2411.10442}, 2024{\natexlab{d}}.

\bibitem[Xia et~al.(2018)Xia, Bai, Ding, Zhu, Belongie, Luo, Datcu, Pelillo, and Zhang]{dota}
Gui-Song Xia, Xiang Bai, Jian Ding, Zhen Zhu, Serge Belongie, Jiebo Luo, Mihai Datcu, Marcello Pelillo, and Liangpei Zhang.
\newblock Dota: A large-scale dataset for object detection in aerial images.
\newblock In \emph{Proceedings of the IEEE conference on computer vision and pattern recognition}, pages 3974--3983, 2018.

\bibitem[Yang et~al.(2018)Yang, Liao, Li, and Rosenhahn]{itcvd}
Michael~Ying Yang, Wentong Liao, Xinbo Li, and Bodo Rosenhahn.
\newblock Deep learning for vehicle detection in aerial images.
\newblock In \emph{2018 25th IEEE International Conference on Image Processing (ICIP)}, pages 3079--3083. IEEE, 2018.

\bibitem[Ying et~al.(2024)Ying, Meng, Wang, Li, Lin, Yang, Zhang, Zhang, Lin, Liu, et~al.]{mmtbench}
Kaining Ying, Fanqing Meng, Jin Wang, Zhiqian Li, Han Lin, Yue Yang, Hao Zhang, Wenbo Zhang, Yuqi Lin, Shuo Liu, et~al.
\newblock Mmt-bench: A comprehensive multimodal benchmark for evaluating large vision-language models towards multitask agi.
\newblock In \emph{International Conference on Machine Learning}, pages 57116--57198. PMLR, 2024.

\bibitem[Young et~al.(2014)Young, Lai, Hodosh, and Hockenmaier]{flickr30k}
Peter Young, Alice Lai, Micah Hodosh, and Julia Hockenmaier.
\newblock From image descriptions to visual denotations: New similarity metrics for semantic inference over event descriptions.
\newblock \emph{Transactions of the Association for Computational Linguistics}, 2:\penalty0 67--78, 2014.

\bibitem[Yu et~al.(2024)Yu, Yang, Li, Wang, Lin, Liu, Wang, and Wang]{mmvet}
Weihao Yu, Zhengyuan Yang, Linjie Li, Jianfeng Wang, Kevin Lin, Zicheng Liu, Xinchao Wang, and Lijuan Wang.
\newblock Mm-vet: Evaluating large multimodal models for integrated capabilities.
\newblock In \emph{International Conference on Machine Learning}, pages 57730--57754. PMLR, 2024.

\bibitem[Zhan et~al.(2023)Zhan, Xiong, and Yuan]{diorrsvg}
Yang Zhan, Zhitong Xiong, and Yuan Yuan.
\newblock Rsvg: Exploring data and models for visual grounding on remote sensing data.
\newblock \emph{IEEE Transactions on Geoscience and Remote Sensing}, 61:\penalty0 1--13, 2023.

\bibitem[Zhang et~al.(2024{\natexlab{a}})Zhang, Li, Zhang, Pu, Cahyono, Hu, Liu, Zhang, Yang, Li, and Liu]{zhang2024lmmsevalrealitycheckevaluation}
Kaichen Zhang, Bo Li, Peiyuan Zhang, Fanyi Pu, Joshua~Adrian Cahyono, Kairui Hu, Shuai Liu, Yuanhan Zhang, Jingkang Yang, Chunyuan Li, and Ziwei Liu.
\newblock Lmms-eval: Reality check on the evaluation of large multimodal models.
\newblock \emph{arXiv preprint arXiv:2407.12772}, 2024{\natexlab{a}}.

\bibitem[Zhang et~al.(2023)Zhang, Wang, Cao, Xu, Ouyang, Zhao, Ding, Zhang, Duan, Yan, et~al.]{InternLM-XComposer}
Pan Zhang, Xiaoyi Dong~Bin Wang, Yuhang Cao, Chao Xu, Linke Ouyang, Zhiyuan Zhao, Shuangrui Ding, Songyang Zhang, Haodong Duan, Hang Yan, et~al.
\newblock Internlm-xcomposer: A vision-language large model for advanced text-image comprehension and composition.
\newblock \emph{arXiv preprint arXiv:2309.15112}, 2023.

\bibitem[Zhang et~al.(2024{\natexlab{b}})Zhang, Dong, Zang, Cao, Qian, Chen, Guo, Duan, Wang, Ouyang, et~al.]{ixc2.5}
Pan Zhang, Xiaoyi Dong, Yuhang Zang, Yuhang Cao, Rui Qian, Lin Chen, Qipeng Guo, Haodong Duan, Bin Wang, Linke Ouyang, et~al.
\newblock Internlm-xcomposer-2.5: A versatile large vision language model supporting long-contextual input and output.
\newblock \emph{arXiv preprint arXiv:2407.03320}, 2024{\natexlab{b}}.

\bibitem[Zhang et~al.(2024{\natexlab{c}})Zhang, Cai, Zhang, Zhuang, and Mao]{earthgpt}
Wei Zhang, Miaoxin Cai, Tong Zhang, Yin Zhuang, and Xuerui Mao.
\newblock Earthgpt: A universal multi-modal large language model for multi-sensor image comprehension in remote sensing domain.
\newblock \emph{IEEE Transactions on Geoscience and Remote Sensing}, 2024{\natexlab{c}}.

\bibitem[{Zhang} et~al.(2016){Zhang}, {Sun}, {Shang}, {Zhang}, and {Wang}]{agriculture_1}
X. {Zhang}, Y. {Sun}, K. {Shang}, L. {Zhang}, and S. {Wang}.
\newblock Crop classification based on feature band set construction and object-oriented approach using hyperspectral images.
\newblock \emph{IEEE J. Sel. Topics Appl. Earth Observ. Remote Sens.}, 9\penalty0 (9):\penalty0 4117--4128, 2016.

\bibitem[Zhang et~al.(2024{\natexlab{d}})Zhang, Wen, Fu, Wang, Zhang, Wang, and Jin]{slime}
Yi-Fan Zhang, Qingsong Wen, Chaoyou Fu, Xue Wang, Zhang Zhang, Liang Wang, and Rong Jin.
\newblock Beyond llava-hd: Diving into high-resolution large multimodal models.
\newblock \emph{arXiv preprint arXiv:2406.08487}, 2024{\natexlab{d}}.

\bibitem[Zhang et~al.(2024{\natexlab{e}})Zhang, Zhang, Tian, Fu, Zhang, Wu, Li, Wang, Wen, Zhang, et~al.]{mmerealworld}
Yi-Fan Zhang, Huanyu Zhang, Haochen Tian, Chaoyou Fu, Shuangqing Zhang, Junfei Wu, Feng Li, Kun Wang, Qingsong Wen, Zhang Zhang, et~al.
\newblock Mme-realworld: Could your multimodal llm challenge high-resolution real-world scenarios that are difficult for humans?
\newblock \emph{arXiv preprint arXiv:2408.13257}, 2024{\natexlab{e}}.

\bibitem[Zheng et~al.(2021)Zheng, Wang, Du, and Lu]{rsivqa}
Xiangtao Zheng, Binqiang Wang, Xingqian Du, and Xiaoqiang Lu.
\newblock Mutual attention inception network for remote sensing visual question answering.
\newblock \emph{IEEE Transactions on Geoscience and Remote Sensing}, 60:\penalty0 1--14, 2021.

\bibitem[Zhou et~al.(2018)Zhou, Xu, and Corso]{youcook2}
Luowei Zhou, Chenliang Xu, and Jason Corso.
\newblock Towards automatic learning of procedures from web instructional videos.
\newblock In \emph{Proceedings of the AAAI Conference on Artificial Intelligence}, 2018.

\bibitem[Zhu et~al.(2024)Zhu, Chen, Shen, Li, and Elhoseiny]{minigpt}
Deyao Zhu, Jun Chen, Xiaoqian Shen, Xiang Li, and Mohamed Elhoseiny.
\newblock Minigpt-4: Enhancing vision-language understanding with advanced large language models.
\newblock In \emph{12th International Conference on Learning Representations, ICLR 2024}, 2024.

\bibitem[Zhu et~al.(2019)Zhu, Zhou, Seto, Stokes, Deng, Pickett, and Taubenböck]{urban_plan}
Zhe Zhu, Yuyu Zhou, Karen~C. Seto, Eleanor~C. Stokes, Chengbin Deng, Steward~T.A. Pickett, and Hannes Taubenböck.
\newblock Understanding an urbanizing planet: Strategic directions for remote sensing.
\newblock \emph{Remote Sensing of Environment}, 228:\penalty0 164--182, 2019.

\end{thebibliography}
}
\clearpage
\appendix

\newcommand{\roundedboxpink}[1]{
  \tikz[baseline=(char.base)]{
    \node[anchor=south west, rounded corners, text height=1.5ex, text depth=.25ex, fill=pink, draw=none, text=black, font=\bfseries] (char) {#1};
  }
}
\newcommand{\roundedboxgreen}[1]{
  \tikz[baseline=(char.base)]{
    \node[anchor=south west, rounded corners, text height=1.5ex, text depth=.25ex, fill=green!30, draw=none, text=black, font=\bfseries] (char) {#1};
  }
}
\newcommand{\roundedboxblue}[1]{
  \tikz[baseline=(char.base)]{
    \node[anchor=south west, rounded corners, text height=1.5ex, text depth=.25ex, fill=blue!30, draw=none, text=black, font=\bfseries] (char) {#1};
  }
}
\newcommand{\roundedboxyellow}[1]{
  \tikz[baseline=(char.base)]{
    \node[anchor=south west, rounded corners, text height=1.5ex, text depth=.25ex, fill=yellow!50, draw=none, text=black, font=\bfseries] (char) {#1};
  }
}
\newcommand{\roundedboxred}[1]{
  \tikz[baseline=(char.base)]{
    \node[anchor=south west, rounded corners, text height=1.5ex, text depth=.25ex, fill=red!30, draw=none, text=black, font=\bfseries] (char) {#1};
  }
}
\newcommand{\roundedboxpurple}[1]{
  \tikz[baseline=(char.base)]{
    \node[anchor=south west, rounded corners, text height=1.5ex, text depth=.25ex, fill=purple!50, draw=none, text=black, font=\bfseries] (char) {#1};
  }
}
\newcommand{\roundedboxbrown}[1]{
  \tikz[baseline=(char.base)]{
    \node[anchor=south west, rounded corners, text height=1.5ex, text depth=.25ex, fill=brown!30, draw=none, text=black, font=\bfseries] (char) {#1};
  }
}
\newcommand{\roundedboxorange}[1]{
  \tikz[baseline=(char.base)]{
    \node[anchor=south west, rounded corners, text height=1.5ex, text depth=.25ex, fill=orange!30, draw=none, text=black, font=\bfseries] (char) {#1};
  }
}
\newcommand{\roundedboxcyan}[1]{
  \tikz[baseline=(char.base)]{
    \node[anchor=south west, rounded corners, text height=1.5ex, text depth=.25ex, fill=cyan!30, draw=none, text=black, font=\bfseries] (char) {#1};
  }
}
\newcommand{\roundedboxgray}[1]{
  \tikz[baseline=(char.base)]{
    \node[anchor=south west, rounded corners, text height=1.5ex, text depth=.25ex, fill=gray!50, draw=none, text=black, font=\bfseries] (char) {#1};
  }
}

\definecolor{customcolorred}{RGB}{225,159,156} 
\definecolor{customcolorgreen}{RGB}{5,204,151} 

\newcommand{\boxedred}[1]{
  \tikz[baseline=(char.base)]{
    \node[anchor=south west, rectangle, text height=1.5ex, text depth=.25ex, fill=customcolorred, draw=none, text=black, font=\bfseries] (char) {#1};
  }
}
\newcommand{\boxedgreen}[1]{
  \tikz[baseline=(char.base)]{
    \node[anchor=south west, rectangle, text height=1.5ex, text depth=.25ex, fill=customcolorgreen, draw=none, text=black, font=\bfseries] (char) {#1};
  }
}


\section{Appendix}

\subsection{Overview of the Appendix}
This appendix supplements the proposed \textbf{XLRS-Bench} with details excluded from the main paper due to space constraints.

The appendix is organized as follows:
\begin{itemize}
\item Sec.~\ref{app-details}: More details of XLRS-Bench.
    \item Sec.~\ref{app-sec1}: Human evaluations on XLRS-Bench.
    \item Sec.~\ref{app-sec2}: More analysis on L-2 capability across various MLLMs.
    \item Sec.~\ref{app-sec3}: Detailed results of specific sub-tasks (L-3 capability).
    \item Sec.~\ref{app-sec4}: Visualizations of samples and challenging cases.
    \item Sec.~\ref{app-datasheets}: Datasheets for the XLRS-Bench dataset.
    \item Sec.~\ref{app-limitation}: Discussion on limitations and societal impact.
\end{itemize}

\subsection{More Details of XLRS-Bench}
\label{app-details}
\begin{table*}[h]
{\renewcommand{\arraystretch}{1.5}
\footnotesize
    \caption{Characteristics and vision-language formats of L3 sub-tasks. }
    \centering
    \resizebox{0.98\textwidth}{!}{
     \begin{tabular}{c|c|cccccc}
        \toprule 
L1-Task & L2-Task & L3-Task & Abbr. & Annotation Format & Annotation Method & Number of Samples & Answer Type\\ \hline
\multirow{10}{*}{Perception} &  \multirow{2}{*}{Counting}  & Overall Counting & OC & VQA & All Human & 370 & Multiple Choice(A/B/C/D)\\
 &  & Regional Counting & RC & VQA & All Human & 972 & Multiple Choice(A/B/C/D)\\
\cline{2-8}
 & \multirow{2}{*}{Scene Classification}  & Overall Land Use Classification & OLUC & VQA & All Human & 904 & Multiple Choice(A/B/C/D)\\
 &  & Regional Land Use Classification & RLUC & VQA & All Human & 1854 & Multiple Choice(A/B/C/D)\\
 \cline{2-8}
 & Object Spatial Relationship & Object Spatial Relationship & OSR & VQA & All Human & 4819 & Multiple Choice(A/B/C/D)\\
\cline{2-8}
 & \multirow{3}{*}{Object Properties}  & Object Classification & OCC & VQA & All Human & 9172 & Multiple Choice(A/B/C/D)\\
 &  & Object Color & OCL & VQA & All Human & 8930 & Multiple Choice(A/B/C/D)\\
 &  & Object Motion State & OMS & VQA & All Human & 640 & Multiple Choice(A/B for Yes/No)\\
 \cline{2-8}
 & Image Captioning & Detailed Image Captioning & - & Caption & Semi-automated & 934 & Plain Text\\
 \cline{2-8}
 & \multirow{2}{*}{Visual Grounding} & Fine-grained Visual Grounding & - & Visual Grounding & All Human & 6310 & Bounding Box\\
\cline{1-1}

\multirow{6}{*}{Reasoning} & & Condition-based Visual Grounding & - & Visual Grounding & All Human & 6305 & Bounding Box\\
\cline{2-8}
 & Route Planning & Route Planning & RP & VQA & All Human & 1130 & Multiple Choice(A/B/C/D)\\
 \cline{2-8}
 & Anomaly Reasoning & Anomaly Detection and Interpretation & AD & VQA & All Human & 1131 & Multiple Choice(A/B/C/D)\\
  \cline{2-8}
 & \multirow{2}{*}{Complex Reasoning}  & Environmental Condition Reasoning & ECR & VQA & All Human & 1125 & Multiple Choice(A/B/C/D)\\
 &  & Counting with Complex Reasoning & CCR & VQA & All Human & 972 & Multiple Choice(A/B/C/D)\\
\cline{2-8}
 & Spatiotemporal Reasoning & Regional Counting with Change Detection & RCCD & VQA & All Human & 270 & Multiple Choice(A/B/C/D)\\
        \bottomrule
    \end{tabular}}
    \label{tab:statistics}
    \vspace{-2mm}
}
\end{table*}
We provide additional details about the dataset, with Table \ref{tab:statistics} presenting statistics for VQA, visual grounding, and image captioning tasks, along with their relationships to the L3 sub-tasks. This clarifies the dataset's structure and composition. Notably, Visual Grounding spans both perception and reasoning, with Fine-grained Visual Grounding classified under perception and Condition-based Visual Grounding under reasoning.

\subsection{Human Evaluations on XLRS-Bench}
\label{app-sec1}
\begin{figure*}[h]
\centering
\includegraphics[width=\linewidth]{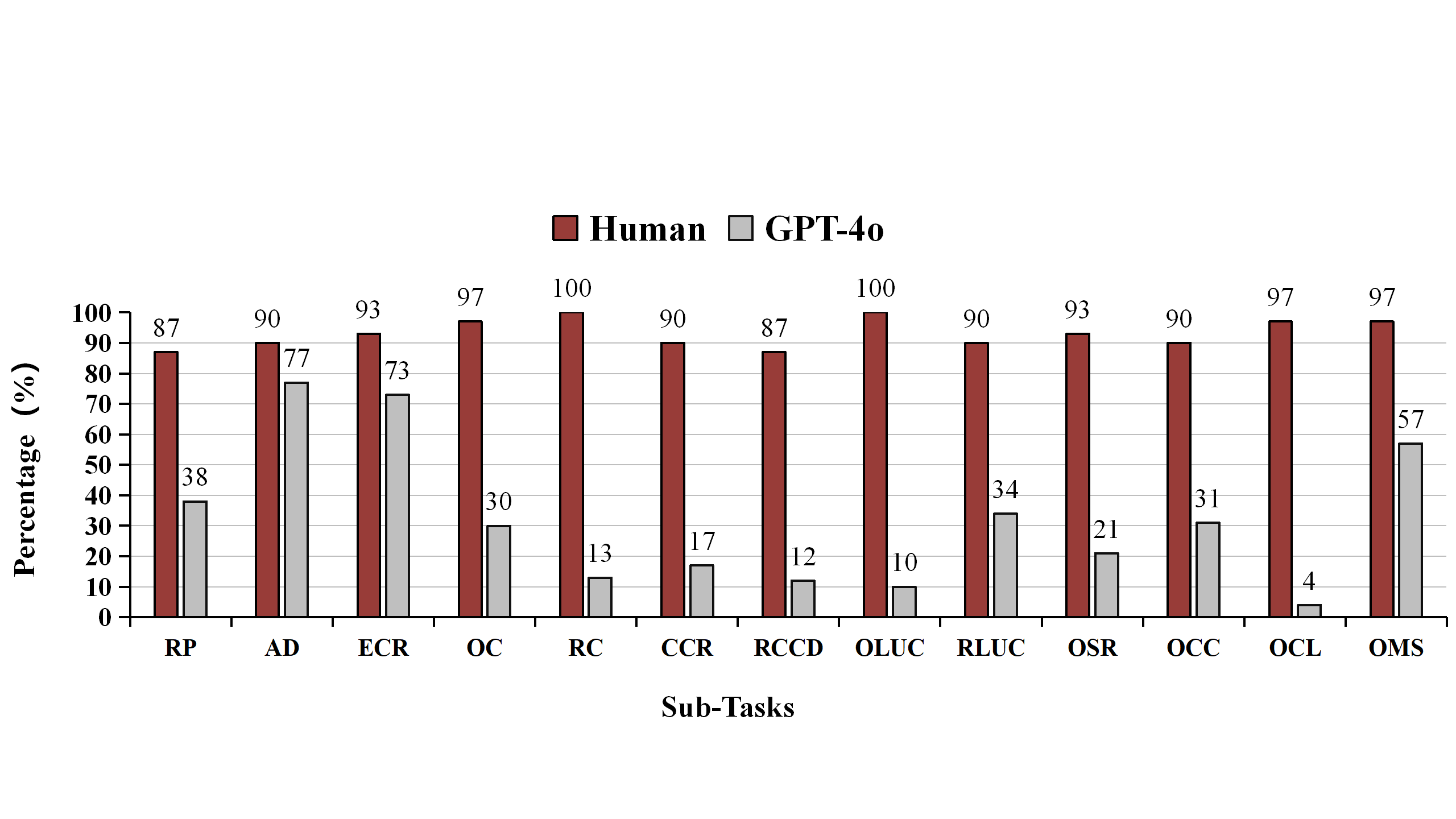}
\caption{\textbf{Evaluation results of XLRS-Bench and MLLMs.} “RP”, “AD”, “ECR”, “OCC”, “RC”, “CCR”, “RCCD”, “OLUC”, “RLUC”, “OSR”, “OCC”, “OCL”, and “OMS” each indicate a specific task domain: Route Planning, Anomaly Detection, Environmental Conditional Reasoning, Overall Counting, Regional Counting, Counting with Complex Reasoning, Regional Counting with Change Detection, Overall Land Use Classification, Regional Land Use Classification, Object Spatial Relationship, Object Classification, Object Color and Object Motion State. }
 
\label{fig:human}
\vspace{-2mm}
\end{figure*}

Human evaluation is essential for assessing dataset effectiveness~\cite{mmbench}. For XLRS-Bench, we randomly selected 30 questions from each VQA sub-task (L-3 dimensions) and had two groups answer them simultaneously. The final accuracy was computed as the average accuracy of both groups. Figure \ref{fig:human} illustrates the evaluation results of MLLMs and humans.

We observed that human accuracy consistently exceeded 90\%, validating the reliability of XLRS-Bench. However, human evaluation is not error-free, as analyzing large ultra-high-resolution RS images demands intense focus and frequent zooming, particularly for tasks like global counting, making it inherently challenging. In contrast, existing MLLMs, such as the closed-source GPT-4o, performed significantly worse, likely due to insufficient training on real ultra-high-resolution RS data. We encourage future research to address these challenges.

\subsection{More Analysis of Results on XLRS-Bench}
\label{app-sec2}

Due to space limitations, more in-depth analyses to advance MLLM research in ultra-high-resolution remote sensing scenarios are provided in the appendix. This section highlights the performance of all L-2 capabilities.

\textbf{Most MLLMs underperform across all 16 evaluation dimensions.} The accuracy of most MLLMs remains below 50\%, in sharp contrast to the 80\%–90\% typically observed in common benchmarks \cite{mmbench,seedbench,singh2019towards,chartqa}. Notably, the high accuracy and minimal variation among advanced models in these benchmarks often obscure their practical utility, reducing the significance of small improvements. The consistently low performance on XLRS-Bench underscores the distinct challenges of ultra-high-resolution remote sensing, driven by a lack of pretraining on annotated data. This highlights the pressing need for specialized models to address these complexities.

\textbf{Performance Gap: Anomaly vs. Spatiotemporal Reasoning} 
A notable performance gap exists between Anomaly Reasoning (AR) and Spatiotemporal Reasoning (SR) tasks. While most models achieve about 70\% accuracy on AR tasks, their performance drops sharply to 15.2\% on SR tasks. This discrepancy arises because AR tasks depend on identifying global anomalies with clear patterns, whereas SR tasks demand intricate local spatiotemporal modeling. Current MLLMs excel at detecting static anomalies but struggle with dynamic pattern comprehension. To bridge this gap, optimizing MLLMs should focus on improving temporal feature modeling, such as enhancing Transformer architectures to better handle sequential data.

\textbf{Limited Benefits of Larger LLMs in Perception Tasks.} In the Counting and Scene Classification (SC) subtasks, LLaVA-Next (Llama3-8B) offers little advantage over the smaller Qwen2-VL (Qwen2-7B), indicating that model size is not a primary determinant of performance. Instead, factors like diverse pretraining data and effective task alignment mechanisms likely play a more significant role. This underscores the reliance of perception tasks on the visual module's capabilities rather than the language model's reasoning. Future efforts could prioritize smaller, more efficient models tailored for perception and explore distillation techniques to enhance visual module performance with reduced model sizes.

\textbf{Poor Performance in Visual Grounding Tasks.}
On XLRS-Bench, MLLMs underperform significantly, as shown in main text. On both the Chinese (XLRS-Bench-ZH) and English (XLRS-Bench-EN) benchmarks, most models achieve less than 1.0\% accuracy in terms of Acc@0.5 and Acc@0.7 metrics, highlighting major limitations in their ability to handle visual localization tasks. Key issues include:  
1. Inadequate local feature extraction, hindering fine-grained localization in ultra-high-resolution images.  
2. Weak cross-modal alignment, limiting accurate matching between language descriptions and complex visual scenes.  
3. Poor generalization to real-world remote sensing scenarios, particularly with high object similarity.  
4. Limited reasoning capability (e.g., condition-based visual grounding, L-3 capability) in ultra-high-resolution settings, underscoring the need for more effective multimodal representation learning.
Future research could focus on better visual feature extraction, enhanced language alignment, and stronger generalization and reasoning in complex, high-resolution contexts.

\subsection{Sub-tasks (L-3 capability) Results on XLRS-Bench}
\begin{table*}[t]

\caption{\textbf{Experimental results of L-3 capability on the perception dimension of VQA tasks.} Models are ranked according to their average performance. Rows corresponding to proprietary models are highlighted in gray for distinction. “OC”, “RC”, “OLUC”, “RLUC”, “OSR”,“OCC”,“OCL” and “OMS” each indicate a specific task domain: Overall counting, Regional Counting, Overall Land Use classification, Regional Land Use Classification, Object Spatial Relationship, Object Classification, Object Color and Object Motion State.}
\label{tab:perception-L3}
\centering
\resizebox{0.95\textwidth}{!}{%
\begin{tabular}{llcccccccccc}
\toprule 
\multicolumn{1}{c}{\textbf{Method}}&\multicolumn{1}{c}{\textbf{LLM}} &\textbf{Language} & \multicolumn{9}{c}{\textbf{Perception}}  \\\cmidrule{1-3} \cmidrule{4-12}
\multicolumn{3}{c}{\textbf{Subtasks (L-3 Capability)}} & \textbf{OC} & \textbf{RC} & \textbf{OLUC} & \textbf{RLUC} & \textbf{OSR}& \textbf{OCC} & \textbf{OCL} & \textbf{OMS} & \textbf{Avg} \\   \hline 

CogVLM2 & Llama3-8B & en & 31.89 & 38.48 & 1.53 & 68.07 & \textbf{35.92} & \textbf{40.77} & 30.93 & \textbf{64.53} & \textbf{37.65} \\
LLaVA-OneVision & Qwen2-7B & en & 29.19 & 38.37 & 1.11 & \textbf{70.50} & 32.60 & 35.63 & \textbf{34.37} & 62.50 & 36.54 \\
Qwen2-VL & Qwen2-7B & en & \textbf{32.43} & \textbf{42.49} & 5.86 & 68.99 & 32.04 & 35.05 & 33.34 & 59.53 & 36.09 \\
\Lgray GPT-4o-mini & - & en & 20.27 & 31.38 & \textbf{18.69} & 58.52 & 29.96 & 39.97 & 30.56 & 63.44 & 35.71 \\
InternVL2 & InternLM2.5-7B & en & 22.97 & 38.07 & 8.19 & 62.84 & 26.60 & 35.02 & 32.79 & 60.94 & 34.37 \\
InternLM-XComposer-2.5 & InternLM2-7B & en & 26.76 & 39.30 & 1.11 & 70.01 & 32.75 & 33.41 & 29.06 & 11.72 & 32.90 \\
LLaVA-Next & Llama3-8B & en & 27.84 & 41.87 & 1.11 & 60.14 & 32.50 & 29.60 & 30.59 & 63.12 & 32.72 \\
\Lgray GPT-4o & - & en & 24.32 & 31.48 & 16.81 & 64.02 & 32.35 & 19.59 & 29.00 & 40.31 & 28.70 \\
LLaVA-1.5 & Vicuna-7B & en & 24.05 & 22.12 & 0.00 & 29.29 & 23.05 & 21.65 & 17.45 & 38.12 & 20.77 \\
GeoChat & Vicuna-7B & en & 24.05 & 22.12 & 1.00 & 27.72 & 23.30 & 21.65 & 17.45 & 38.13 & 20.74 \\
\midrule
Qwen2-VL & Qwen2-7B & zh & \textbf{32.16} & 42.28 & 1.77 & \textbf{72.44} & 33.26 & \textbf{40.07} & \textbf{34.06} & 60.00 & \textbf{38.29} \\
InternVL2 & InternLM2.5-7B & zh & 22.43 & 38.99 & 3.65 & 56.42 & 34.24 & 39.60 & 33.75 & 60.78 & 36.97 \\
LLaVA-OneVision & Qwen2-7B & zh & 27.03 & 41.46 & 1.11 & 69.47 & 31.73 & 39.00 & 28.33 & 62.50 & 35.56 \\
\Lgray GPT-4o-mini & - & zh & 20.00 & 33.13 & \textbf{25.66} & 53.13 & 29.59 & 37.58 & 31.18 & \textbf{63.28} & 34.98 \\
InternLM-XComposer-2.5 & InternLM2-7B & zh & 23.24 & \textbf{43.21} & 1.22 & 63.59 & 32.62 & 33.92 & 31.55 & 51.72 & 34.44 \\
CogVLM2 & Llama3-8B & zh & 27.84 & 39.40 & 11.50 & 62.51 & \textbf{34.57} & 29.26 & 30.56 & 62.81 & 33.37 \\
LLaVA-Next & Llama3-8B & zh & 29.73 & 34.36 & 1.33 & 58.14 & 31.98 & 26.58 & 29.85 & 61.88 & 31.00 \\
\Lgray GPT-4o & - & zh & 18.11 & 23.87 & 12.39 & 61.60 & 31.25 & 13.91 & 33.57 & 62.19 & 27.95 \\
GeoChat & Vicuna-7B & zh & 24.05 & 22.12 & 0.72 & 28.16 & 23.05 & 21.65 & 17.45 & 38.13 & 20.72 \\
LLaVA-1.5 & Vicuna-7B & zh & 24.05 & 22.12 & 0.66 & 28.05 & 23.01 & 21.65 & 17.45 & 38.12 & 20.70 \\

\bottomrule

\end{tabular}%
}
\vspace{-1mm}
\end{table*}

\begin{table*}[t]
\caption{\textbf{Experimental results of L-3 capability on the reasoning dimension of VQA tasks.} Models are ranked according to their average performance. Rows corresponding to proprietary models are highlighted in gray for distinction. “RP”, “AD”, “ECR”, “CCR” and “RCCD” each indicate a specific task domain: Route Planning, Anomaly Detection, Environmental Conditional Reasoning, Counting with Complex Reasoning and Regional Counting with Change Detection. }
\label{tab:reasoning-L3}
\centering
\resizebox{0.95\textwidth}{!}{%
\begin{tabular}{llccccccc}
\toprule 
\multicolumn{1}{c}{\textbf{Method}}&\multicolumn{1}{c}{\textbf{LLM}} &\textbf{Language} & \multicolumn{6}{c}{\textbf{Reansoning}}  \\\cmidrule{1-3} \cmidrule{4-9}
\multicolumn{3}{c}{\textbf{Subtasks (L-3 Capability)}} & \textbf{RP} & \textbf{AD} & \textbf{ECR} & \textbf{CCR} & \textbf{RCCD}   & \textbf{Avg} \\   \hline 

InternVL2 & InternLM2.5-7B & en & 33.01 & \textbf{74.54} & 77.07 & \textbf{54.12} & 44.44 & \textbf{58.97} \\
InternLM-XComposer-2.5 & InternLM2-7B & en & 35.31 & 69.50 & 77.87 & 49.59 & 32.22 & 56.83 \\
Qwen2-VL & Qwen2-7B & en & 32.12 & 68.35 & 79.29 & 46.81 & \textbf{45.93} & 56.33 \\
LLaVA-Next & Llama3-8B & en & 26.02 & 69.10 & 76.00 & 45.47 & 32.22 & 53.14 \\
LLaVA-OneVision & Qwen2-7B & en & 24.07 & 71.88 & \textbf{79.91} & 37.76 & 37.78 & 53.00 \\
\Lgray GPT-4o & - & en & \textbf{41.24} & 72.06 & 75.29 & 26.34 & 21.85 & 52.79 \\
CogVLM2 & Llama3-8B & en & 34.16 & 69.85 & 73.07 & 45.37 & - & 52.70 \\
\Lgray GPT-4o-mini & - & en & 33.81 & 72.06 & 75.02 & 31.49 & 15.19 & 51.60 \\
LLaVA-1.5 & Vicuna-7B & en & 38.67 & 34.70 & 41.51 & 25.00 & 29.26 & 34.97 \\
GeoChat & Vicuna-7B & en & 32.12 & 33.25 & 34.84 & 25.10 & - & 29.71 \\
\midrule
InternVL2 & InternLM2.5-7B & zh & 40.09 & \textbf{76.57} & \textbf{85.51} & 48.97 & 44.07 & \textbf{62.14} \\
Qwen2-VL & Qwen2-7B & zh & 24.34 & \textbf{76.57} & 83.11 & \textbf{49.69} & \textbf{44.44} & 57.89 \\
LLaVA-OneVision & Qwen2-7B & zh & 29.12 & 74.80 & 82.13 & 39.92 & 30.37 & 55.51 \\
\Lgray GPT-4o-mini & - & zh & \textbf{41.86} & 74.71 & 73.78 & 34.05 & 21.85 & 54.84 \\
InternLM-XComposer-2.5 & InternLM2-7B & zh & 29.47 & 69.14 & 80.89 & 42.28 & 24.44 & 54.06 \\
CogVLM2 & Llama3-8B & zh & 26.19 & 70.47 & 75.47 & 41.15 & - & 50.60 \\
LLaVA-Next & Llama3-8B & zh & 21.59 & 69.85 & 73.51 & 31.58 & 25.56 & 48.33 \\
\Lgray GPT-4o & - & zh & 27.08 & 69.41 & 66.76 & 20.78 & 15.19 & 45.05 \\
LLaVA-1.5 & Vicuna-7B & zh & 38.67 & 37.58 & 41.60 & 25.10 & 29.26 & 35.72 \\
GeoChat & Vicuna-7B & zh & 22.79 & 23.74 & 24.44 & 25.10 & - & 22.58 \\

\bottomrule
\end{tabular}%
}

\end{table*}

\begin{table*}[h]
\footnotesize
    \caption{Visual grounding performance of L-3 capability on XLRS-Bench. }
    \centering
    \resizebox{0.98\textwidth}{!}{
     \begin{tabular}{c|ccccccccccccc}
        \toprule 
        \textbf{L-3 Capability}& \textbf{Language} & \textbf{Method}  & \textbf{GPT-4o} & \textbf{GPT-4o-mini} & \textbf{Qwen2-VL} & \textbf{LLaVA-OneVision} & \textbf{LLaVA-Next} & \textbf{LLaVA-1.5} & \textbf{CogVLM2} & \textbf{InternLM-XComposer-2.5} & \textbf{InternVL2} & \textbf{GeoChat}\\
        \hline
      Fine-grained&\multirow{2}{*}{en}
          &Acc@0.5 &  \textbf{0.70} & 0.17 & 0.21 & 0.25 & 0.16 & 0.11 & 0.02 & 0.03 & 0.46 & 0.21 \\ 
        Visual Grounding&  &Acc@0.7  &  0.10 & 0.06 & 0.03 & 0.00 & 0.08 & 0.00 & 0.00 & 0.02 & \textbf{0.17} & 0.02 \\
        \midrule
       Condition-based &\multirow{2}{*}{en}
       &Acc@0.5  &  \textbf{0.21} & 0.00 & 0.08 & 0.06 & 0.19 & 0.06 & 0.00 & 0.00 & 0.19 & 0.06 \\
    Visual Grounding&  &Acc@0.7  & 0.00 & 0.00 & 0.02 & 0.00 & 0.00 & 0.00 & 0.00 & 0.00 & \textbf{0.06} & 0.00 \\
    \midrule
        Fine-grained &\multirow{2}{*}{zh}
        &Acc@0.5 &    \textbf{0.76} & 0.22 & 0.22 & 0.24 & 0.05 & 0.13 & 0.02 & 0.08 & 0.38 & 0.21 \\
      Visual Grounding& &Acc@0.7  &  0.05 & 0.05 & 0.02 & 0.02 & 0.00 & 0.02 & 0.00 & 0.00 & \textbf{0.11} & 0.02 \\
      \midrule
       Condition-based &\multirow{2}{*}{zh}
       &Acc@0.5  &   0.14 & \textbf{0.20} & 0.06 & 0.02 & 0.08 & 0.11 & 0.03 & 0.03 & 0.17 & 0.06 \\
     Visual Grounding& &Acc@0.7  &   0.00 & 0.00 & 0.00 & 0.00 & \textbf{0.03} & 0.02 & 0.00 & 0.00 & \textbf{0.03} & 0.00 \\
        \bottomrule
    \end{tabular}}
    \label{tab:grounding-L3}
    \vspace{-2mm}
\end{table*}
\label{app-sec3}

This section highlights the performance of MLLMs across all L-3 capabilities. The VQA task is split into perception and reasoning dimensions, with results shown in Tables \ref{tab:perception-L3} and \ref{tab:reasoning-L3}, respectively. L-3 capabilities for the Visual Grounding task are summarized in Table \ref{tab:grounding-L3}.

\textbf{MLLMs generally excel in reasoning tasks compared to perception tasks in XLRS-Bench.} 
On most benchmarks~\cite{mmbench,mmerealworld}, MLLMs excel in perception tasks but struggle with reasoning, which requires complex conditional interpretation. However, XLRS-Bench presents a reversed trend: MLLMs perform worse in perception due to its ultra-high-resolution images averaging 8,500$\times$8,500 pixels—24 times higher than those in MME-Realworld~\cite{mmerealworld}. With a 4K resolution limit, current MLLMs cannot process such detailed imagery effectively. In contrast, reasoning tasks, often based on global patterns, are less resolution-dependent. XLRS-Bench highlights the need for next-generation MLLMs capable of handling ultra-high-resolution data, a crucial step for real-world remote sensing applications.

\textbf{Capturing local features is crucial for perception tasks.} Ultra-high-resolution visual tasks like Object Spatial Relationship (OSR) and Object Color (OCL), as shown in Tables \ref{tab:perception-L3}-\ref{tab:grounding-L3}, exhibit significantly lower accuracy. For example, LLaVA1.5 achieves only 17.45\% on the OCL task in the Chinese benchmark. This shortfall arises from three main factors:
1. Limited sensitivity to sparse details. Ultra-high-resolution images (e.g., 8,500$\times$8,500 pixels) feature sparse yet critical details, such as small object contours and intricate local relationships. Existing MLLMs, optimized for global feature extraction, struggle to detect these fine-grained signals, impairing performance on detail-oriented tasks.
2. Imbalanced global and local feature modeling. Current visual encoders emphasize global semantics over local-global interactions. For instance, OCL tasks require identifying subtle color variations among objects in localized regions, yet an overemphasis on global features can lead to misclassification of these local color details.
3. Resolution and computational constraints. MLLMs face input resolution limits (e.g., 4K), necessitating downsampling that degrades local features and hampers detailed modeling.
Ultra-high-resolution scenarios demand enhanced local feature modeling to meet their complex requirements. Strengthening this capability is essential for improving performance in these perception tasks.

\subsection{Samples and Hard Cases of XLRS-Bench}
\label{app-sec4}
\begin{figure*}[htbp]
\centering
\includegraphics[width=1\linewidth]{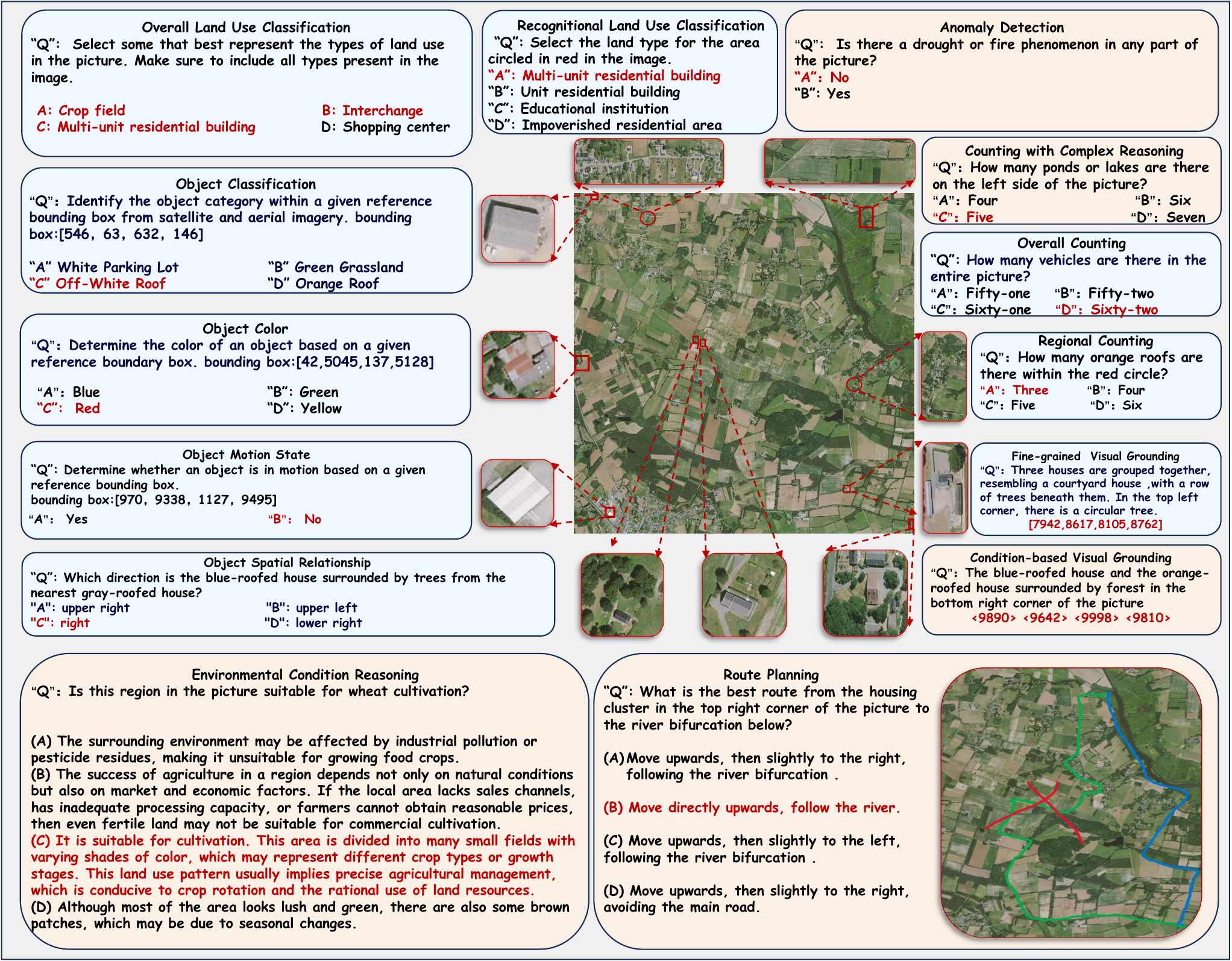}
\caption{Example of XLRS-Bench in English. XLRS-Bench focuses on large-size ultra-high-resolution remote sensing imagery, integrating over 10 multimodal perception and reasoning tasks within the same image.}
\label{fig:all-en}
\vspace{-1mm}
\end{figure*}

\begin{figure*}[t!]
\centering
\includegraphics[width=1\linewidth]{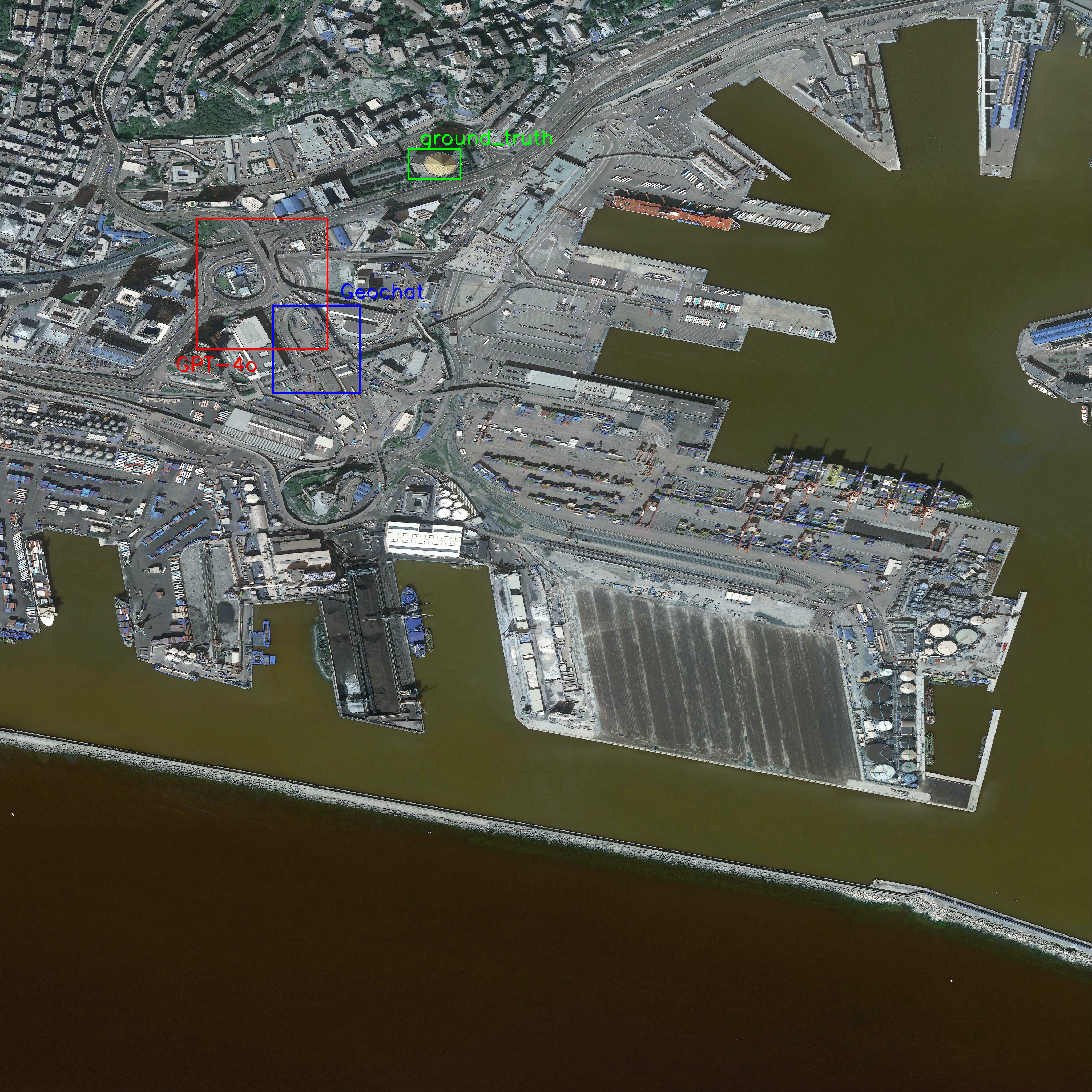}
\caption{\textbf{Visual Grounding Results of XLRS-Bench.} Question:\textit{"The multi-sided building in the left central area of the picture."} The "multi-sided building" required in the ground truth is relatively small and therefore difficult to identify. The GPT-4o model incorrectly classified a similarly shaped roundabout as a polygonal building, while GeoChat~\cite{geochat} misidentified an irregularly shaped parking lot as a polygonal building.}
\label{fig:grounding}
\vspace{-1mm}
\end{figure*}

\begin{figure*}[t!]
\centering
\includegraphics[width=0.8\linewidth]{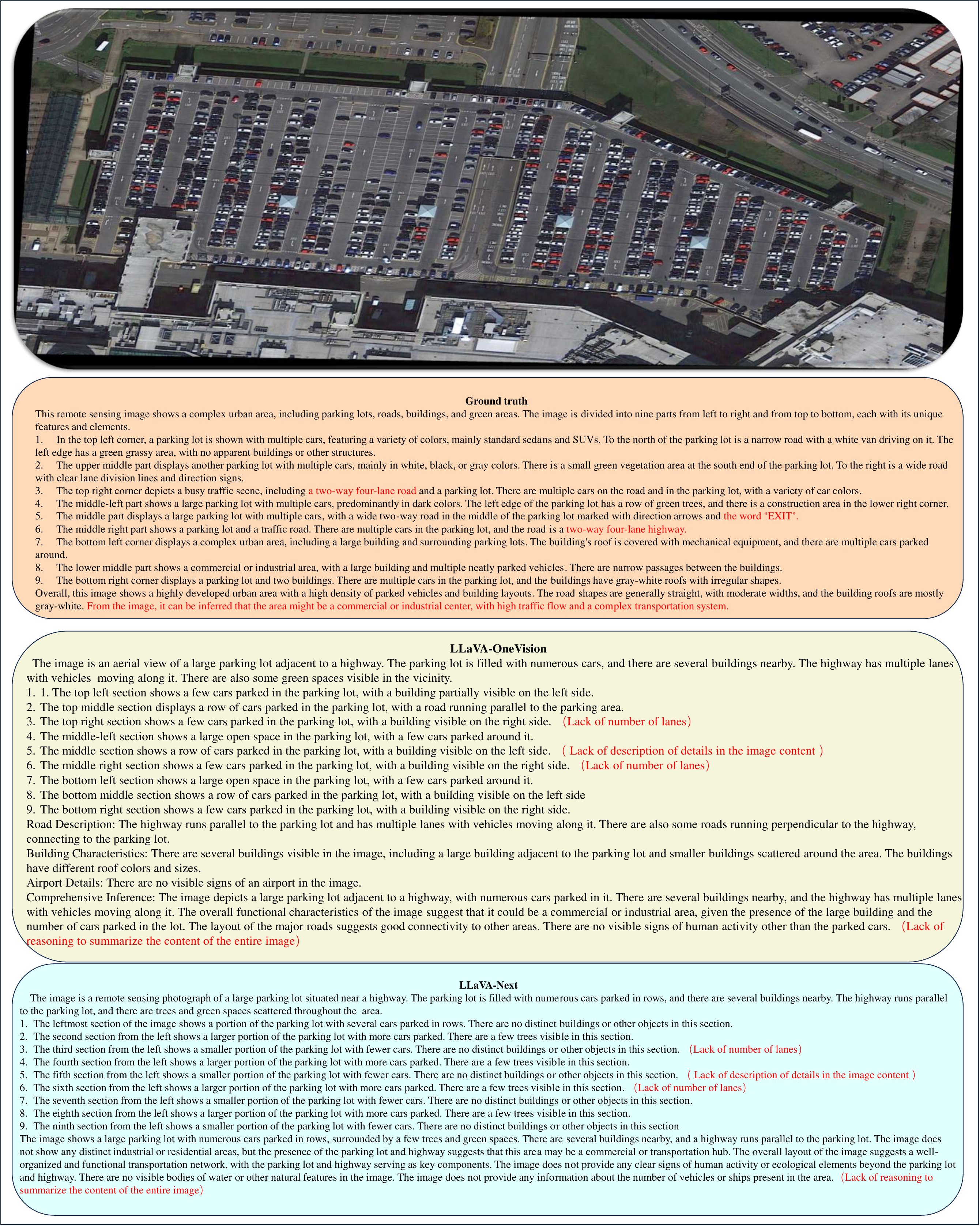}
\caption{\textbf{Image Captioning Results of XLRS-Bench in English.} LLaVA-OneVision and LLaVA-Next face challenges in capturing image details, particularly in conveying critical information like lane counts and vehicle types. Their descriptions often lack depth, failing to convey the richness and nuances of the images. The language is overly rigid and mechanical, struggling to naturally align with the images' context and overall environment.}
\label{fig:caption-en}
\vspace{-1mm}
\end{figure*}

\begin{figure*}[t!]
\centering
\includegraphics[width=0.75\linewidth]{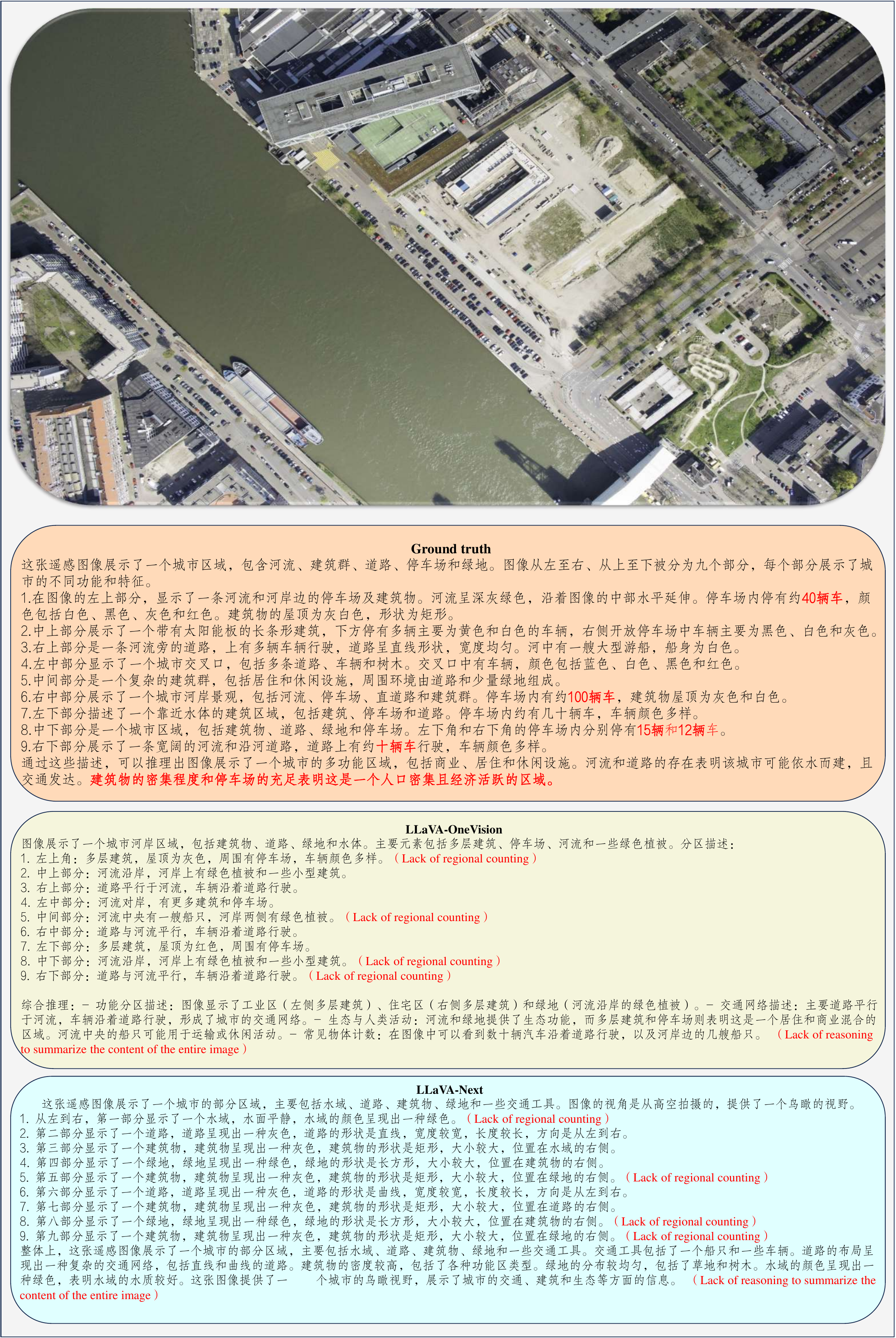}
\caption{\textbf{Image Captioning Results of XLRS-Bench in Chinese.} Both LLaVA-Next and LLaVA-OneVision share a key weakness: the inability to perform local counting. LLaVA-Next shows notable limitations and a narrow focus when describing localized features, resulting in overly simplistic outputs. While LLaVA-OneVision offers greater diversity and detail, it still lacks sufficient complexity. More importantly, both models are confined to directly describing objects in images without engaging in deeper reasoning or analysis, restricting their practical utility.}
\label{fig:caption-cn}
\vspace{-1mm}
\end{figure*}

In this section, we present examples of the VQA (Fig. \ref{fig:all-en}), image captioning (Fig. \ref{fig:caption-en} and Fig. \ref{fig:caption-cn}), and visual grounding tasks (Fig. \ref{fig:grounding} ). What's more, we construct a detailed table (Tab. \ref{tab:error_case}) analyzing model performance and error causes for each L-3 subtask. We then use examples to thoroughly illustrate the errors for each subtask.

\begin{figure*}
    \centering
    \includegraphics[width=0.99\linewidth]{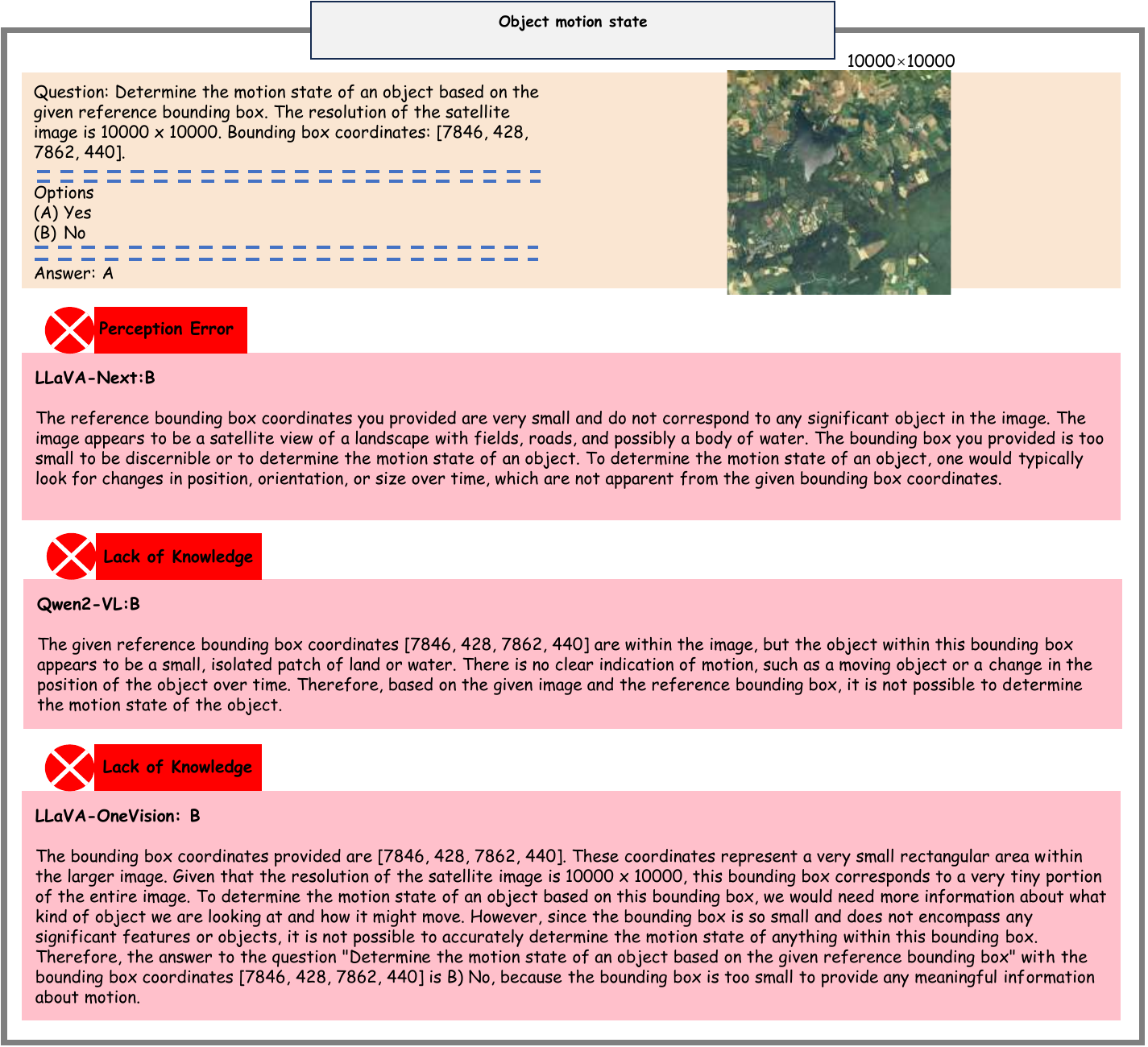}
    \caption{A sample case of Object motion state. }
    \label{fig:error_12}
\end{figure*}

\begin{figure*}
    \centering
    \includegraphics[width=1\linewidth]{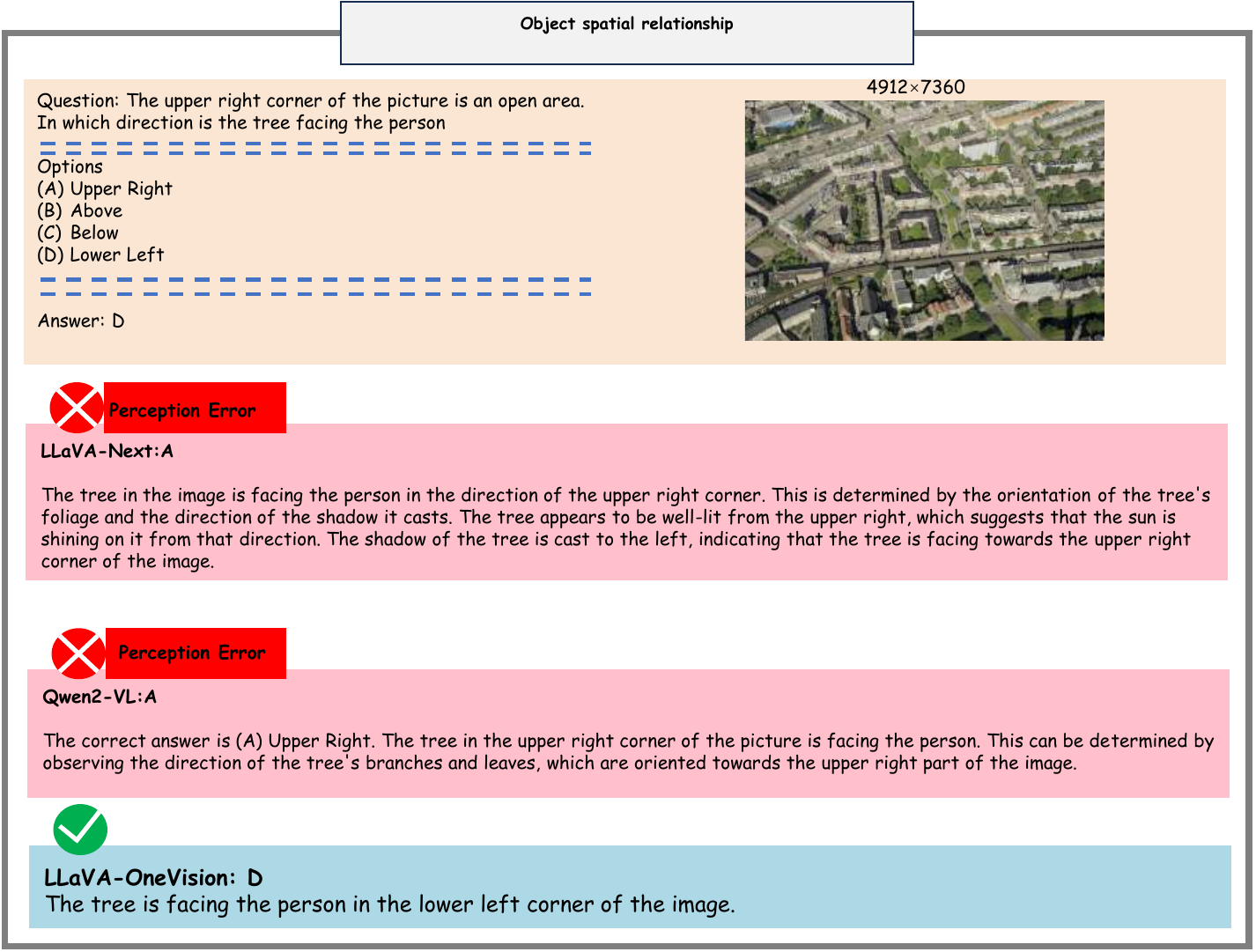}
    \caption{A sample case of Object spatial relationship. }
    \label{fig:error_13}
\end{figure*}

In this section, we present a case study analysis of the error types made by LLaVa-Next, Qwen2-VL, and LLaVA-OneVision on various sub-tasks in XLRS-Bench. We classify the errors into the following 5 categories, following the MMT-Bench~\cite{mmtbench}:

\roundedboxyellow{Perception Error}: MLLMs often struggle to recognize, classify, or detect objects and content in images, largely due to the limited representational power of their visual encoders, making this the most prevalent error. This perceptual limitation is especially evident in ultra-high-resolution images, where MLLMs often struggle to detect objects with minimal pixel representation. See examples in Fig.~\ref{fig:error_12}, Fig.~\ref{fig:error_13}, etc.

\begin{figure*}
    \centering
    \includegraphics[width=1\linewidth]{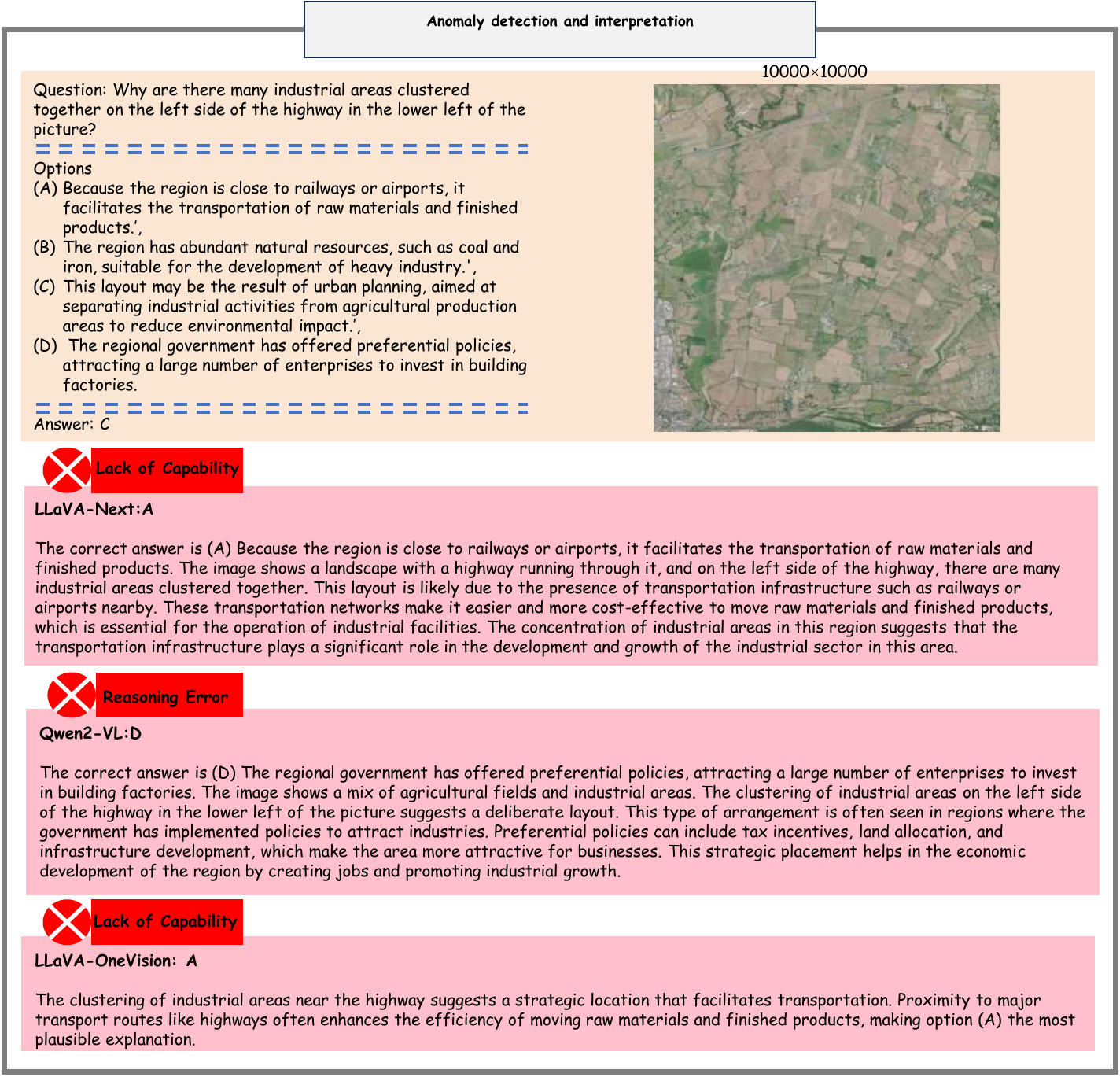}
    \caption{A sample case of Anomaly detection and interpretation. }
    \label{fig:error_1}
\end{figure*}

\begin{figure*}
    \centering
    \includegraphics[width=1\linewidth]{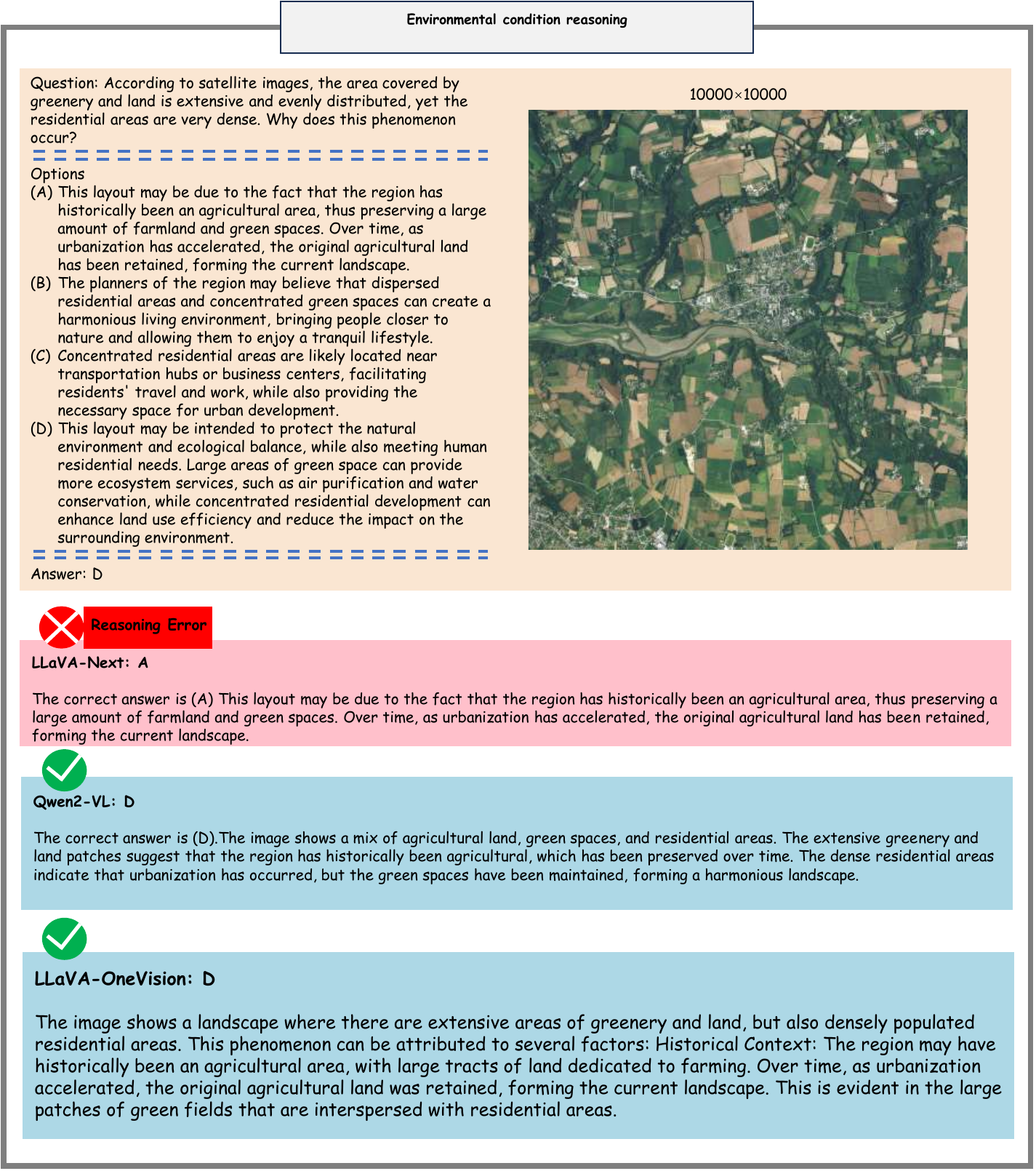}
    \caption{A sample case of Environmental condition reasoning. }
    \label{fig:error_2}
\end{figure*}

\roundedboxred{Reasoning Error}: MLLMs accurately perceive visual content but fail in reasoning, resulting in incorrect answers.. See examples in Fig.~\ref{fig:error_1}, Fig.~\ref{fig:error_2}, etc.

\roundedboxblue{Lack of Knowledge}: MLLMs lack the domain-specific knowledge needed to answer specialized questions, such as identifying ship wake information in remote sensing images (see Fig.~\ref{fig:error_12}).

\begin{figure*}
    \centering
    \includegraphics[width=1\linewidth]{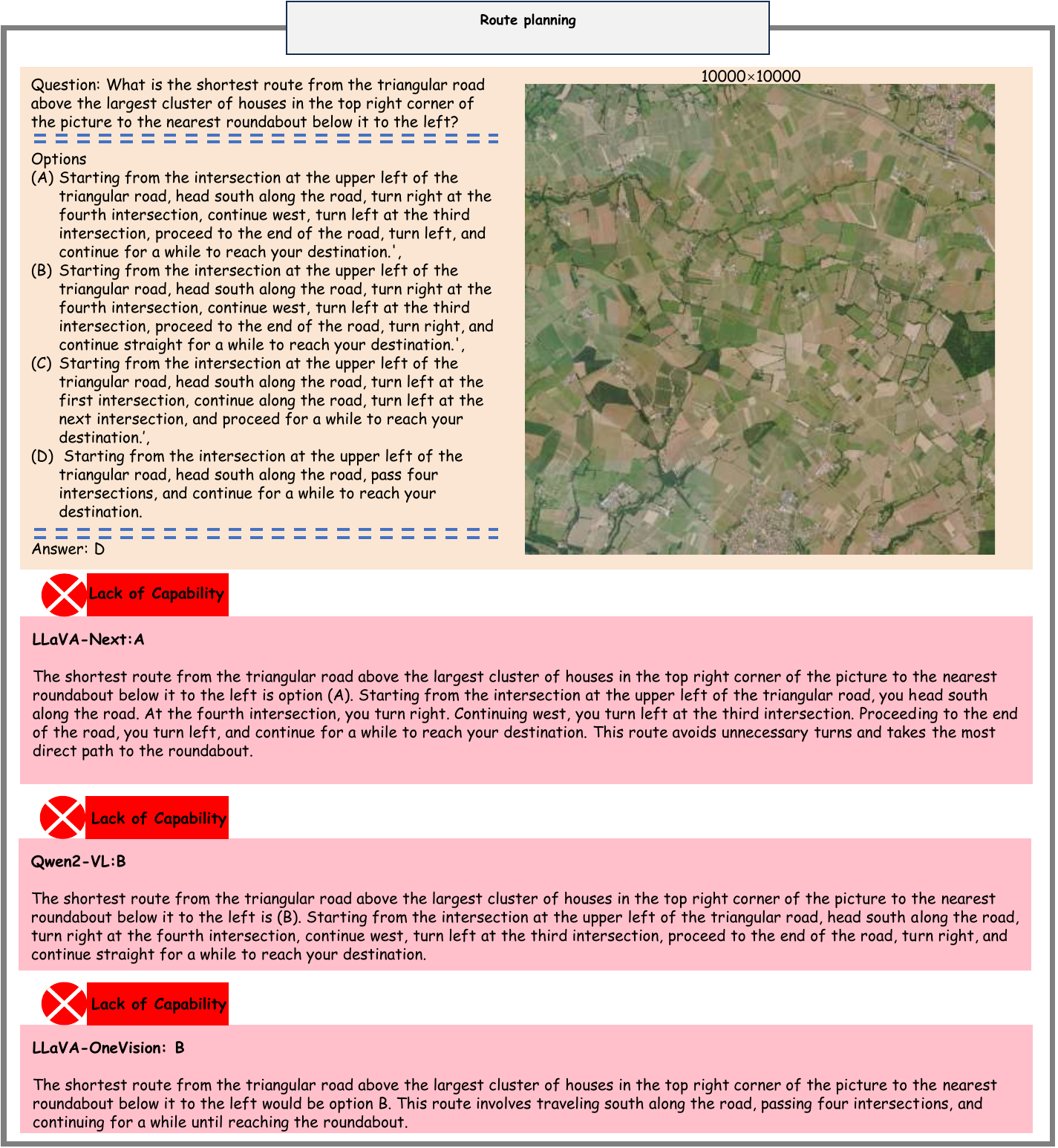}
    \caption{A sample case of Route planning. }
    \label{fig:error_3}
\end{figure*}

\roundedboxcyan{Lack of Capability}: MLLMs do not have the capability to solve the corresponding tasks. See examples in Fig.~\ref{fig:error_1}, Fig.~\ref{fig:error_3}.

\begin{figure*}
    \centering
    \includegraphics[width=1\linewidth]{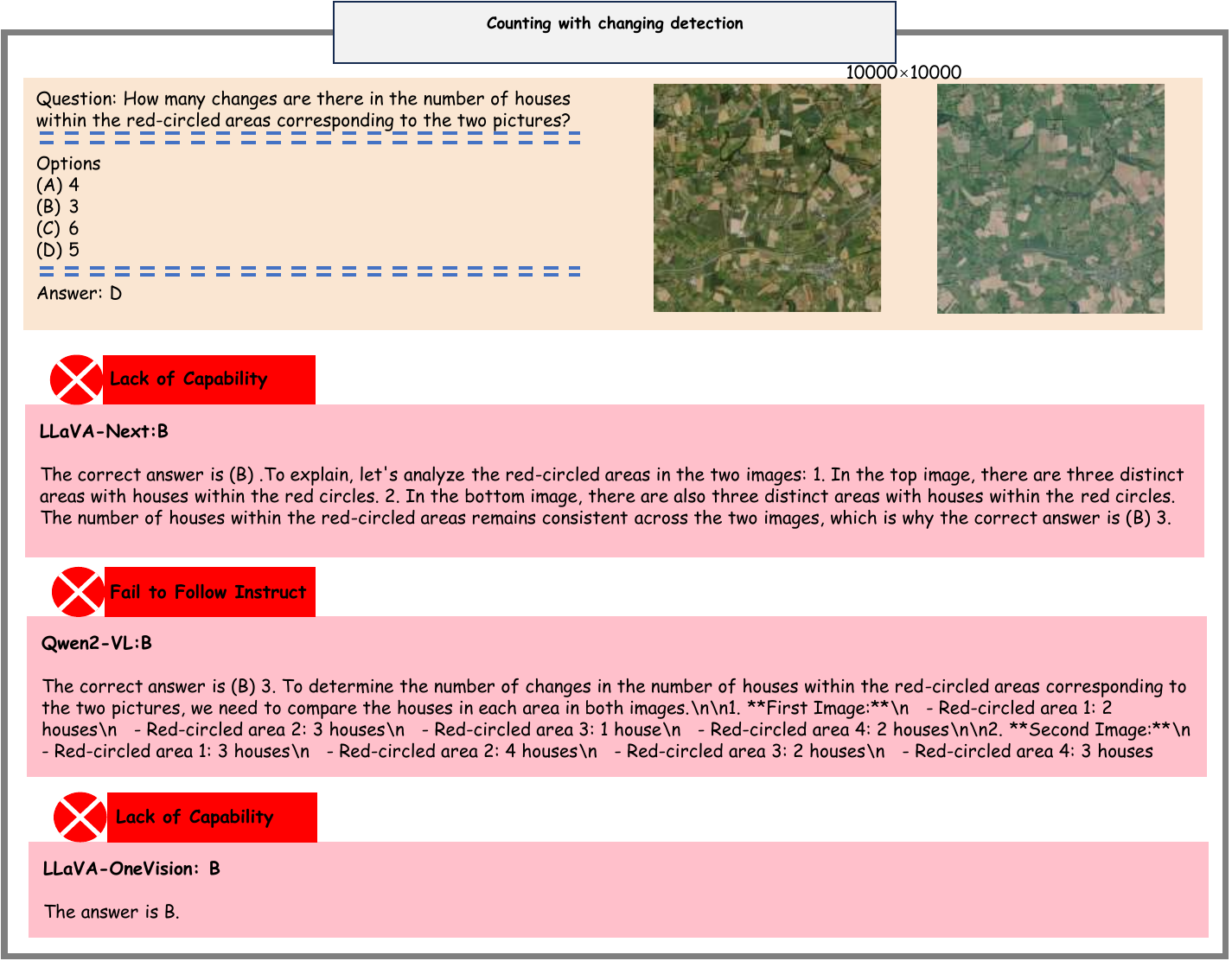}
    \caption{A sample case of Counting with change detection. }
    \label{fig:error_4}
\end{figure*}

\begin{figure*}
    \centering
    \includegraphics[width=1\linewidth]{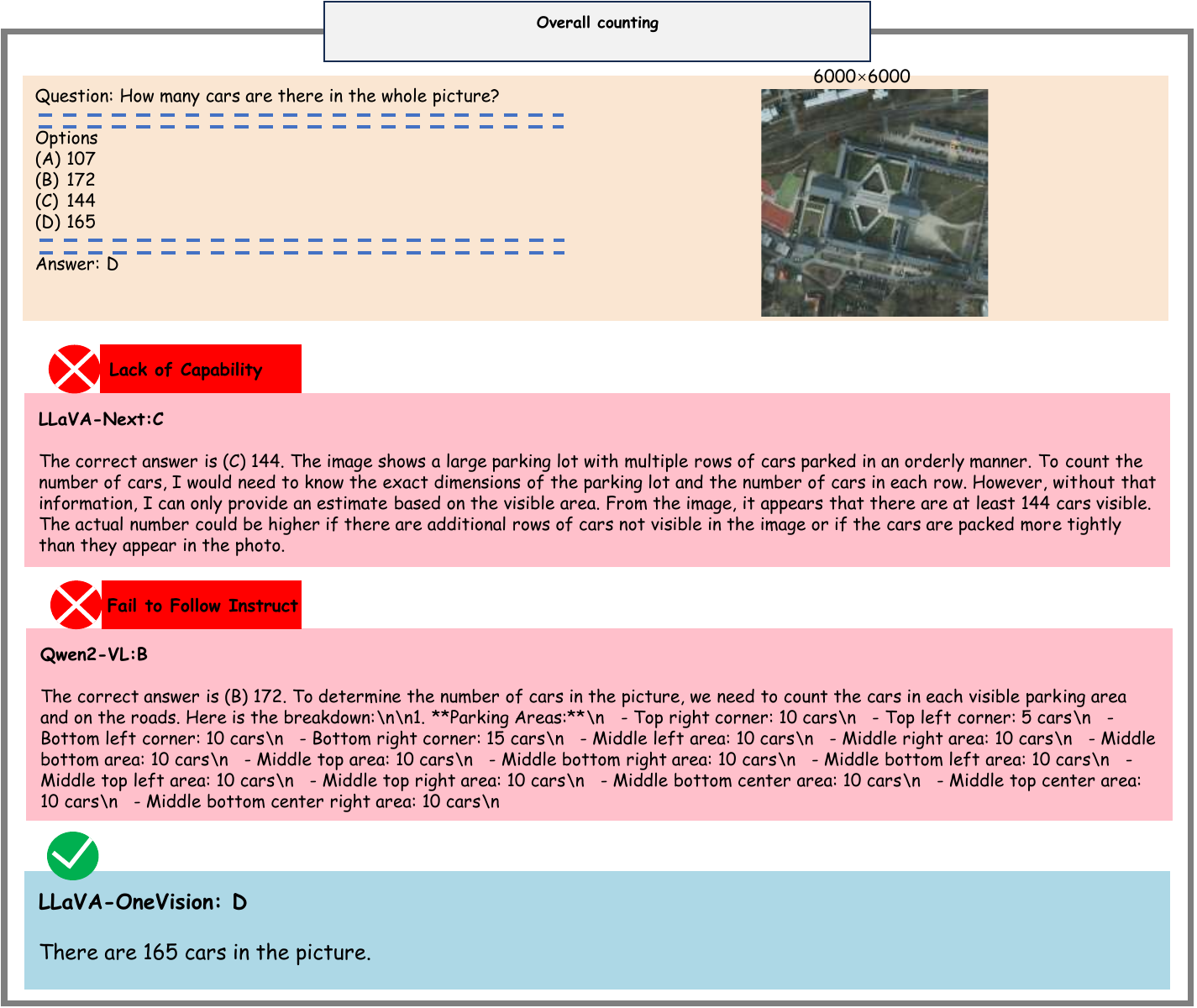}
    \caption{A sample case of Overall counting. }
    \label{fig:error_6}
\end{figure*}

\roundedboxorange{Fail to Follow Instruct}: 
MLLMs often misinterpret instructions, leading to errors. For instance, they may misunderstand specific conditions (see Fig.~\ref{fig:error_4}) or disregard the instructions entirely, generating errors for the image instead (see Fig.~\ref{fig:error_6}).

\clearpage

\begin{table*}[htp]
\centering
\caption{Table index of case study figures by sub-tasks (L-3 capability) with associated (error) categories for each MLLM.}
\label{tab:error_case}
\resizebox{1.\textwidth}{!}{%
\begin{tabular}{lllccc}
\toprule
Case Figure & L-2 task & L-3 task & LLaVa-Next & Qwen2-VL & LLaVA-OneVision \\
     \midrule
      \textcolor{red}{Fig.~\ref{fig:error_1}}    & Anomaly detection & Anomaly detection and interpretation & \roundedboxcyan{Lack of Capability} & \roundedboxred{Reasoning Error} & \roundedboxcyan{Lack of Capability} \\
      \textcolor{red}{Fig.~\ref{fig:error_2}}& Complex reasoning & Environmental condition reasoning & \roundedboxred{Reasoning Error} &  \roundedboxgreen{Correct} &  \roundedboxgreen{Correct} \\
      \textcolor{red}{Fig.~\ref{fig:error_3}}& Planning & Route planning & \roundedboxcyan{Lack of Capability} & \roundedboxcyan{Lack of Capability} & \roundedboxcyan{Lack of Capability} \\
      \textcolor{red}{Fig.~\ref{fig:error_4}}& Spatiotemporal reasoning & Counting with change detection &\roundedboxcyan{Lack of Capability} & \roundedboxorange{Fail to Follow Instruct}&\roundedboxcyan{Lack of Capability} \\
     \textcolor{red}{Fig.~\ref{fig:error_5}}& Complex reasoning & Counting with complex reasoning&\roundedboxcyan{Lack of Capability} & \roundedboxorange{Fail to Follow Instruct} &  \roundedboxgreen{Correct} \\
      \textcolor{red}{Fig.~\ref{fig:error_6}}& Counting & Overall counting &\roundedboxcyan{Lack of Capability} &\roundedboxcyan{Lack of Capability}&\roundedboxcyan{Lack of Capability} \\
      \textcolor{red}{Fig.~\ref{fig:error_7}}& Counting & Regional counting &\roundedboxcyan{Lack of Capability} &\roundedboxcyan{Lack of Capability}&\roundedboxcyan{Lack of Capability}\\
      \textcolor{red}{Fig.~\ref{fig:error_8}}& Scene classification & Overall land use classification & \roundedboxyellow{Perception Error} &  \roundedboxgreen{Correct} & \roundedboxyellow{Perception Error} \\
      \textcolor{red}{Fig.~\ref{fig:error_9}}&Scene classification & Regional land use classification &\roundedboxyellow{Perception Error} & \roundedboxgreen{Correct} & \roundedboxgreen{Correct}\\
      \textcolor{red}{Fig.~\ref{fig:error_10}}& Object properties & Object classification & \roundedboxgreen{Correct} & \roundedboxgreen{Correct}&\roundedboxcyan{Lack of Capability}\\
      \textcolor{red}{Fig.~\ref{fig:error_11}}& Object properties & Object color&  \roundedboxgreen{Correct} & \roundedboxyellow{Perception Error} & \roundedboxyellow{Perception Error} \\
      \textcolor{red}{Fig.~\ref{fig:error_12}}& Object properties & Object motion state & \roundedboxyellow{Perception Error} & \roundedboxblue{Lack of Knowledge} & \roundedboxblue{Lack of Knowledge} \\
      \textcolor{red}{Fig.~\ref{fig:error_13}}& Object spatial relationship & Object spatial relationship & \roundedboxyellow{Perception Error} & \roundedboxyellow{Perception Error}&  \roundedboxgreen{Correct} \\

\bottomrule
\end{tabular}%
}
\end{table*}

\clearpage

\begin{figure*}
    \centering
    \includegraphics[width=1\linewidth]{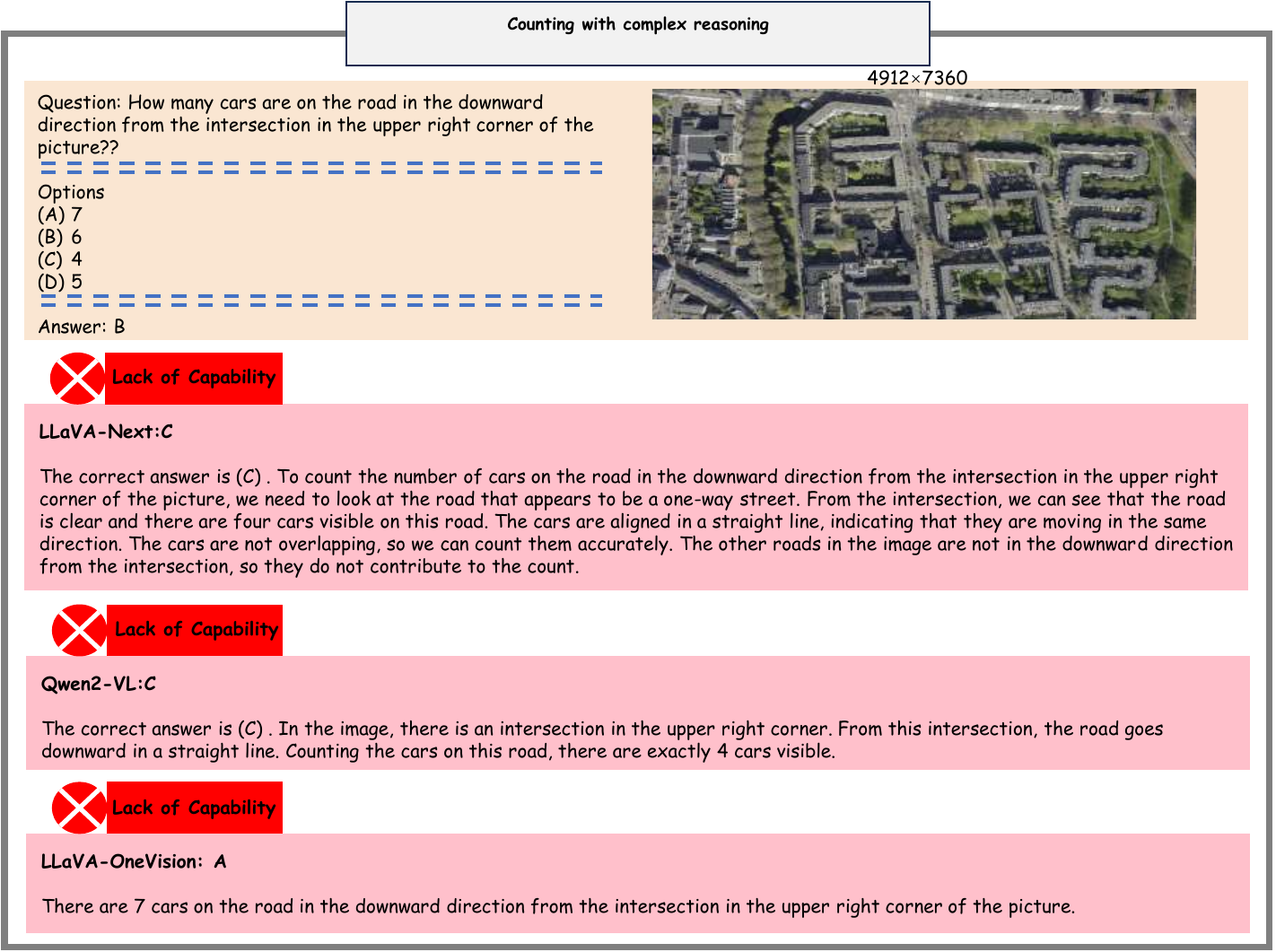}
    \caption{A sample case of Counting with complex reasoning. }
    \label{fig:error_5}
\end{figure*}
\begin{figure*}
    \centering
    \includegraphics[width=1\linewidth]{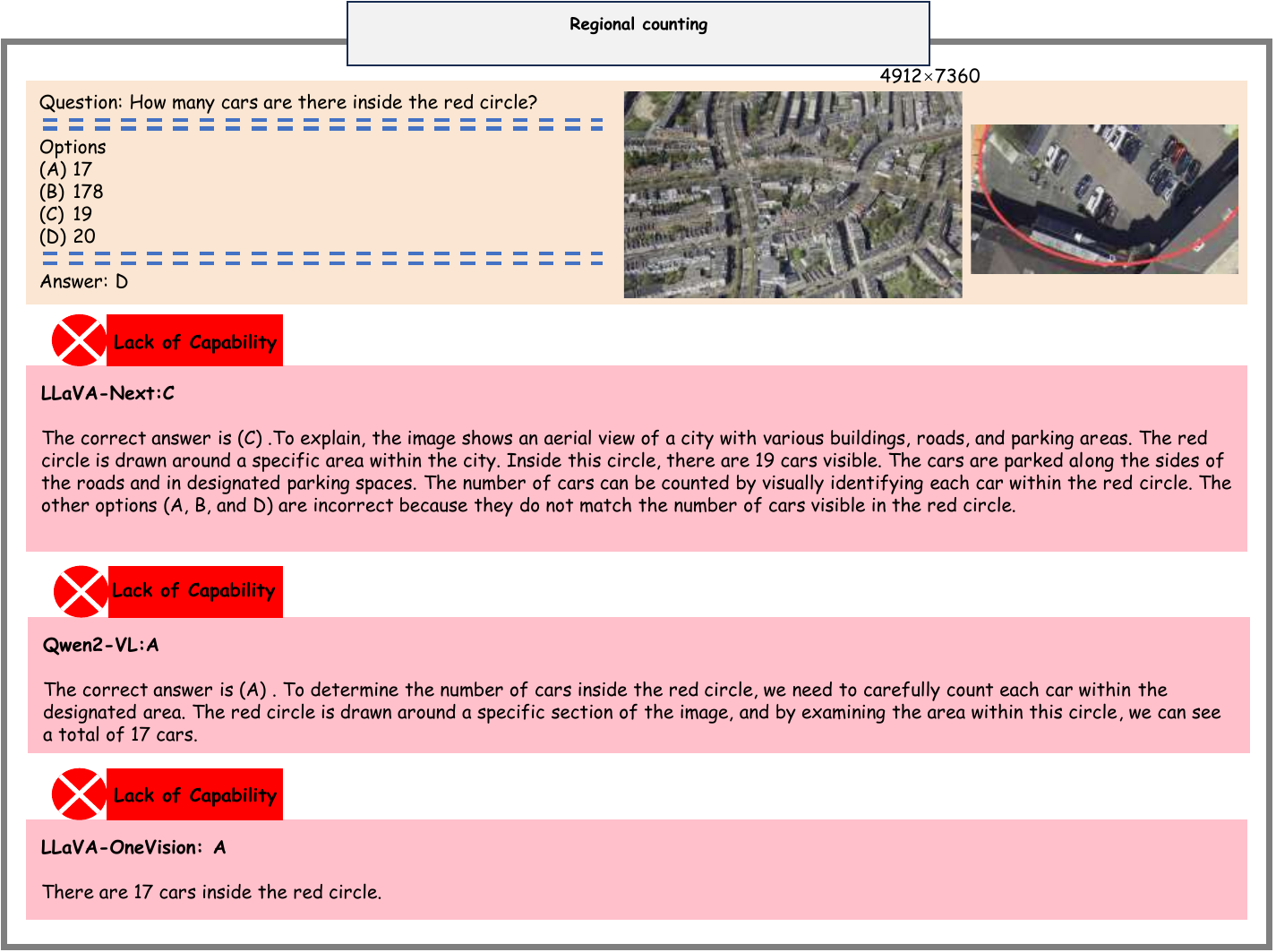}
    \caption{A sample case of Regional counting. }
    \label{fig:error_7}
\end{figure*}
\begin{figure*}
    \centering
    \includegraphics[width=1\linewidth]{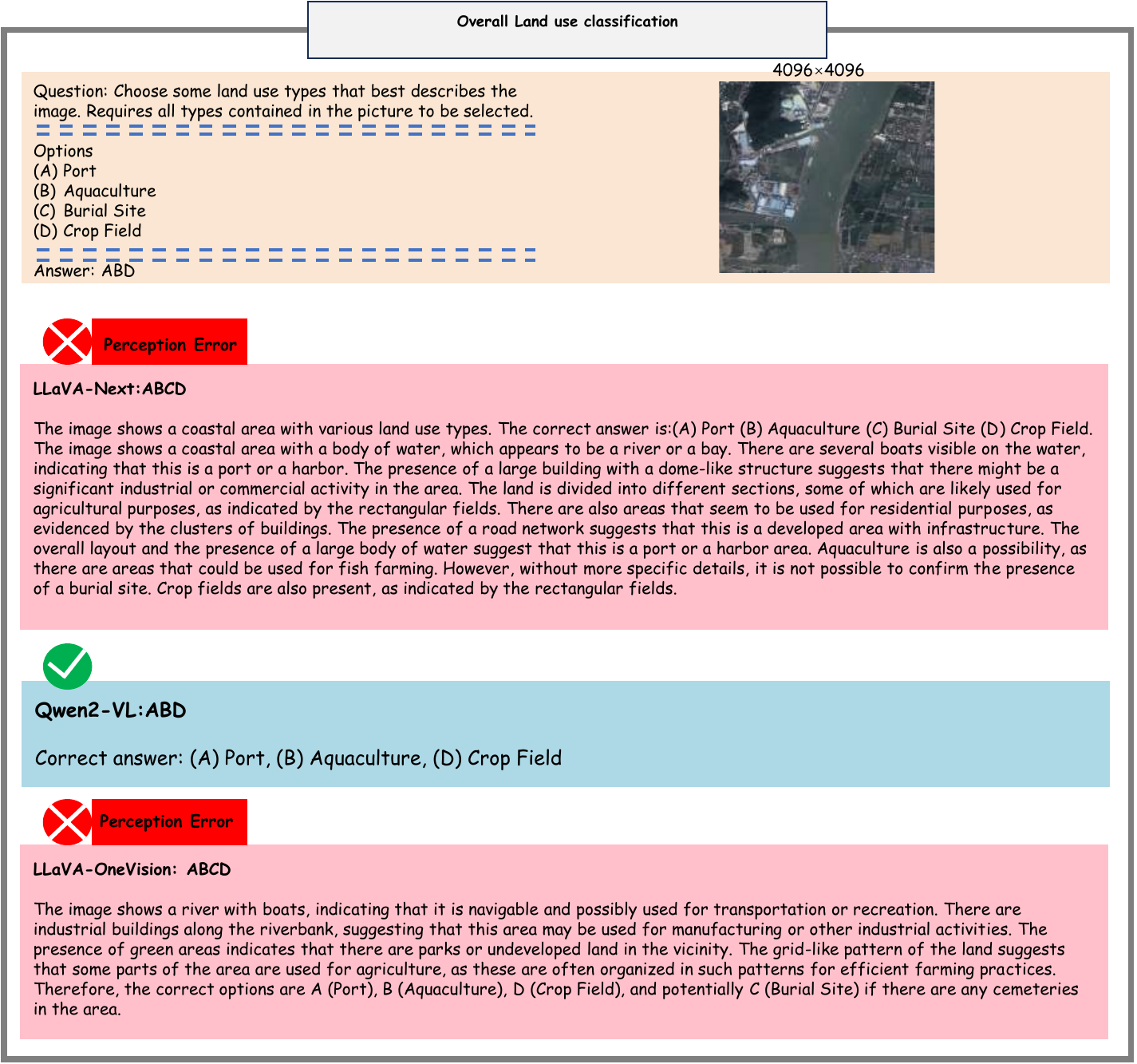}
    \caption{A sample case of Overall land use classification. }
    \label{fig:error_8}
\end{figure*}
\begin{figure*}
    \centering
    \includegraphics[width=1\linewidth]{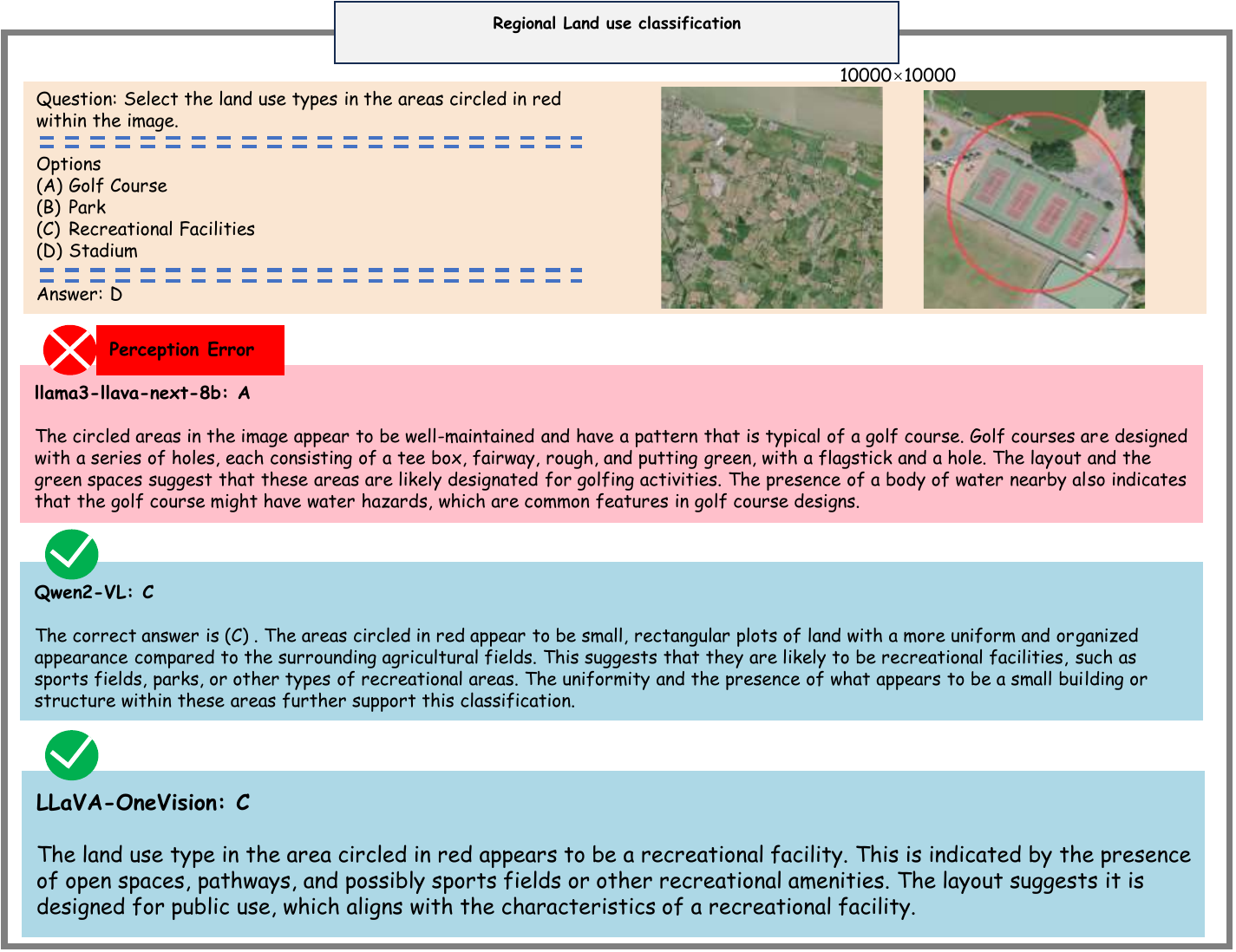}
    \caption{A sample case of Regional land use classification. }
    \label{fig:error_9}
\end{figure*}\begin{figure*}
    \centering
    \includegraphics[width=1\linewidth]{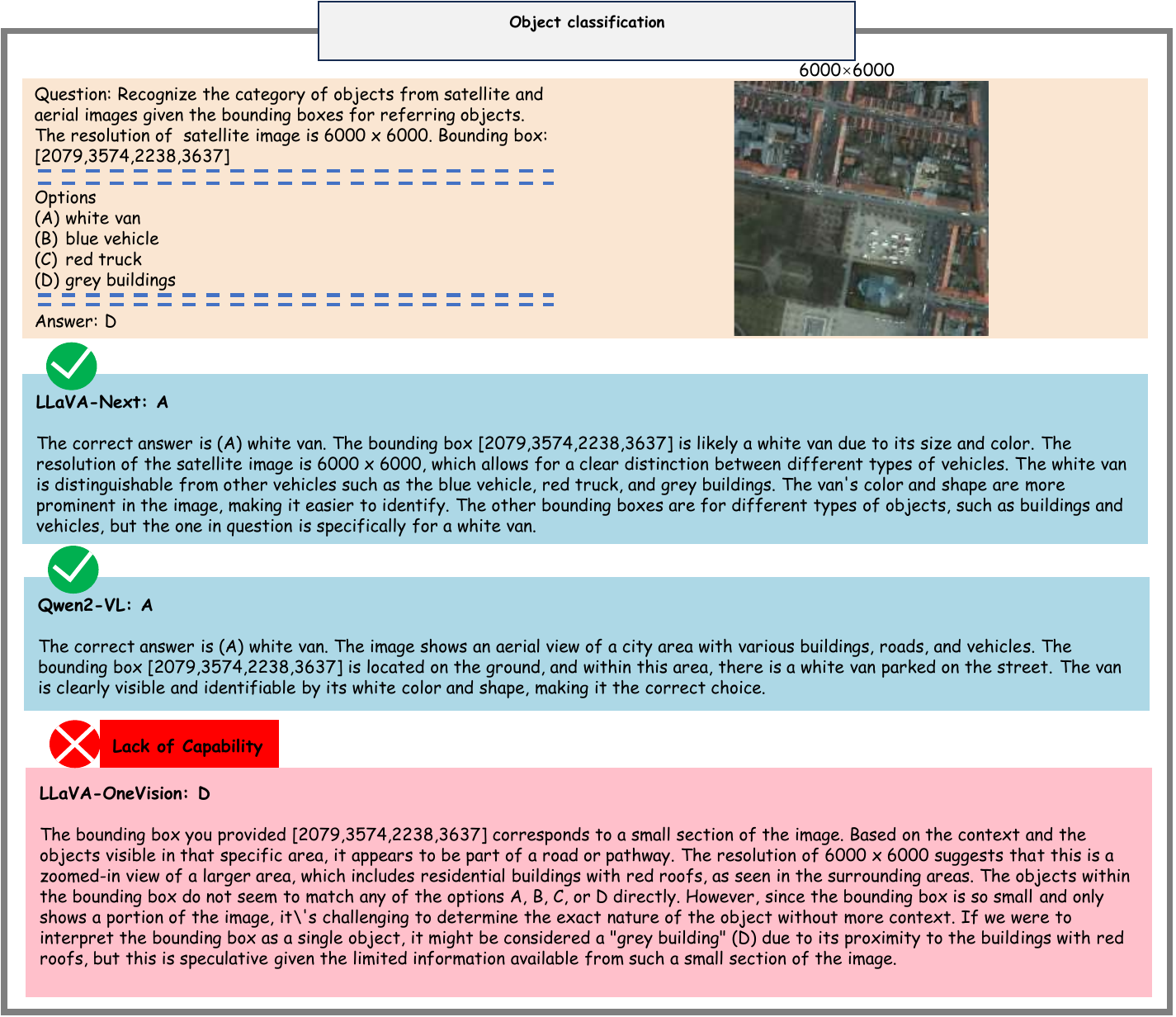}
    \caption{A sample case of Object classification. }
    \label{fig:error_10}
\end{figure*}
\begin{figure*}
    \centering
    \includegraphics[width=1\linewidth]{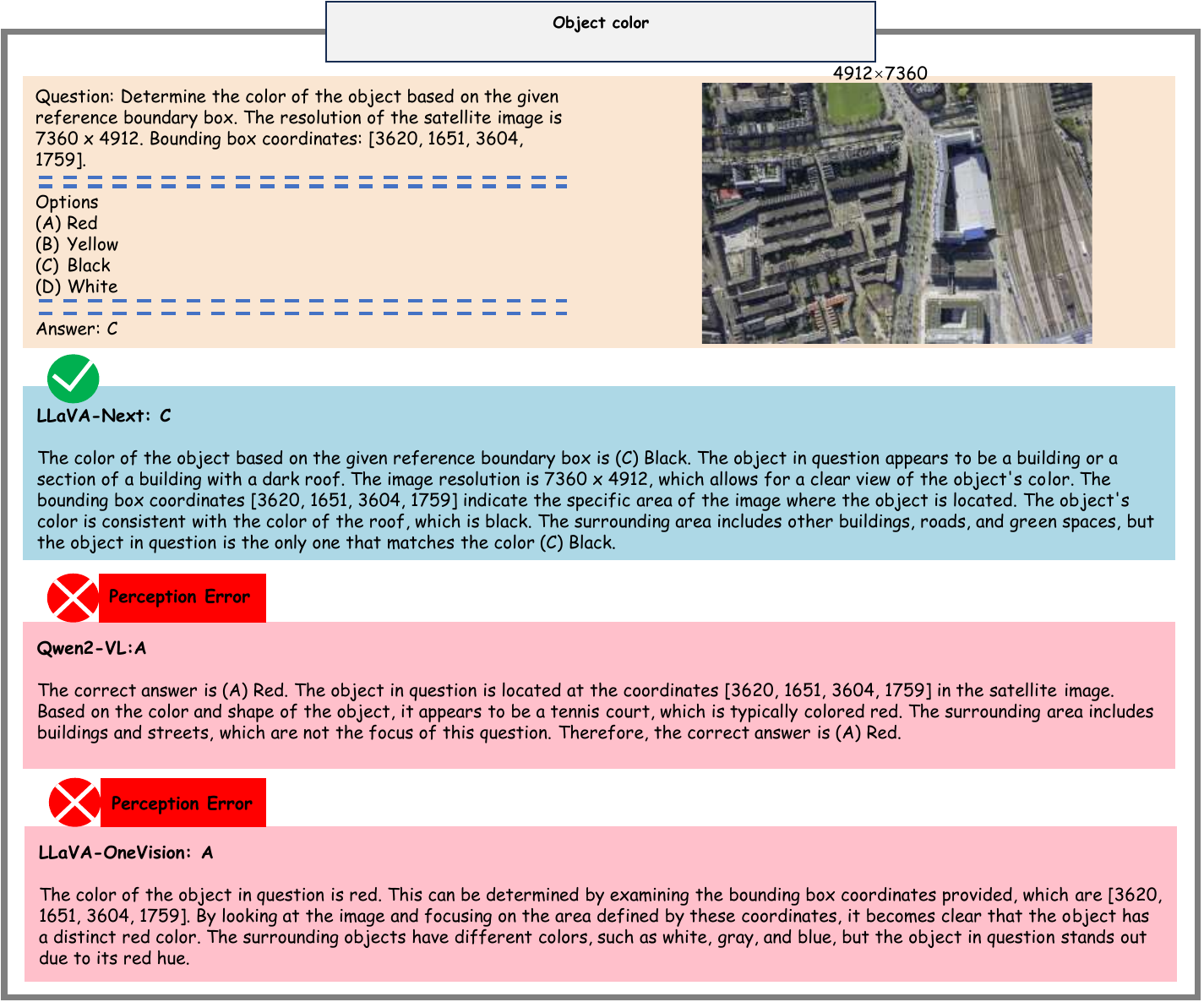}
    \caption{A sample case of Object color. }
    \label{fig:error_11}
\end{figure*}

\clearpage
\subsection{Datasheets}
\label{app-datasheets}

In this section, we document essential details about the proposed datasets and benchmarks following the CVPR Dataset and Benchmark guidelines and the template provided by Gebru \textit{et al.} \cite{datasheets}.

\subsubsection{Motivation}
The questions in this section are primarily intended to encourage dataset creators to clearly articulate their reasons for creating the dataset and to promote transparency about funding interests. The latter may be particularly relevant for datasets created for research purposes.
\begin{enumerate}
    \item \textit{``For what purpose was the dataset created?''}
    
    \textcolor{BurntOrange}{\textbf{A:}} 
    Existing benchmarks often use significantly smaller image sizes than those found in real-world RS scenarios, have limited annotation quality, and fail to account for key evaluation dimensions. To address these limitations, we introduce XLRS-Bench, a comprehensive benchmark designed to assess the perception and reasoning capabilities of MLLMs in ultra-high-resolution RS contexts.
    
    \item \textit{``Who created the dataset (\textit{e.g.}, which team, research group) and on behalf of which entity?''}
    
    \textcolor{BurntOrange}{\textbf{A:}} The dataset was created by the following authors:
    \begin{itemize}
      \item Fengxiang Wang (National University of Defense Technology)
      \item Hongzhen Wang (Tsinghua University)
      \item Mingshuo Chen (Beijing University of Posts and Telecommunications)
      \item Di Wang (Wuhan University, Zhongguancun Academy)
      \item Yulin Wang (Tsinghua University)
      \item Zonghao Guo (Tsinghua University)
      \item Qiang Ma (Tsinghua University)
      \item Long Lan (National University of Defense Technology)
      \item Wenjing Yang (National University of Defense Technology)
      \item Jing Zhang (Wuhan University)
      \item Zhiyuan Liu (Tsinghua University)
      \item Maosong Sun (Tsinghua University)
    \end{itemize}
    
    \item \textit{``Who funded the creation of the dataset?''}
    
    \textcolor{BurntOrange}{\textbf{A:}}
    The dataset creation was funded by the affiliations of the authors involved in this work.
\end{enumerate}

\subsubsection{Composition}
Most of the questions in this section are intended to provide dataset consumers with the information they need to make informed decisions about using the dataset for their chosen tasks. Some of the questions are designed to elicit information about compliance with the EU’s General Data Protection Regulation (GDPR) or comparable regulations in other jurisdictions. Questions that apply only to datasets that relate to people are grouped together at the end of the section. We recommend taking a broad interpretation of whether a dataset relates to people. For example, any dataset containing text that was written by people relates to people.
\begin{enumerate}
    \item \textit{``What do the instances that comprise our datasets represent (\textit{e.g.}, documents, photos, people, countries)?''}
    
    \textcolor{BurntOrange}{\textbf{A:}} The dataset primarily consists of ultra-high-resolution remote sensing images captured by satellites, along with their corresponding textual annotations. All datasets utilized in XLRS-Bench are publicly accessible and nonprofit.
    
    \item \textit{``How many instances are there in total (of each type, if appropriate)?''}
    
    \textcolor{BurntOrange}{\textbf{A:}} XLRS-Bench includes 1,400 ultra-high-resolution images, with 840 reaching a resolution of 10,000 $\times$ 10,000. Additionally, for these ultra-high-resolution images, we have provided 934 detailed captions, 32,389 VQA pairs, and 12,619 visual grounding instances.

    \item \textit{``Does the dataset contain all possible instances or is it a sample (not necessarily random) of instances from a larger set?''}
    
    \textcolor{BurntOrange}{\textbf{A:}} The images in XLRS-Bench are sourced from existing detection \cite{dota,itcvd} and segmentation \cite{minifrance,hrscd} datasets, but all textual annotations were independently created by us.
    
    \item \textit{``Is there a label or target associated with each instance?''}
    
    \textcolor{BurntOrange}{\textbf{A:}} Yes, for these ultra-high-resolution images, we have provided 934 detailed captions, 32,389 VQA pairs, and 12,619 visual grounding instances.
    
    \item \textit{``Is any information missing from individual instances?''}
    
    \textcolor{BurntOrange}{\textbf{A:}} No, each individual instance is complete.
    
    \item \textit{``Are relationships between individual instances made explicit (\textit{e.g.}, users’ movie ratings, social network links)?''}
    
    \textcolor{BurntOrange}{\textbf{A:}} Yes, the relationship between individual instances is explicit.
    
    \item \textit{``Are there recommended data splits (\textit{e.g.}, training, development/validation, testing)?''}
    
    \textcolor{BurntOrange}{\textbf{A:}} 
    The dataset is designed to evaluate the perception and reasoning abilities of MLLMs, so we recommend using it in its entirety as a test set.
    
    \item \textit{``Is the dataset self-contained, or does it link to or otherwise rely on external resources (\textit{e.g.}, websites, tweets, other datasets)?''}
    
    \textcolor{BurntOrange}{\textbf{A:}} XLRS-Bench is self-contained and will be open-sourced on platforms like Hugging Face, integrated into evaluation tools such as LLMs-Eval~\cite{zhang2024lmmsevalrealitycheckevaluation,lmms_eval2024} for easy use.
    
    \item \textit{``Does the dataset contain data that might be considered confidential (\textit{e.g.}, data that is protected by legal privilege or by doctor–patient confidentiality, data that includes the content of individuals’ non-public communications)?''}
    
    \textcolor{BurntOrange}{\textbf{A:}} No, all data are clearly licensed.
    
    \item \textit{``Does the dataset contain data that, if viewed directly, might be offensive, insulting, threatening, or might otherwise cause anxiety?''}
    
    \textcolor{BurntOrange}{\textbf{A:}} No, XLRS-Bench does not contain any data with negative information.
\end{enumerate}

\subsubsection{Collection Process}
In addition to the goals outlined in the previous section, the questions in this section are designed to elicit information that may help researchers and practitioners create alternative datasets with similar characteristics. Again, questions that apply only to datasets that relate to people are grouped together at the end of the section.
\begin{enumerate}
    \item \textit{``How was the data associated with each instance acquired?''}
    
    \textcolor{BurntOrange}{\textbf{A:}} 
    The images in XLRS-Bench are sourced from existing detection \cite{dota,itcvd} and segmentation \cite{minifrance,hrscd} datasets. We enrich these ultra-high-resolution images with manual annotations, including 934 detailed captions, 32,389 VQA pairs, and 12,619 visual grounding instances.
    
    \item \textit{``What mechanisms or procedures were used to collect the data (\textit{e.g.}, hardware apparatuses or sensors, manual human curation, software programs, software APIs)?''}
    
    \textcolor{BurntOrange}{\textbf{A:}} We employed professional annotation and quality control teams to complete the annotations for VQA and Visual Grounding tasks. For the Image Captioning task, we developed a semi-automated pipeline. Detailed information can be found in Section 3.2 of the main text.
    
    \item \textit{``If the dataset is a sample from a larger set, what was the sampling strategy (\textit{e.g.}, deterministic, probabilistic with specific sampling probabilities)?''} 
    
    \textcolor{BurntOrange}{\textbf{A:}} Please refer to the details listed in the main text Section 3.2.
\end{enumerate}

\subsubsection{Preprocessing, Cleaning, and Labeling}
The questions in this section are intended to provide dataset
consumers with the information they need to determine whether the “raw” data has been processed in ways that are compatible with their chosen tasks. For example, text that has been converted into a ``bag-of-words" is not suitable for tasks involving word order.
\begin{enumerate}
    \item \textit{``Was any preprocessing/cleaning/labeling of the data done (\textit{e.g.}, discretization or bucketing, tokenization, part-of-speech tagging, SIFT feature extraction, removal of instances, processing of missing values)?''}
    
    \textcolor{BurntOrange}{\textbf{A:}} Yes. During image collection, we prioritized selecting valuable satellite images for annotation. For linguistic annotation, three Level-3 subtasks—Regional Land Use Classification, Regional Counting, and Regional Counting with Change Detection—were marked with red circles. This method, mimicking human interaction, was essential for providing clear, fine-grained region-level analysis on ultra-high-resolution images.

    \item \textit{``Was the `raw' data saved in addition to the preprocessed/cleaned/labeled data (\textit{e.g.}, to support unanticipated future uses)?''} 
    
    \textcolor{BurntOrange}{\textbf{A:}} Yes, raw data is accessible.
    
    \item \textit{``Is the software that was used to preprocess/clean/label the data available?''} 
    
    \textcolor{BurntOrange}{\textbf{A:}} Yes, the necessary software used to preprocess and clean the data is publicly available.
\end{enumerate}

\subsubsection{Uses}
The questions in this section are intended to encourage dataset creators to reflect on tasks for which the dataset should and should not be used. By explicitly highlighting these tasks, dataset creators can help dataset consumers make informed decisions, thereby avoiding potential risks or harms.
\begin{enumerate}
    \item \textit{``Has the dataset been used for any tasks already?''} 
    
    \textcolor{BurntOrange}{\textbf{A:}} 
    No.
    
    \item \textit{``Is there a repository that links to any or all papers or systems that use the dataset?''} 
    
    \textcolor{BurntOrange}{\textbf{A:}} Yes, we will provide such links in the GitHub and the Huggingface repository.
    
    \item \textit{``What (other) tasks could the dataset be used for?''} 
    
    \textcolor{BurntOrange}{\textbf{A:}} XLRS-Bench provides extensive annotations for VQA, Grounding, and Captioning tasks. In addition to evaluating the perception and reasoning capabilities of existing MLLMs, it can also be used to assess models specifically designed for these tasks.
    
    \item \textit{``Is there anything about the composition of the dataset or the way it was collected and preprocessed/cleaned/labeled that might impact future uses?''} 
    
    \textcolor{BurntOrange}{\textbf{A:}} No.
    
    \item \textit{``Are there tasks for which the dataset should not be used?''} 
    
    \textcolor{BurntOrange}{\textbf{A:}} N/A.
\end{enumerate}

\subsubsection{Distribution}
Dataset creators should provide answers to these questions prior to distributing the dataset either internally within the entity on behalf of which the dataset was created or externally to third parties.
\begin{enumerate}
    \item \textit{``Will the dataset be distributed to third parties outside of the entity (\textit{e.g.}, company, institution, organization) on behalf of which the dataset was created?''} 
    
    \textcolor{BurntOrange}{\textbf{A:}} No. The datasets will be made publicly accessible to the research community.
    
    \item \textit{``How will the dataset be distributed (\textit{e.g.}, tarball on website, API, GitHub)?''} 
    
    \textcolor{BurntOrange}{\textbf{A:}} We will provide XLRS-Bench in the GitHub and the Huggingface repository.
    
    \item \textit{``When will the dataset be distributed?''} 
    
    \textcolor{BurntOrange}{\textbf{A:}} We will create a repository to release the data once the paper is officially published, ensuring compliance with the anonymity principle.
    
    \item \textit{``Will the dataset be distributed under a copyright or other intellectual property (IP) license, and/or under applicable terms of use (ToU)?''} 
    
    \textcolor{BurntOrange}{\textbf{A:}} Yes, the dataset will be released under the Creative Commons Attribution-NonCommercial-ShareAlike 4.0 International License.
    
    \item \textit{``Have any third parties imposed IP-based or other restrictions on the data associated with the instances?''} 
    
    \textcolor{BurntOrange}{\textbf{A:}} No.
    
    \item \textit{``Do any export controls or other regulatory restrictions apply to the dataset or to individual instances?''} 
    
    \textcolor{BurntOrange}{\textbf{A:}} No.    
\end{enumerate}

\subsubsection{Maintenance}
As with the questions in the previous section, dataset creators should provide answers to these questions prior to distributing the dataset. The questions in this section are intended to encourage dataset creators to plan for dataset maintenance and communicate this plan to dataset consumers.
\begin{enumerate}
    \item \textit{``Who will be supporting/hosting/maintaining the dataset?''} 
    
    \textcolor{BurntOrange}{\textbf{A:}} The authors of this work serve to support, host, and maintain the datasets.
    
    \item \textit{``How can the owner/curator/manager of the dataset be contacted (\textit{e.g.}, email address)?''} 
    
    \textcolor{BurntOrange}{\textbf{A:}} The curators can be contacted via the email addresses listed on our paper or webpage.
    
    \item \textit{``Is there an erratum?''} 
    
    \textcolor{BurntOrange}{\textbf{A:}} There is no explicit erratum; updates and known errors will be specified in future versions.
    
    \item \textit{``Will the dataset be updated (\textit{e.g.}, to correct labeling errors, add new instances, delete instances)?''} 
    
    \textcolor{BurntOrange}{\textbf{A:}} Future updates (if any) will be posted on the dataset website.
    
    \item \textit{``Will older versions of the dataset continue to be supported/hosted/maintained?''} 
    
    \textcolor{BurntOrange}{\textbf{A:}} 
    Yes. This initial release will be updated in the future, with older versions replaced as new updates are posted.
    
    \item \textit{``If others want to extend/augment/build on/contribute to the dataset, is there a mechanism for them to do so?''} 
    
    \textcolor{BurntOrange}{\textbf{A:}} Yes, we will provide detailed instructions for future extensions.
\end{enumerate}

\subsection{Limitation and Potential Societal Impact}
\label{app-limitation}
In this section, we discuss the limitations and potential societal impact of this work.

\subsubsection{Potential Limitations}
While \textbf{XLRS-Bench} provides a comprehensive benchmark for evaluating the perception and reasoning capabilities of MLLMs, there are several limitations to consider:
\begin{itemize}
    \item \textbf{Scope of Sensors:} Although our benchmark includes 1,400 utlra-high-resolution visible light remote sensing images, it may not cover all possible real-world scenarios. There could be additional sensor data, like multispectral data that were not included in this study, potentially limiting the generalizability of our findings.

    \item \textbf{Model and Dataset Diversity:} 
    In this paper, we extensively evaluated both general-purpose and RS-specific MLLMs. As new models emerge, their evaluation results will be added to our open-source leaderboard. Additionally, XLRS-Bench will also be expanded in dataset size and task diversity.

    \item \textbf{Multilingual Support:} XLRS-Bench currently supports both Chinese and English, surpassing the single-language limitations of most remote sensing benchmarks~\cite{vrsbench}. In the future, we aim to extend support to languages like Spanish and French.

\end{itemize}

\subsubsection{Potential Negative Societal Impact}
\begin{itemize}
    \item \textbf{Safety Risks:} XLRS-Bench is designed to evaluate the performance of vision-language multimodal models in ultra-high-resolution remote sensing scenarios. However, excessive reliance on evaluation datasets may lead to overconfidence in autonomous systems, such as multimodal large models. It is crucial to implement adequate safety measures and human supervision when deploying these MLLMs to ensure public safety.

    \item \textbf{Environmental Impact:} 
    Training MLLMs on large datasets and evaluating them using XLRS-Bench requires a certain amount of computational resources. To facilitate future research, we will maintain a leaderboard of MLLMs, removing the need for repeated evaluations of existing models.

    \item \textbf{Bias and Fairness:} 
    XLRS-Bench, with its 16 Level-3 capabilities, is tailored for evaluating ultra-high-resolution remote sensing scenarios. However, it remains limited in comprehensiveness and may exhibit biases. For instance, disaster prediction in anomaly reasoning relies solely on satellite imagery, providing warnings but reflecting inherent biases. Effective decision-making demands the integration of local meteorological and hydrological data. In the future, we aim to expand the evaluation dimensions and datasets to deepen insights into ultra-high-resolution remote sensing applications.
    
\end{itemize}



\end{document}